\documentclass{article} 
\usepackage{iclr2027_conference,times}

\usepackage{amsmath,amsfonts,bm}

\def\eqref#1{equation~\ref{#1}}

\def\1{\bm{1}}

\DeclareMathAlphabet{\mathsfit}{\encodingdefault}{\sfdefault}{m}{sl}
\SetMathAlphabet{\mathsfit}{bold}{\encodingdefault}{\sfdefault}{bx}{n}

\usepackage{hyperref}
\usepackage{url}
\usepackage{booktabs}
\usepackage{multirow}
\usepackage{makecell}
\usepackage{graphicx}
\usepackage{capt-of}
\usepackage{tabularx}
\usepackage{array}
\usepackage{ragged2e}
\usepackage{graphicx}
\usepackage{booktabs}
\usepackage{multirow}
\usepackage{capt-of}
\usepackage{wrapfig}
\usepackage{amssymb}
\usepackage{marvosym}

\definecolor{ModelSmallBase}{HTML}{756BB1}
\definecolor{ModelMediumBase}{HTML}{70A84B}
\definecolor{ModelLargeBase}{HTML}{E8BE48}
\newcommand{\modelshade}[2]{{%
  \setlength{\fboxsep}{1.8pt}%
  \colorbox{#1}{\makebox[10em][l]{\strut #2}}}}
\newcommand{\modelsmall}[1]{\modelshade{ModelSmallBase!12}{#1}}
\newcommand{\modelmedium}[1]{\modelshade{ModelMediumBase!12}{#1}}
\newcommand{\modellarge}[1]{\modelshade{ModelLargeBase!14}{#1}}
\title{AnesTRACE: Benchmarking Intraoperative Anesthesia from Multimodal Perception to Multi-step Decision-Making}

\author{
Ziwei Huang$^{1}$,
Qi Gao$^{2}$,
Zhe Ji$^{3}$\thanks{Work done while at Zhejiang University.},
Yuanyuan Yao$^{2}$,
Fengjiang Zhang$^{2}$,
Min Yan$^{2}$\textsuperscript{\Letter}
\\[0.35em]
\textbf{
Zhongle Xie$^{1}$\textsuperscript{\Letter},
Gang Chen$^{4}$
}\\[0.9em]
$^{1}$School of Software Technology, Zhejiang University \\[0.2em]
$^{2}$Department of Anesthesiology, The Second Affiliated Hospital of Zhejiang University School of Medicine \\[0.2em]
$^{3}$School of Software, Central South University \\[0.2em]
$^{4}$Zhejiang University
}

\iclrfinalcopy 
\begin{document}

\maketitle
\fancyhead{}
\renewcommand{\headrulewidth}{0pt}

\begin{abstract}

Intraoperative anesthesia requires systems to interpret evolving multimodal evidence, recommend timely management, and revise decisions as patient states change, yet existing benchmarks usually isolate perception or single-point reasoning.
We introduce AnesTRACE, an evaluation suite comprising AnesTRACE-Bench and AnesTRACE-Eval.
Built from public perioperative datasets with anesthesiologist annotation, AnesTRACE-Bench evaluates Intraoperative Perception, Single-point Anesthesia Decision-Making, and Multi-step Anesthesia Decision-Making.
AnesTRACE-Eval assesses open-ended responses through anesthesiologist-defined criteria for Clinical Correctness, Evidence Grounding, Task Completeness, and Safety, with Temporal Consistency for multi-step decisions; its domain-specific evaluator is trained by supervised fine-tuning and preference alignment on expert-reviewed judgments.
Across more than 30 models, fine-grained visual grounding and intervention selection remain difficult: the leading model reaches only 32.2 mIoU for TEE visual grounding and retains a 17.5\% Major/Critical Safety Error Rate in multi-step management.
Evaluator alignment with anesthesiologists improves across both training stages, while the best decision quality is accompanied by a 74.3-second P95 Latency.
These results show that aggregate performance alone does not establish safe, timely longitudinal decision-making.
We release our code at \url{https://zjudbxai.github.io/AnesTRACE/}.

\end{abstract}

\section{Introduction}

\begin{figure*}[t]
    \centering
    \setlength{\abovecaptionskip}{0.2cm}
    \setlength{\belowcaptionskip}{-0.3cm}
    \includegraphics[width=\textwidth]{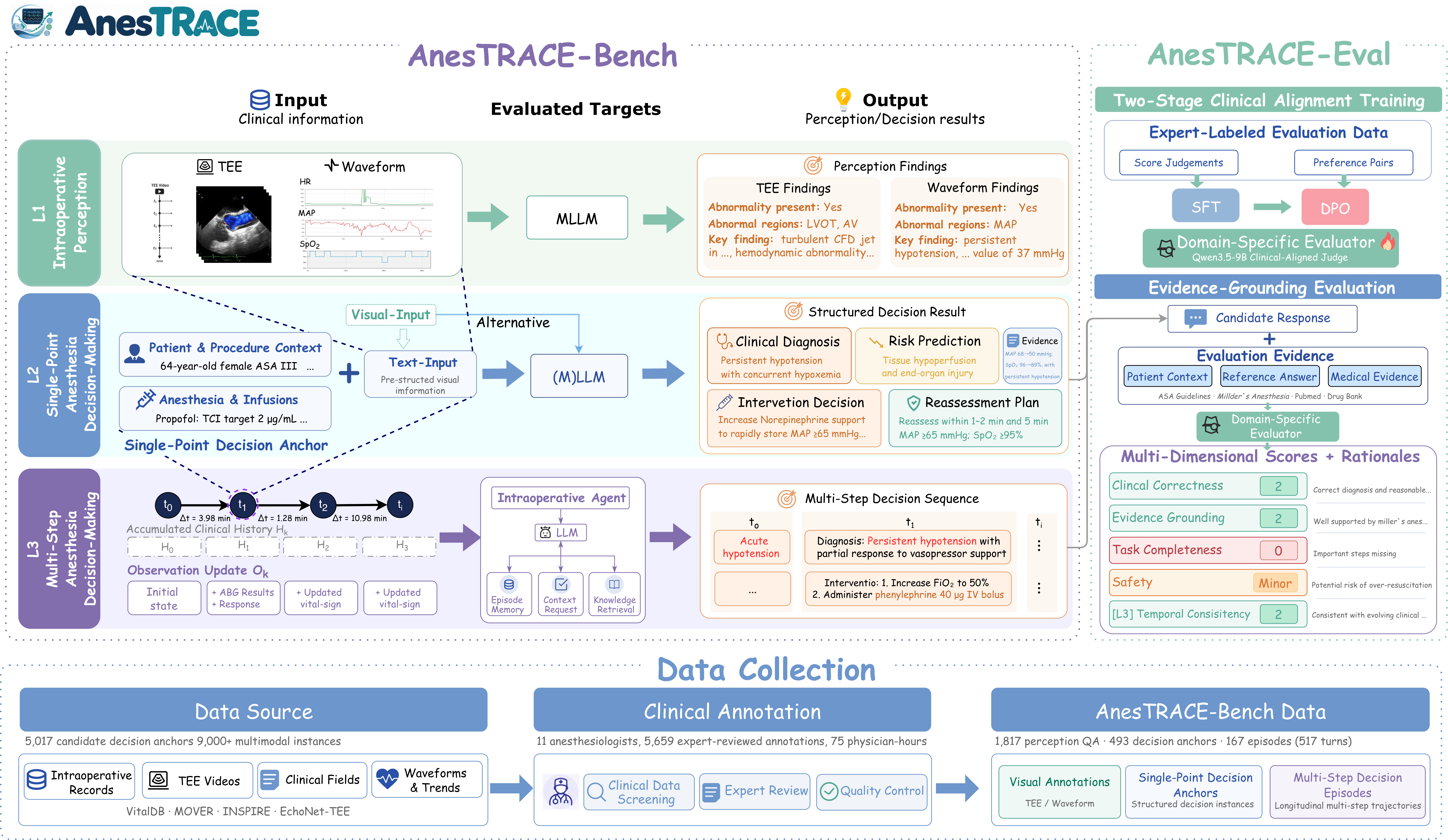}
    \caption{\textbf{Overview of AnesTRACE.} The bottom shows data collection and annotation, the upper left presents the three-level AnesTRACE-Bench task hierarchy, and the upper right shows the AnesTRACE-Eval framework.}
    \label{fig:anestrace_overview}
    \vspace{-3mm}
\end{figure*}

Intraoperative anesthesia couples interpretation, intervention, and reassessment under changing physiological conditions, where delayed or inappropriate management can have immediate consequences~\citep{nair2017intraoperative,asa2025monitoring,saugel2025haemodynamic}.
Anesthesiologists integrate heterogeneous signals, procedural context, ongoing treatment, and responses to prior interventions; the same physiological change can therefore require different management across intraoperative trajectories~\citep{saugel2025haemodynamic}.
Medical AI intended for this setting must connect available evidence to management, update decisions over time, and respond within seconds.

\enlargethispage{\baselineskip}

Existing benchmarks cover complementary pieces of the process but not the complete intraoperative decision pathway.
Knowledge benchmarks emphasize self-contained questions~\citep{medQA, medxpertqa, multimedQA}; multimodal benchmarks test interpretation without subsequent management~\citep{omnimedvqa, ecg_qa}; and agent or longitudinal benchmarks introduce tool use and evolving records outside repeated intraoperative management~\citep{agentclinic, medagentbench, longmedbench}.
Anesthesiology benchmarks remain largely text-based and centered on specialty knowledge, consultation, or single-response clinical reasoning\mbox{\citep{Anessuite,CAB}}.
A shared testbed is therefore needed to separate perception, management reasoning, and decision updating while measuring response latency.

Open-ended management also complicates evaluation because several plans may be acceptable, while consequential omissions or unsafe recommendations remain clinically important.
Repeated anesthesiologist review is costly, whereas holistic LLM judging may overlook decision-time evidence, temporal dependencies, or safety failures~\citep{healthbench,medhelm}.
Scalable evaluation consequently requires explicit evidence-grounded, temporal, and safety-sensitive criteria.

We introduce \textbf{AnesTRACE}, comprising \textit{AnesTRACE-Bench} and \textit{AnesTRACE-Eval} (Figure~\ref{fig:anestrace_overview}).
Intraoperative practice requires different capabilities as clinicians move from interpreting monitoring data to selecting an intervention and reassessing it after the patient state changes.
To reflect this workflow, AnesTRACE-Bench varies both the evaluated object and the information interface across \textbf{Intraoperative Perception (L1)}, \textbf{Single-point Anesthesia Decision-Making (L2)}, and \textbf{Multi-step Anesthesia Decision-Making (L3)}.
Accordingly, L1 tests whether existing multimodal large language models (MLLMs) can extract findings from raw visual and physiological evidence, while L2 tests whether models can convert a standardized patient-state description into management at one decision time.
L3 then places language models in a common agent framework to test information requests, memory, and management revision across successive decision points.
The agents share bounded patient-information and knowledge tools, episode memory, and trajectory replay that reveals recorded states without allowing their proposals to alter the clinical trajectory.
This controlled progression localizes whether a failure arises from perception, intervention reasoning, evidence acquisition, or longitudinal decision updating.

AnesTRACE-Eval addresses the aforementioned evaluation problem by translating anesthesiologist-defined atomic criteria into separate judgments for Clinical Correctness, Evidence Grounding, Task Completeness, and safety, while Temporal Consistency evaluates decision updating over an L3 episode.
The criteria permit clinically acceptable alternatives, and safety remains separately reported so that unsafe recommendations are not obscured by stronger performance on other dimensions.
A domain-specific evaluator learns this protocol through supervised fine-tuning and preference alignment on expert-reviewed judgments, enabling scalable evaluation while retaining explicit rationales for each applicable dimension.

Through benchmarking more than 30 models, we find that fine-grained perception and intervention selection remain weak, while the leading model retains a 17.5\% L3 Safety Error Rate and requires 74.3 seconds for a full response for multi-step decisions, exceeding the 30-second operating-standard requirement.
Clinical history, memory, and tools improve decision updating, while evaluator alignment with anesthesiologists improves after both training stages.

To summarize, our contributions are threefold:
\begin{itemize}
    \item We introduce AnesTRACE-Bench, which diagnoses three complementary intraoperative capabilities by varying evidence representation, information access, and temporal scope within a shared clinical domain.
    \item We develop AnesTRACE-Eval, an anesthesiologist-aligned protocol and domain-specific evaluator for evidence-grounded, safety-sensitive, and temporally consistent assessment of open-ended management responses.
    \item We provide a systematic evaluation of 30 models and controlled analyses of the factors that shape clinical quality, latency, and evidence-acquisition cost across the three settings.
\end{itemize}
\section{Related Work}
\label{sec:related_work}

\noindent\textbf{Perioperative Prediction and Clinical Decision-Making.}
Perioperative AI has primarily targeted constrained prediction or control, including hypotension prediction~\citep{hatib2018machine,wijnberge2020effect}, infusion adjustment~\citep{miyaguchi2021predicting}, and closed-loop anesthetic administration~\citep{puri2016multicenter,pasin2017closed}.
These studies establish the value of perioperative signals but do not jointly test heterogeneous evidence integration, intervention selection, and decision updating.

\noindent\textbf{Anesthesiology Benchmarks and LLM-Based Evaluation.}
CAB, Hypnos, and AnesSuite evaluate anesthesiology knowledge or reasoning through questions, consultations, and clinical scenarios~\citep{CAB,wang2025hypnos,Anessuite}.
Other medical benchmarks separately cover multimodal perception~\citep{omnimedvqa,ecg_qa}, information gathering and EHR tools~\citep{hager2024mimiccdm,medagentbench}, or evolving records~\citep{longmedbench}.
AnesTRACE differs by separating raw-evidence perception, single-point management, and multi-step decision updating within one anesthesia-specific domain and controlled information interfaces.

For open-ended evaluation, Prometheus and JudgeLM train rubric-based evaluators~\citep{kim2024prometheus,zhu2025judgelm}, while HealthBench and LLMEval-Med use physician-defined criteria and expert-validated judgments~\citep{healthbench,zhang2025llmeval}.
AnesTRACE-Eval extends criterion-based clinical evaluation to decision-time evidence, safety, and Temporal Consistency for intraoperative management. A comparison with related benchmarks and evaluators is summarized
in Appendix~\ref{app:evaluation_metrics}.

\section{Task Formulation}
AnesTRACE separates intraoperative perception, management reasoning, and decision updating through three complementary information settings.

\subsection{Clinical State}
For decision-oriented tasks, let $t_k$ denote the decision time associated with the $k$-th clinical decision anchor, a point at which the available evidence suggests a change in patient state that may warrant clinical reassessment or management adjustment.
An anchor marks an opportunity for reassessment and does not require that a new intervention was recorded at that instant.
A recent observation window selects dynamic measurements leading up to the decision:
\begin{equation}
W_k^{\mathrm{obs}}=[t_k-\Delta_{\mathrm{obs}},t_k),
\label{eq:revised_observation_window}
\end{equation}
where $\Delta_{\mathrm{obs}}$ is the predefined observation horizon.
The patient state available at $t_k$ is represented as
\begin{equation}
S_k =
\left(
C,\,
M_{t_k},\,
O_{W_k^{\mathrm{obs}}}
\right),
\label{eq:revised_patient_state}
\end{equation}
where $C$ denotes patient and procedural context, $M_{t_k}$
denotes ongoing management, and $O_{W_k^{\mathrm{obs}}}$
contains recent measurements and trends together with events
and investigation results available by $t_k$.
Inputs and patient-information tools expose only evidence available
at the decision time; reference answers and screening labels remain
hidden. Temporal alignment and filtering are detailed in
Appendix~\ref{app:data_sources}.
We denote the initially supplied information by $I_k$, which may
contain only part of $S_k$.

\subsection{Evaluation Task}
\textbf{L1: Intraoperative Perception} maps a TEE frame or video, physiological waveform, or vital-sign trend plot and task prompt to an option, spatial localization, or structured clinical finding.
Given raw input $x$ and prompt $q$, the model produces
$\widehat{y}=f_{\theta}(x,q)$ as an option selection, spatial
localization, or structured finding job out of 12 tasks (see Appendix~\ref{app:evaluation_metrics}), without formulating a management recommendation.

\textbf{L2: Single-point Anesthesia Decision-Making} evaluates one management response $\widehat{A}_k=\pi_{\theta}(I_k,q_k)$ at the fixed decision time $T_0=t_k$ from standardized textual evidence, without additional patient-information requests.
In this expression, $q_k$ is the task prompt, $\pi_{\theta}$ is the evaluated model with parameters $\theta$, and $\widehat{A}_k$ is its assessment and management response.
Prompts cover diagnosis, short-term risk, intervention selection, and reassessment planning.

\textbf{L3: Multi-step Anesthesia Decision-Making} evaluates information acquisition and decision updating by a language-model agent with bounded tools and episode memory.
An episode $\mathcal{T}_i$ contains at least two decision points from recorded surgical case $i$:
\begin{equation}
\mathcal{T}_i=\bigl((t_{i,k},S_{i,k},A_{i,k}^{\mathrm{rec}})\bigr)_{k=1}^{K_i},\qquad K_i\geq 2.
\label{eq:revised_episode}
\end{equation}
Here $K_i$ is the number of decision points, $S_{i,k}$ is the patient state at decision time $t_{i,k}$, and $A_{i,k}^{\mathrm{rec}}$ is a documented intervention when available.
At decision point $k$, the agent starts from $I_k$ and may access history $\mathcal{H}_k$ and the permitted tool set $\mathcal{U}$ to produce retrieved information $R_k$ and a structured management response $\widehat{A}_k$:
\begin{equation}
(R_k,\widehat{A}_k)=\Pi_{\theta}(I_k,\mathcal{H}_k;\mathcal{U}),
\label{eq:revised_l3_policy}
\end{equation}
$\Pi_{\theta}$ denotes the complete agent policy, including tool selection, memory access, and response generation by the underlying model.
Tools expose only information available at $t_k$, and memory distinguishes agent proposals from recorded interventions.
After submission, trajectory replay reveals the next recorded state and intervention outcome; note that $\widehat{A}_k$ does not alter the trajectory.

\noindent\textbf{Design Rationale and Clinical Usefulness.}
The three protocols mirror distinct requirements of intraoperative care: extracting findings from heterogeneous monitoring data, converting an available patient state into an immediate management plan, and reassessing that plan after the state or documented treatment response changes.
Keeping these interfaces separate is clinically meaningful because an incorrect recommendation may originate from missed evidence, faulty management reasoning, or failure to incorporate temporal context, each of which calls for a different improvement strategy.
The design is also operationally useful: standardized L2 inputs permit controlled comparison of management reasoning, while bounded L3 tools and trajectory replay test information-seeking and decision continuity without exposing future records or claiming to simulate the physiological effects of model-proposed actions.
\section{AnesTRACE-Bench}
\label{sec:anestrace_bench}

\begin{table*}[t]
    \centering
    \caption{Three-level task hierarchy of AnesTRACE-Bench, covering intraoperative perception, single-point management, and multi-step management.}
    \label{tab:benchmark_composition}
    \begingroup
    \scriptsize
    \setlength{\tabcolsep}{4pt}
    \renewcommand{\arraystretch}{1.12}

    \begin{tabular*}{\textwidth}{@{\extracolsep{\fill}}
        c l l l l l l
    @{}}
        \toprule
        \textbf{Level} &
        \textbf{Task} &
        \textbf{Input} &
        \textbf{Question} &
        \textbf{Response} &
        \textbf{Evaluation} &
        \textbf{Scale} \\
        \midrule

        \multirow[c]{5}{*}{\textbf{L1}} &
        \multirow[c]{3}{*}{TEE perception} &
        \multirow[c]{3}{*}{TEE frames/videos} &
        Option selection & Option ID &
        \multirow[c]{3}{*}{\makecell[l]{Accuracy\\mIoU / F1}} &
        \multirow[c]{3}{*}{817 QA} \\
        & & & Grounding & Coordinates & & \\
        & & & Open-ended & Structured text & & \\
        \cmidrule(lr){2-7}

        & \multirow[c]{2}{*}{Waveform perception} &
        \multirow[c]{2}{*}{Waveforms/trends} &
        Option selection & Option ID(s) &
        \multirow[c]{2}{*}{\makecell[l]{Accuracy\\F1}} &
        \multirow[c]{2}{*}{1,000 QA} \\
        & & & Open-ended & Structured text & & \\
        \midrule

        \textbf{L2} &
        Single-point management &
        Clinical state $S_k$ &
        Open-ended &
        Structured text &
        \mbox{AnesTRACE-Eval} &
        493 anchors \\
        \midrule

        \textbf{L3} &
        Multi-step management &
        \makecell[l]{State + history\\$S_k,\,\mathcal{H}_k$} &
        Multi-turn &
        Structured text/turn &
        \mbox{AnesTRACE-Eval} &
        \makecell[l]{167 episodes\\517 turns} \\

        \bottomrule
    \end{tabular*}
    \endgroup
    \vspace{-3mm}
\end{table*}

\textbf{Overall Design.} Figure~\ref{fig:anestrace_overview} summarizes the three task settings and their controlled information interfaces.
Task-specific metrics assess L1, while AnesTRACE-Eval assesses L2 and L3 responses and adds episode-level Temporal Consistency for L3 (Section~\ref{sec:anestrace_eval}).
Controlled comparisons vary input representation, clinical history, memory, and tool access; Appendix~\ref{app:agent_framework} provides configurations and budgets.
This organization does not assume that success at one level implies success at the next; instead, it makes distinct failure sources observable within a common clinical domain.

\textbf{Data Curation.}
We draw on VitalDB~\citep{vitaldb} and MOVER~\citep{mover} for perioperative records and physiological measurements, INSPIRE~\citep{inspire} for structured perioperative records, and EchoNet-TEE~\citep{echo_net} for TEE material.
Dataset-specific parsers harmonize context, management, observations, units, and relative timestamps; records with unreliable alignment, insufficient evidence, or unresolved artifacts are excluded.
Perception questions are constructed from selected TEE and physiological inputs, while candidate clinical decision anchors are identified from clinically meaningful changes, procedural phases, and newly available results.
Expert-informed rules account for contexts such as cardiopulmonary bypass and deep hypothermic circulatory arrest, in which ordinary physiological thresholds may not indicate the same clinical state.
The initial pool contains 5,017 candidate clinical decision anchors and more than 9,000 multimodal instances.
A retained anchor supplies an L2 decision point; ordered anchors form L3 episodes with future states withheld.
L2 therefore isolates one decision from evidence available at a fixed time, whereas L3 reveals later states and documented intervention outcomes only after each agent response.
Appendix~\ref{app:data_sources} details source alignment and inclusion criteria.

\textbf{Expert Review and Quality Guarantee.}
An open-source language model assists candidate generation, after which anesthesiologists review clinical relevance, correctness, decision-time evidence, anchor validity, and evidence sufficiency.
Reference responses permit clinically acceptable alternatives and identify consequential errors.
Temporal validation removes cases whose reference response relies on information that became available only after the relevant decision time.
Appendix~\ref{app:datasets_annotation} details annotation and temporal validation.

After expert review, L1 contains 817 TEE and 1,000 waveform or trend question--answer instances; L2 contains 493 clinical decision anchors from 467 surgical cases; and L3 contains 167 episodes comprising 517 turns from 124 surgical cases.
These units are reported separately because questions, anchors, episodes, and turns are not interchangeable; Appendix~\ref{app:task_specifications} provides task definitions, formats, counts, and metrics.
The retrospective design and selected episodes limit extrapolation to unrepresented settings and to outcomes from executing model recommendations.
\section{AnesTRACE-Eval}
\label{sec:anestrace_eval}

\begin{table}[t]
\centering
\small
\caption{Evaluation dimensions, values, and applicable units in AnesTRACE-Eval.}
\label{tab:eval_dimension_summary}
\setlength{\tabcolsep}{4pt}
\renewcommand{\arraystretch}{1.15}
\begin{tabularx}{\linewidth}{@{}>{\raggedright\arraybackslash}p{0.20\linewidth}>{\raggedright\arraybackslash}X>{\raggedright\arraybackslash}p{0.14\linewidth}>{\raggedright\arraybackslash}p{0.23\linewidth}@{}}
\toprule
\textbf{Dimension} & \textbf{Assessment} & \textbf{Values} & \textbf{Evaluation unit} \\
\midrule
Clinical Correctness & Clinical conclusions and management & 0, 1, 2 & L2 response; L3 decision turn \\
Evidence Grounding & Support from available evidence & 0, 1, 2 & L2 response; L3 decision turn \\
Task Completeness & Coverage of requested content & 0, 1, 2 & L2 response; L3 decision turn \\
Temporal Consistency & Decision updating as evidence evolves & 0, 1, 2 & Complete L3 episode \\
Safety & Unsafe actions, consequential omissions, and inappropriate clinical delays & \textit{safe}, \textit{minor}, \textit{major}, \textit{critical} & L2 intervention and reassessment outputs; L3 intervention decisions \\
\bottomrule
\end{tabularx}
\par\smallskip
\begin{minipage}{\linewidth}
\end{minipage}
\vspace{-3mm}
\end{table}


AnesTRACE-Eval combines an anesthesiologist-defined clinical protocol with a domain-specific evaluator for L2 and L3 responses, extending criterion-based medical evaluation with decision-time evidence and temporal assessment~\citep{healthbench,medhelm,zhang2025llmeval}.

\noindent\textbf{Criterion-Based Clinical Evaluation.} As shown in Figure~\ref{fig:anestrace_overview}, AnesTRACE-Eval combines a candidate response with three groups of evaluation evidence: patient context, a reference answer, and medical evidence.
For L2, patient context contains the information available at the decision point; for L3, it additionally contains the episode history of patient states, agent recommendations, and documented interventions.
Patient context establishes the case facts, the reference answer describes an expert-supported response without defining the only acceptable plan, and authoritative anesthesiology, ASA, and pharmacological sources support clinical interpretation~\citep{miller_anesthesia_2024,asa2025monitoring,knox2024drugbank}.
Atomic criteria specify the clinical requirements applicable to a case without forcing one reference wording or one acceptable management plan.
The evaluator jointly assigns dimension-level judgments because individual criteria provide evidence for the response-level clinical assessment rather than independent benchmark scores.

Given input $x$, evaluator $J_{\phi}$ jointly produces applicable dimension-level judgments and rationales:
\begin{equation}
J_{\phi}(x)=\bigl\{(s_d,r_d)\bigr\}_{d\in\mathcal{D}(x)},
\label{eq:eval_dimensions}
\end{equation}
Here $\mathcal{D}(x)$ contains the applicable dimensions and $(s_d,r_d)$ pairs each judgment with its rationale (Table~\ref{tab:eval_dimension_summary}).

For L3, Temporal Consistency is assessed once per episode by whether recommendations are maintained or revised appropriately as evidence evolves.
This episode-level judgment complements turn-level Clinical Correctness, Evidence Grounding, and Task Completeness by testing whether later decisions incorporate newly revealed states and documented intervention outcomes.

Ordinal scores are averaged over their response-, turn-, or episode-level units.
Safety is reported separately for applicable L2 and L3 outputs as the Major/Critical Safety Error Rate:
\begin{equation}
\mathrm{Safety\ Error}
=\frac{1}{N}\sum_{k=1}^{N}
\mathbb{I}\!\left[s_{\mathrm{saf},k}\in
\{\textit{major},\textit{critical}\}\right],
\label{eq:eval_safety_rate}
\end{equation}
Here $N$ counts applicable outputs with valid safety labels.
Safety remains separate from average clinical scores, and Appendix~\ref{app:evaluation_metrics} provides the complete rubrics and aggregation rules.
Separating safety prevents a low-frequency but consequential unsafe action or omission from disappearing inside an average of more common clinical judgments.

\noindent\textbf{Anesthesiologist-Aligned Evaluator Training.} 
We train the domain-specific evaluator as a companion to AnesTRACE-Bench using supervision derived from benchmark cases.
DeepSeek-V4-Pro~\citep{xu2026deepseek} generates response variants, judgments, rationales, and preference pairs, which anesthesiologists correct and rank.
Starting from Qwen3.5-9B, supervised fine-tuning uses approximately 7,000 expert-reviewed examples, followed by Direct Preference Optimization on approximately 4,000 expert-ranked pairs~\citep{dpo}.
Each response-level SFT example contains the judgments and rationales for all applicable dimensions, while a trajectory-level example is counted once per episode rather than once per turn or criterion.
Appendix~\ref{app:evaluator_training} provides annotation and optimization details.

Because evaluator-training cases overlap with benchmark cases by design, validation measures agreement with anesthesiologists on benchmark responses and does not establish generalization to unseen clinical cases.
Section~\ref{sec:experiments} compares base, SFT-only, and preference-aligned variants on 100 L2 responses and 117 L3 decision-turn responses reviewed by three anesthesiologists.
\section{Experiments}
\label{sec:experiments}

\begin{table*}[t]
\centering
\renewcommand{\arraystretch}{1.25}
\setlength{\tabcolsep}{5pt}
\caption{
Performance of variant MLLMs on the intraoperative perception tasks of AnesTRACE-Bench.
Bold indicates the best result in each column.
Background colors denote model size:
{\setlength{\fboxsep}{0.8pt}\colorbox{ModelSmallBase!12}{\small\strut Small}} ($<10$B),
{\setlength{\fboxsep}{0.8pt}\colorbox{ModelMediumBase!12}{\small\strut Medium}} ($10$--$40$B), and
{\setlength{\fboxsep}{0.8pt}\colorbox{ModelLargeBase!14}{\small\strut Large}} ($>40$B).
}
\label{tab:model_comparison}
\resizebox{\textwidth}{!}{%
\begin{tabular}{llcccccccc}
\toprule
& \multirow{3}{*}{\textbf{Model}}
& \multicolumn{5}{c}{\textbf{TEE}}
& \multicolumn{3}{c}{\textbf{Waveform}} \\
\cmidrule(lr){3-7}\cmidrule(lr){8-10}
& & \makecell{Foundational\\Perception}
& \multicolumn{2}{c}{\makecell{Visual\\Grounding}}
& \multicolumn{2}{c}{\makecell{Functional\\Assessment}}
& \makecell{Value\\Extraction}
& \makecell{Trend\\Reasoning}
& \makecell{Anomaly\\Perception} \\
\cmidrule(lr){3-3}\cmidrule(lr){4-5}\cmidrule(lr){6-7}
\cmidrule(lr){8-8}\cmidrule(lr){9-9}\cmidrule(lr){10-10}
& & ACC & ACC & mIoU & ACC & Factual-F1 & ACC & Set-F1 & Metric-F1 \\
\midrule
\multirow{3}{*}{\rotatebox[origin=c]{90}{\small\makecell{\textbf{Proprietary}\\\textbf{MLLMs}}}}
& \modellarge{GPT-6-Astra}
& 73.9 & \textbf{62.9} & \textbf{32.2} & 40.6 & 27.5
& \textbf{91.5} & 61.1 & \textbf{64.8} \\
& \modellarge{Gemini 3 Pro Preview}
& 71.6 & 46.1 & 24.6 & 34.4 & 23.7 & 77.0 & 62.6 & 54.8 \\
& \modellarge{Claude-Fable-5}
& \textbf{74.5} & 42.7 & 23.4 & 37.8 & 26.9 & 80.5 & 68.7 & 59.6 \\
\midrule
\multirow{7}{*}{\rotatebox[origin=c]{90}{\small\makecell{\textbf{General}\\\textbf{Open-Weight MLLMs}}}}
& \modellarge{Qwen3.8-MAX}
& 73.2 & 51.7 & 22.8 & \textbf{48.2} & \textbf{29.9} & 72.0 & 69.6 & 42.8 \\
& \modelmedium{Qwen3.8-27B}
& 58.5 & 33.7 & 14.8 & 44.5 & 14.0
& 57.5 & \textbf{72.5} & 31.3 \\
& \modelmedium{Qwen3.5-27B}
& 58.5 & 43.8 & 18.9 & 43.2 & 16.5 & 56.0 & 65.3 & 36.5 \\
& \modelsmall{Qwen3.5-9B}
& 54.2 & 52.8 & 15.5 & 42.0 & 8.7 & 54.5 & 60.7 & 33.1 \\
& \modelsmall{Qwen3-VL-8B}
& 51.5 & 23.6 & 14.4 & 35.6 & 5.7 & 45.0 & 58.2 & 23.5 \\
& \modelsmall{LLaVA-OneVision-2-8B}
& 49.9 & 50.6 & 14.3 & 37.1 & 9.7 & 49.5 & 60.6 & 23.6 \\
& \modelsmall{InternVL3.5-8B}
& 54.0 & 48.3 & 13.9 & 34.8 & 6.3 & 41.0 & 70.1 & 17.3 \\
\midrule
\multirow{4}{*}{\rotatebox[origin=c]{90}{\scriptsize\makecell{\textbf{Medical-Domain}\\\textbf{Open-Weight}\\\textbf{MLLMs}}}}
& \modelmedium{Fleming-VL-38B}
& 50.3 & 37.1 & 14.8 & 28.2 & 11.2 & 41.5 & 59.4 & 24.0 \\
& \modelmedium{Lingshu-32B}
& 55.0 & 46.1 & 12.1 & 35.2 & 17.6 & 49.5 & 64.6 & 20.0 \\
& \modelsmall{Lingshu-I-8B}
& 53.5 & 56.2 & 13.5 & 33.4 & 9.2 & 39.0 & 59.5 & 23.3 \\
& \modelsmall{Fleming-VL-8B}
& 47.2 & 46.1 & 8.7 & 12.9 & 7.3 & 38.5 & 41.1 & 17.6 \\
\bottomrule
\end{tabular}%
}
\end{table*}

\begin{table*}[t]
\centering
\caption{Performance of variant LLMs on the decision-making tasks of AnesTRACE-Bench. Higher is better except for Safety Error and latency; bold indicates the best result in each column. Background colors denote model size:
{\setlength{\fboxsep}{0.8pt}\colorbox{ModelSmallBase!12}{\small\strut Small}} ($<10$B),
{\setlength{\fboxsep}{0.8pt}\colorbox{ModelMediumBase!12}{\small\strut Medium}} ($10$--$40$B), and
{\setlength{\fboxsep}{0.8pt}\colorbox{ModelLargeBase!14}{\small\strut Large}} ($>40$B).
}
\renewcommand{\arraystretch}{1.25}
\setlength{\tabcolsep}{4.5pt}
\resizebox{\textwidth}{!}{%
\begin{tabular}{ll*{12}{c}}
\toprule
& \multirow{2}{*}{\textbf{Model}}
& \multicolumn{7}{c}{\textbf{Single-Point Decision Making}}
& \multicolumn{5}{c}{\textbf{Multi-Step Decision Making}} \\
\cmidrule(lr){3-9}
\cmidrule(lr){10-14}
& & \makecell{\textbf{Risk}$\uparrow$}
& \makecell{\textbf{Diagnosis}$\uparrow$}
& \makecell{\textbf{Intervention}$\uparrow$}
& \makecell{\textbf{Reassessment}$\uparrow$}
& \makecell{\textbf{Avg.}$\uparrow$}
& \makecell{\textbf{Safety}\\\textbf{Error}$\downarrow$}
& \makecell{\textbf{P95}\\\textbf{Latency (s)}$\downarrow$}
& \makecell{\textbf{Turn-level}\\\textbf{Avg.}$\uparrow$}
& \makecell{\textbf{Temporal}\\\textbf{Consistency}$\uparrow$}
& \makecell{\textbf{Safety}\\\textbf{Error}$\downarrow$}
& \makecell{\textbf{Evidence}\\\textbf{Acquisition}$\uparrow$}
& \makecell{\textbf{P95}\\\textbf{Latency (s)}$\downarrow$} \\
\midrule
\multirow{3}{*}{\rotatebox[origin=c]{90}{\small\makecell{\textbf{Proprietary}\\\textbf{LLMs}}}}
& \modellarge{GPT-6-Astra}
& 80.8 & \textbf{93.7} & \textbf{96.0} & \textbf{98.4}
& \textbf{92.2} & \textbf{3.4} & 33.41
& \textbf{82.7} & \textbf{87.6} & \textbf{17.5} & 61.8 & 74.3 \\
& \modellarge{Gemini 3 Pro Preview}
& 71.1 & 74.7 & 68.9 & 86.1 & 75.2 & 13.5 & 26.95
& 66.1 & 78.0 & 38.5 & 74.4 & 101.4 \\
& \modellarge{Claude-Fable-5}
& \textbf{87.1} & 91.9 & 87.8 & 96.3 & 90.8 & 9.4 & 39.96
& 79.4 & 82.5 & 24.3 & 84.4 & 69.2 \\
\midrule
\multirow{8}{*}{\rotatebox[origin=c]{90}{\small\makecell{\textbf{General}\\\textbf{Open-Weight LLMs}}}}
& \modellarge{DeepSeek-V4-Pro}
& 78.2 & 83.5 & 79.0 & 93.8 & 83.6 & 14.8 & 14.04
& 80.0 & 81.9 & 19.2 & 89.7 & 14.4 \\
& \modellarge{Qwen3.8-Max}
& 70.7 & 79.6 & 71.6 & 87.5 & 77.4 & 15.1 & 23.51
& 67.1 & 85.3 & 40.1 & 62.8 & 22.9 \\
& \modelmedium{Qwen3.5-27B}
& 69.4 & 74.6 & 70.2 & 77.9 & 73.0 & 28.6 & 22.14
& 69.5 & 81.6 & 32.5 & 83.3 & 34.4 \\
& \modelmedium{Qwen3.8-27B}
& 71.4 & 75.5 & 66.1 & 75.0 & 72.0 & 21.5 & 22.94
& 66.1 & 69.2 & 34.1 & 86.9 & 39.9 \\
& \modelmedium{GPT-OSS-20B}
& 69.1 & 65.8 & 50.7 & 66.1 & 62.9 & 37.5 & 17.90
& 57.1 & 67.2 & 53.4 & 59.5 & 106.6 \\
& \modelmedium{GLM-4.7-Flash}
& 58.5 & 58.6 & 46.1 & 59.7 & 55.7 & 48.2 & 36.62
& 44.1 & 49.9 & 70.8 & 95.4 & 63.6 \\
& \modelsmall{Qwen3.5-9B}
& 65.3 & 70.0 & 57.0 & 71.9 & 66.1 & 27.7 & 15.18
& 55.1 & 64.2 & 61.9 & 89.5 & 27.2 \\
& \modelsmall{Qwen3-8B}
& 58.2 & 52.2 & 37.6 & 50.8 & 49.7 & 57.6 & 24.50
& 30.2 & 45.2 & 64.1 & 32.7 & \textbf{10.3} \\
\midrule
\multirow{8}{*}{\rotatebox[origin=c]{90}{\small\makecell{\textbf{Medical-Domain}\\\textbf{Open-Weight LLMs}}}}
& \modelmedium{Morpheus-32B}
& 55.0 & 50.4 & 41.7 & 58.0 & 51.3 & 56.9 & 20.31
& 46.2 & 56.0 & 66.0 & 67.5 & 86.9 \\
& \modelmedium{Fleming-R1-32B}
& 63.9 & 63.9 & 46.4 & 67.0 & 60.3 & 37.6 & 14.55
& 52.8 & 66.9 & 56.5 & 11.3 & 22.8 \\
& \modelmedium{HuatuoGPT-3-32B}
& 62.7 & 60.7 & 47.6 & 59.7 & 57.7 & 50.9 & 25.05
& 49.9 & 61.7 & 64.7 & 73.4 & 31.3 \\
& \modelmedium{Baichuan-M2-32B}
& 57.9 & 62.0 & 48.2 & 51.5 & 54.9 & 42.1 & 75.10
& 49.5 & 56.1 & 66.7 & \textbf{96.0} & 33.0 \\
& \modelmedium{MedGemma-27B-text-it}
& 64.7 & 61.2 & 41.5 & 63.3 & 57.7 & 43.7 & 12.44
& 48.7 & 54.3 & 59.4 & 40.5 & 34.4 \\
& \modelsmall{HuatuoGPT-3-8B}
& 44.3 & 38.8 & 35.6 & 45.5 & 41.1 & 85.9 & 9.87
& 33.8 & 48.2 & 86.6 & 29.1 & 99.9 \\
& \modelsmall{Morpheus-7B}
& 47.0 & 42.3 & 39.0 & 50.4 & 44.7 & 67.4 & 26.80
& 35.1 & 46.7 & 78.2 & 8.3 & 50.4 \\
& \modelsmall{Fleming-R1-7B}
& 55.0 & 52.5 & 40.5 & 55.2 & 50.8 & 59.0 & \textbf{9.19}
& 44.5 & 56.6 & 69.4 & 4.6 & 33.1 \\
\bottomrule
\end{tabular}%
}
\label{tab:l2_l3_results}
\vspace{-3mm}
\end{table*}

\subsection{Experimental Setup}
\label{sec:experimental_setup}

\noindent\textbf{Baselines and Evaluation Protocols.}
We evaluate more than 30 proprietary, general-purpose open-weight, and medical-domain models~\citep{bai2025qwen3,wang2025internvl3,li2024llava,qwen35,yang2025qwen3,zeng2025glm,guo2025deepseek,agarwal2025gpt,xu2026lingshu,liu2025fleming,zhang2023huatuogpt,dou2025baichuan,sellergren2025medgemma,Anessuite}.
Note that we vary model size because real-world hospitals face finite local compute~\citep{de2014effect,wijnberge2020effect}, and report latency because high-quality responses may still be unusable outside the clinical response window\footnote{We set it 30 seconds indicated by our anesthesiologist collaborator.}.
L1 uses raw visual inputs, L2 uses standardized textual evidence, and L3 uses a shared agent framework with episode memory and bounded tools; Appendix~\ref{app:experiments} provides complete settings.
For open-ended L1 responses, Factual F1 and Metric-F1 match extracted clinical triplets against anesthesiologist-annotated references; L2 and L3 use AnesTRACE-Eval presented in Section~\ref{sec:anestrace_eval}.

\subsection{Main Results on AnesTRACE}
\label{sec:main_results}

\textbf{From recognition to clinically usable perception.} Table~\ref{tab:model_comparison} shows that strong basic recognition does not extend uniformly to fine-grained intraoperative interpretation.
GPT-6-Astra reaches 32.2 mIoU for TEE grounding and 27.5 Factual-F1 for functional assessment, while the best waveform anomaly Metric-F1 is 64.8, indicating that \textit{fine-grained evidence extraction is one of the primary perception bottlenecks}.
For clinical use, the result means that model-derived findings should remain linked to localized source evidence before they inform intraoperative management.

\textbf{From patient state to safe sequential management.} Table~\ref{tab:l2_l3_results} shows that intervention selection trails overall L2 performance for most models, including GPT-OSS-20B (50.7 versus 62.9) and MedGemma-27B-text-it (41.5 versus 57.7).
Standardized evidence therefore reduces variation in perception but \textit{does not resolve the challenge of converting the patient state and ongoing management into an appropriate action}.
In L3, GPT-6-Astra leads Turn-level Avg., Temporal Consistency, and Safety, whereas medical-domain open-weight models show no consistent advantage: Fleming-R1-32B reaches 52.8 in Turn-level Avg. versus 69.5 for Qwen3.5-27B.
Safety errors increase from 3.4\% to 17.5\% for GPT-6-Astra and from 9.4\% to 24.3\% for Claude-Fable-5 between L2 and L3, showing that \textit{strong individual decisions do not ensure safe management across a trajectory}.

\textbf{Technical and clinical implications.} Across the three settings, the principal technical bottleneck shifts from fine-grained evidence extraction to intervention selection and then to safe decision updating.
These shifts indicate that \textit{reliable intraoperative support requires joint gains in perception, reasoning, safety, and efficiency}.
For clinical use, the results indicate that current models should support anesthesiologists by presenting traceable evidence and updated recommendations under human oversight; aggregate quality alone is insufficient for entrusting AI with independent intraoperative management.
Appendix~\ref{app:performance_analyses} provides full dimension-wise, subgroup, efficiency, and error analyses.

\begin{figure}[t]
  \centering
  \begin{minipage}[t]{0.48\textwidth}
    \vspace{0pt}
    \centering
    \includegraphics[width=\linewidth]{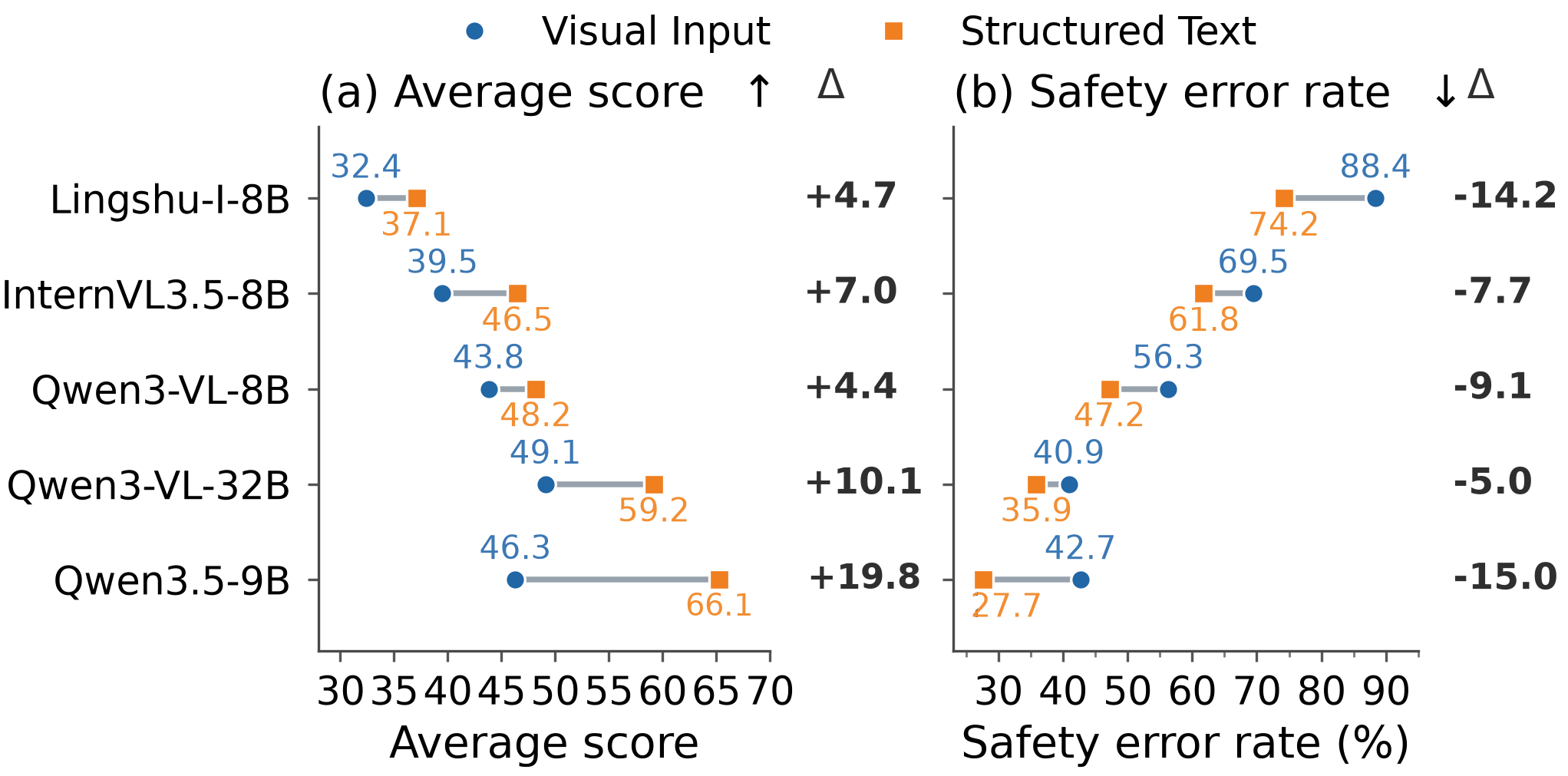}
    \caption{Effect of input representation ablation for five MLLMs on L2 tasks. $\Delta$ denotes structured text minus visual input.}
    \label{fig:perception_impact}
  \end{minipage}\hfill
  \begin{minipage}[t]{0.49\textwidth}
    \vspace{0pt}
    \captionof{table}{Comparison of AnesTRACE-Eval variants in agreement with anesthesiologist ratings. Each cell reports L2/L3. Spear. and Kend. denote Spearman's $\rho$ and Kendall's $\tau_b$, respectively.}
    \label{tab:evaluator_human_agreement}
    \centering
    \scriptsize
    \setlength{\tabcolsep}{1pt}
    \renewcommand{\arraystretch}{1.12}
    \begin{tabular*}{\linewidth}{@{\extracolsep{\fill}}lcccccc@{}}
      \toprule
      \multirow{2}{*}{\textbf{Dimension}}
      & \multicolumn{2}{c}{\textbf{Base}}
      & \multicolumn{2}{c}{\textbf{SFT}}
      & \multicolumn{2}{c}{\textbf{SFT + DPO}} \\
      \cmidrule(lr){2-3}\cmidrule(lr){4-5}\cmidrule(lr){6-7}
      & \textbf{Spear.} & \textbf{Kend.} 
      & \textbf{Spear.} & \textbf{Kend.}
      & \textbf{Spear.} & \textbf{Kend.} \\
        \midrule
        \textbf{Correct.}  & .46/.48 & .43/.44 & .57/.68 & .53/.63 & \textbf{.60/.71} & \textbf{.56/.66} \\
        \textbf{Grounding} & .44/.35 & .41/.32 & .48/.66 & .45/.62 & \textbf{.52/.69} & \textbf{.49/.65} \\
        \textbf{Complete.} & .39/.45 & .32/.38 & .50/.46 & .47/.42 & \textbf{.59/.68} & \textbf{.54/.63} \\
        \textbf{Total}     & .49/.47 & .42/.39 & .53/.72 & .45/.62 & \textbf{.58/.75} & \textbf{.50/.65} \\
        \textbf{Safety}    & .38/.47 & .34/.41 & .48/.77 & .43/.70 & \textbf{.51/.80} & \textbf{.47/.73} \\
        \textbf{Temporal}  & --/.37  & --/.34  & --/.55  & --/.52  & --/\textbf{.58}  & --/\textbf{.55}   \\
        \bottomrule
    \end{tabular*}
  \end{minipage}
  \vspace{-3mm}
\end{figure}

\begin{figure}[t]
  \centering
  \begin{minipage}[t]{0.48\textwidth}
    \vspace{0pt}
    \centering
    \includegraphics[width=\linewidth]{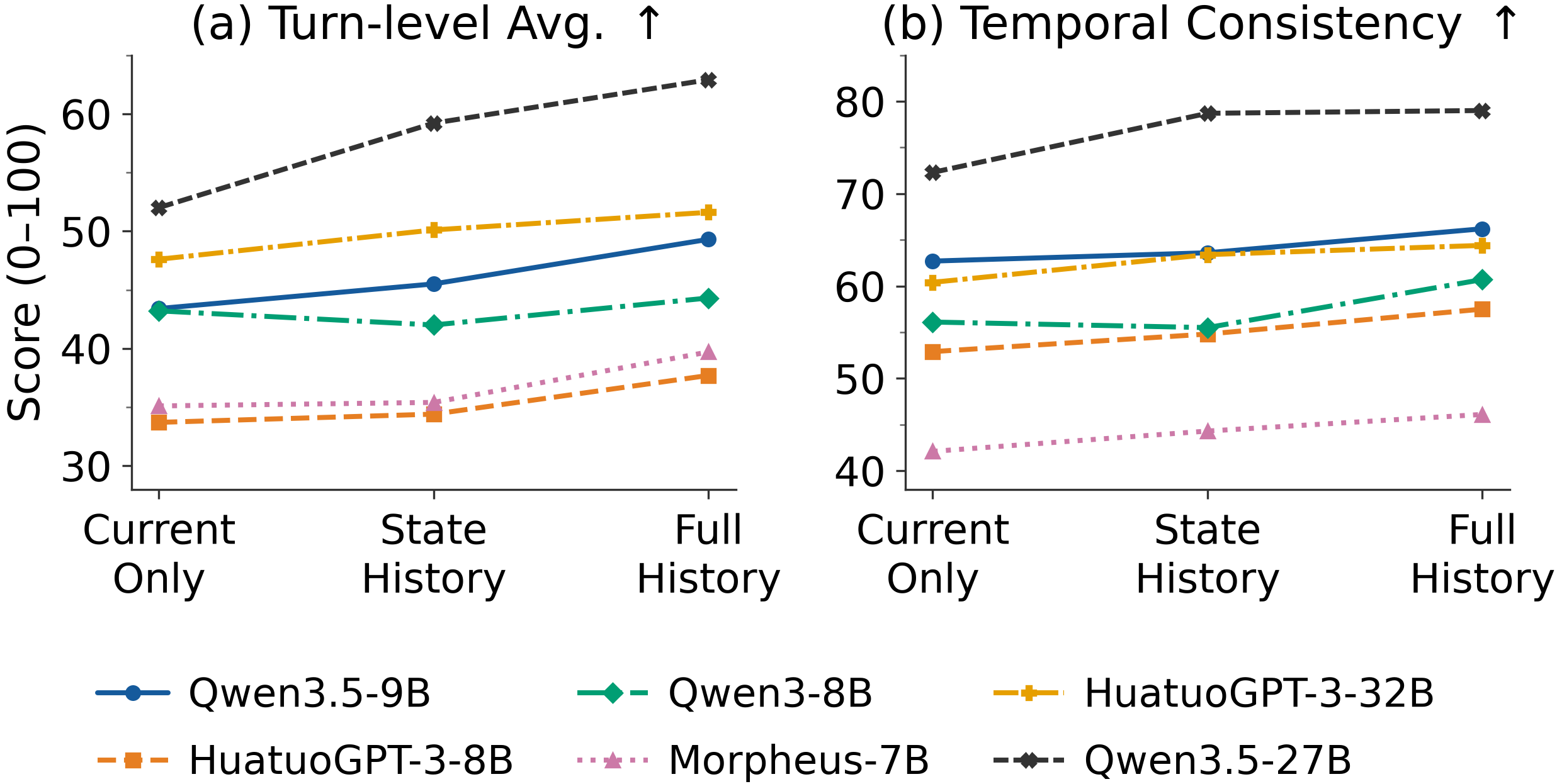}
    \caption{Clinical-history ablation.
    Avg. averages correctness, grounding, and completeness.}
    \label{fig:history_ablation_example}
  \end{minipage}\hfill
  \begin{minipage}[t]{0.49\textwidth}
    \vspace{0pt}
    \captionof{table}{Effect of agent components on multi-step anesthesia decision-making with Qwen3.5-27B.}
    \label{tab:agent_component_ablation}
    \centering
    \scriptsize
    \setlength{\tabcolsep}{2pt}
    \renewcommand{\arraystretch}{1.18}
    \begin{tabular*}{\linewidth}{@{\extracolsep{\fill}}lccc@{}}
      \toprule
      \textbf{Method} & \textbf{Avg.}$\uparrow$
      & \textbf{Temporal Consistency}$\uparrow$ & \textbf{Safety Error}$\downarrow$ \\
      \midrule
      Full Agent & 69.1 & 81.6 & 32.5 \\
      \midrule
      w/o Memory & 63.8 & 67.5 & 35.3 \\
      w/o Knowledge Tools & 63.2 & 80.5 & 43.8 \\
      w/o Clinical Requests & 48.7 & 79.2 & 54.3 \\
      w/o All Tools & 37.9 & 64.8 & 57.4 \\
      \bottomrule
    \end{tabular*}
  \end{minipage}
  \vspace{-3mm}
\end{figure}

\subsection{Ablation and Clinical Alignment}

\textbf{Impact of Input Representation on Decision-Making.}
Across five MLLMs, replacing visual waveform or trend inputs with matched structured text improves L2 Avg. and reduces Safety Error in Figure~\ref{fig:perception_impact}; 
Qwen3.5-9B changes from 46.3 to 66.1 and from 42.7\% to 27.7\%, respectively.
Because the paired inputs hold the clinical case fixed, the improvement isolates the burden of visual evidence extraction; the remaining 27.7--74.2\% Safety Error shows that \textit{structured inputs reduce the perception burden but do not resolve unsafe management reasoning}.
In practice, structured summaries may improve decision support when raw-signal interpretation is unreliable, but clinically useful systems must preserve source evidence and jointly validate both perception and the resulting intervention.

\textbf{Ablation on Clinical History, Tools, and Episode Memory.}
Full History yields the highest Turn-level Avg. and Temporal Consistency across all six models (Figure~\ref{fig:history_ablation_example}); for Qwen3.5-27B, these rise from 52.0 to 62.9 and from 72.3 to 79.0 over Current Only.
In the separate agent-component configuration of Table~\ref{tab:agent_component_ablation}, removing memory lowers Temporal Consistency from 81.6 to 67.5, removing clinical requests lowers Turn-level Avg. from 69.1 to 48.7, and removing knowledge tools raises Safety Error from 32.5\% to 43.8\%.
Together, these results show that \textit{history, memory, and tool access make distinct contributions to decision updating}.
Clinically, an intraoperative assistant must retain prior actions and responses, request missing patient information, and retrieve relevant knowledge so that each recommendation remains connected to the evolving case.

\begin{wrapfigure}[14]{l}{0.49\textwidth}
  \centering
  \vspace{-0.4\baselineskip}
  \includegraphics[width=\linewidth]{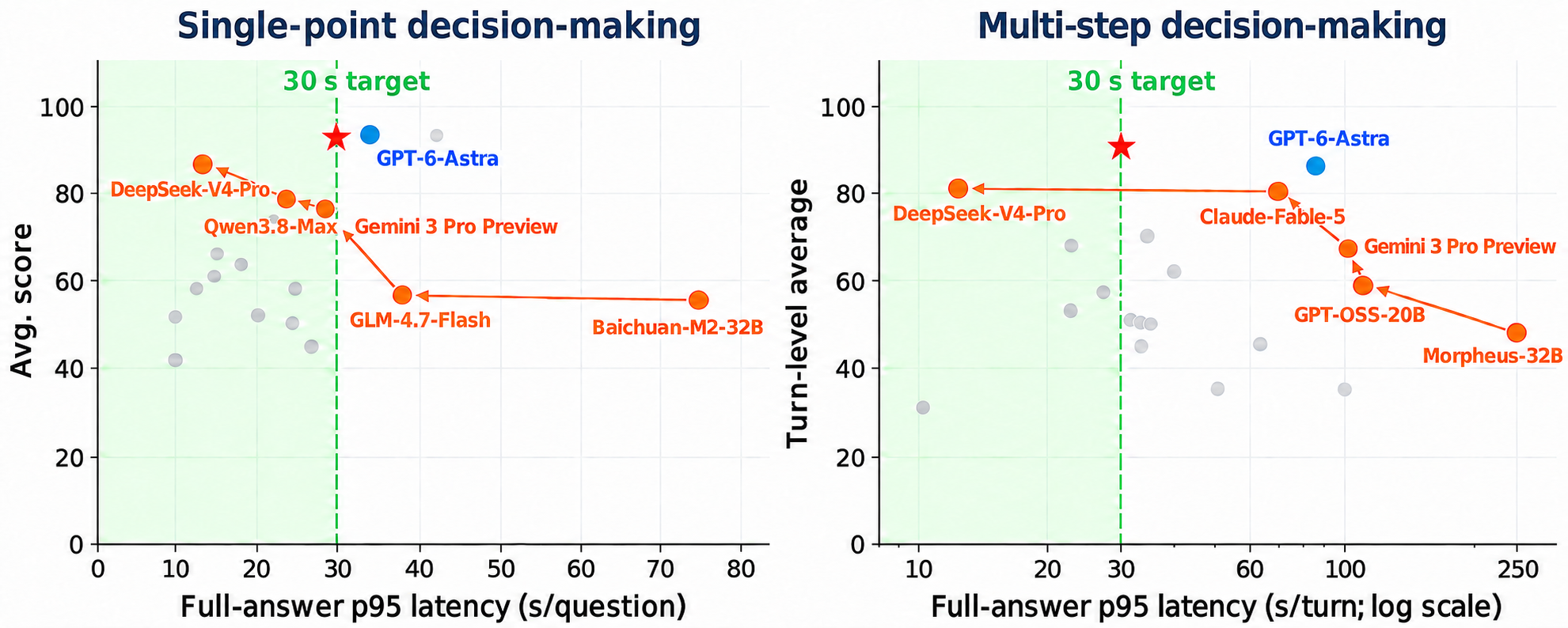}
  \caption{Quality--latency landscape for decision-making tasks. Dots represent evaluated models, orange arrows indicate ideal improvement scenarios, and red stars mark hypothetical operating points at the 30-second target.}
  \label{fig:quality_latency_illustration}
  \vspace{-0.5\baselineskip}
\end{wrapfigure}

\textbf{Alignment of AnesTRACE-Eval with Anesthesiologists.}
Three anesthesiologists reviewed 100 L2 responses and 117 L3 decision-turn responses from 39 episodes; from the base model to SFT+DPO, total-score Spearman correlation rises from .49 to .58 in L2 and from .47 to .75 in L3, while L3 Safety Error correlation rises from .47 to .80 (Table~\ref{tab:evaluator_human_agreement}).
Together, these gains show that \textit{task-specific SFT and DPO improve alignment with anesthesiologists across all evaluated dimensions}.
Clinically, this agreement supports scalable screening and comparison of model responses, while the remaining disagreement requires anesthesiologist review for safety-critical interpretation; Appendix~\ref{app:evaluator_validation} details annotation and aggregation.

\textbf{Decision Quality and Inference Efficiency.}
Figure~\ref{fig:quality_latency_illustration} shows that GPT-6-Astra attains an L3 Turn-level Avg. of 82.7 at 74.3 seconds, whereas DeepSeek-V4-Pro reaches 80.0 at 14.4 seconds; the red star marks the hypothetical combination of the best observed quality and the 30-second operating-room deployment requirement.
This comparison shows that \textit{inference efficiency is an independent requirement for clinical deployment}.
For time-sensitive intraoperative deterioration, systems must return useful evidence and recommendations within the clinical response window~\citep{de2014effect,wijnberge2020effect}; meeting that window still does not establish safety and therefore cannot replace prospective clinical validation and anesthesiologist oversight.

\subsection{Hospital-oriented model selection}

Applying the preceding production criteria changes how the results should be used: externally hosted models, such as GPT-6-Astra and Claude-Fable-5, define a capability reference, whereas local candidates must jointly satisfy quality, safety, latency, and resource constraints.

Under these constraints, Qwen3.5-9B is the most practical starting point among the evaluated locally deployed general-purpose models: its 9B scale and 27.2-second L3 latency accompany a 55.1 Turn-level Avg. and 89.5 Evidence Acquisition.
AnesTRACE supports this choice by mapping the intended role to L1--L3, applying deployment gates, and using controlled comparisons to decide whether perception, episode memory, patient-information requests, or knowledge retrieval should be added to the agent.
However, its 61.9\% L3 Safety Error and moderate decision quality show that this choice still remains \textit{far from patient-care readiness}, and \textit{no evaluated local model simultaneously satisfies} the desired quality, safety, latency, and resource profile.
Nonetheless, the benchmark defines both a current selection frontier and a measurable development path toward efficient, clinically grounded agents with abstention, escalation, and prospective validation under anesthesiologist oversight.

\section{Conclusion}

To address the lack of a shared testbed that connects multimodal evidence interpretation, intervention selection, and longitudinal decision-making, we introduce AnesTRACE, comprising AnesTRACE-Bench and AnesTRACE-Eval, in this paper.
AnesTRACE-Bench evaluates Intraoperative Perception, Single-point Anesthesia Decision-Making, and Multi-step Anesthesia Decision-Making, while AnesTRACE-Eval provides anesthesiologist-aligned assessment of open-ended responses across grounding, temporal consistency, and safety.
Across more than 30 models, results expose persistent gaps in fine-grained perception and intervention selection; even the leading model retains a 17.5\% Safety Error and requires 74.3 seconds.
AnesTRACE therefore provides the field with a clinically structured diagnostic framework for identifying where current systems fail and for guiding the development of evidence-grounded, efficient, and clinician-supervised intraoperative support.
Future work could separate evaluator training from external validation, extend coverage of rare and high-risk scenarios, and prospectively study workflow integration without treating retrospective benchmark performance as evidence of readiness for patient care.

\section*{Acknowledgments}
We thank Xiaoxia Zhou, Junyu Zhou, Jingcheng Zou, Han Cao, Wenjiao Wu, Jiajun Chen, Jiayao Zhang, Feifei Hu, Yanda Cen, and Tingting Wang from the Department of Anesthesiology, The Second Affiliated Hospital of Zhejiang University School of Medicine, for their contributions to clinical annotation and review.

\section*{AI Use Statement}
In this work, we used generative AI tools to assist benchmark construction,
generate candidate responses for expert annotation, and support the training
and application of AnesTRACE-Eval. LLMs also served as evaluated systems.
Additionally, we used generative AI tools for proofreading, sentence
refinement, and improving linguistic clarity. These constitute all uses
of generative AI in this work. AI-assisted manuscript revisions were
reviewed and edited by the authors, and model-assisted annotation materials
were reviewed by anesthesiologists. Details of model use and annotation
procedures are provided in the corresponding sections of the main text
and appendix. We take responsibility for the final content of this work,
including text, claims, and artifacts produced with the aid of generative AI.

\section*{Ethics Statement}

AnesTRACE is constructed from retrospective, de-identified records
obtained from the VitalDB, MOVER, INSPIRE, and EchoNet-TEE research
datasets. We collected no new patient data and had no interaction with
patients. The benchmark retains only de-identified source identifiers
and relative timestamps. We respected source access restrictions,
made no attempt to re-identify individuals, and do not redistribute
restricted raw records, cross-dataset linkage tables, or licensed
reference materials. All experiments were conducted retrospectively,
and no model output informed patient care. AnesTRACE is intended for
research evaluation and does not establish clinical readiness:
model responses may contain unsafe recommendations or consequential
omissions. Any downstream use must respect patient privacy and applicable dataset
licenses; clinical use would require appropriate validation and
qualified human oversight.

\section*{Reproducibility Statement}

We provide anonymous code for decision-anchor construction, preprocessing
of TEE videos, physiological waveforms, and vital-sign trends, benchmark
inference and evaluation, and AnesTRACE-Eval training. The main text and
appendix describe data curation and annotation, task definitions,
evaluation metrics, model configurations, and evaluator training.
Upon acceptance, we will release all AnesTRACE-Bench datasets and
AnesTRACE-Eval checkpoints, subject to applicable data licenses and
access requirements. Access to restricted source data must be obtained
from the respective providers. The code is available at
\url{https://anonymous.4open.science/r/AnesTRACE-027C/}.

\bibliography{iclr2027_conference}
\bibliographystyle{iclr2027_conference}

\appendix
\clearpage
\onecolumn

\section*{\centering Appendix}
\label{app:overview}

\begingroup
\setlength{\parindent}{0pt}
\setlength{\parskip}{0pt}
\newcommand{\appsectionentry}[3]{%
  \noindent%
  \textbf{\hyperref[#1]{#2. #3}}%
  \nobreak\dotfill\nobreak\pageref{#1}\par
  \vspace{0.80em}%
}
\newcommand{\appsubsectionentry}[3]{%
  \noindent\hspace*{1.8em}%
  {\small\itshape\hyperref[#1]{#2. #3}}%
  \nobreak\dotfill\nobreak{\small\pageref{#1}}\par
  \vspace{0.50em}%
}

\appsectionentry{app:dataset_construction}{A}{Dataset Construction and Annotation}
\appsubsectionentry{app:data_sources}{A.1}{Data Sources and Harmonization}
\appsubsectionentry{app:clinical_terminology}{A.2}{Clinical Terminology}
\appsubsectionentry{app:task_construction}{A.3}{Decision Anchors and Episodes}
\appsubsectionentry{app:datasets_annotation}{A.4}{Expert Annotation and Dataset Statistics}

\appsectionentry{app:evaluation_metrics}{B}{Task Definitions and Evaluation Criteria}
\appsubsectionentry{app:task_specifications}{B.1}{Task Taxonomy}
\appsubsectionentry{app:perception_metrics}{B.2}{Perception Metrics and Clinical Triplet Matching}
\appsubsectionentry{app:clinical_rubrics}{B.3}{Clinical Rubrics and Score Aggregation}

\appsectionentry{app:evaluator_training}{C}{AnesTRACE-Eval Training and Validation}
\appsubsectionentry{app:evaluator_data}{C.1}{Evaluator Data and Clinical Evidence}
\appsubsectionentry{app:evaluator_configuration}{C.2}{Training Configuration}
\appsubsectionentry{app:evaluator_validation}{C.3}{Agreement with Anesthesiologists}

\appsectionentry{app:model_setup}{D}{Experimental and Agent Protocols}
\appsubsectionentry{app:model_configuration}{D.1}{Model Inventory and Runtime Configuration}
\appsubsectionentry{app:inference_protocols}{D.2}{Inference Protocols and Prompt Templates}
\appsubsectionentry{app:agent_framework}{D.3}{Agent Framework, Memory, and Tools}
\appsubsectionentry{app:agent_metrics}{D.4}{Agent Behavior and Efficiency Metrics}

\appsectionentry{app:experiments}{E}{Complete Experiments Results}
\appsubsectionentry{app:l1_results}{E.1}{Intraoperative Perception Results}
\appsubsectionentry{app:single_point_decision_results}{E.2}{Single-point Decision-Making Results}
\appsubsectionentry{app:multi_step_decision_results}{E.3}{Multi-step Decision-Making Results}

\appsectionentry{app:performance_analyses}{F}{Ablation and Performance Analysis}
\appsubsectionentry{app:scale_specialization}{F.1}{Model Scale and Domain Specialization}
\appsubsectionentry{app:dimension_analysis}{F.2}{Dimension-wise Performance Analysis}
\appsubsectionentry{app:perception_reasoning}{F.3}{Perception--Reasoning Bottleneck Analysis}
\appsubsectionentry{app:quality_efficiency}{F.4}{Quality--Efficiency Trade-offs}
\appsubsectionentry{app:surgical_subgroup}{F.5}{Models performance in Different Surgical Subgroup}
\appsubsectionentry{app:error_type_distribution}{F.6}{Error Analysis for Decision-Making Tasks}

\appsectionentry{app:case_studies}{G}{Qualitative Case Study}

\endgroup

\clearpage

\section{Dataset Construction and Annotation}
\label{app:dataset_construction}
\subsection{Data Sources and Harmonization}
\label{app:data_sources}

AnesTRACE derives patient states from four open-source datasets: VitalDB, MOVER, INSPIRE, and EchoNet-TEE~\citep{vitaldb,mover,inspire,echo_net}. L1 uses data from VitalDB, MOVER-SIS, and EchoNet-TEE; L2 uses VitalDB; and L3 uses linked VitalDB--INSPIRE records and MOVER-EPIC.

For VitalDB--INSPIRE alignment, we exactly match INSPIRE \texttt{operations.csv.case\_id} to VitalDB \texttt{caseid}, yielding 2,594 locally verifiable linked cases. Procedure start aligns VitalDB seconds with INSPIRE minutes:

\begin{equation}
t^{\mathrm{INSPIRE}}
= t_{\mathrm{op}}^{\mathrm{INSPIRE}}
+ \frac{t^{\mathrm{VDB}}-t_{\mathrm{op}}^{\mathrm{VDB}}}{60}.
\label{eq:vitaldb_inspire_alignment}
\end{equation}

We retain only temporally valid records within each procedure interval and treat INSPIRE as low-frequency perioperative context rather than second-level observations. Cases are included only when the clinical context, timeline, evidence, and recorded actions can be reliably reconstructed from information available at the decision point; invalid linkage, insufficient coverage, unresolved units, artifacts, and hindsight-dependent references are excluded. Dataset-specific parsers map records into the unified $(C, M, O)$ representation while preserving missing or conflicting values. For L3, trajectories are serialized as ordered turns, with references and future information removed before inference. Released task files contain only de-identified identifiers and relative time, and translation preserves all structured clinical values and metadata.

\subsection{Clinical Terminology}
\label{app:clinical_terminology}

Table~\ref{tab:clinical_terminology} summarizes the recurring medical terms
represented in our AnesTRACE-Bench. Terms occurring in at
least five records were retained after normalizing case, spelling, and close
synonyms; closely related concepts were combined to keep the glossary compact.

\begin{table*}[!t]
\centering
\caption{Clinical terms and abbreviations used throughout AnesTRACE-Bench.}
\label{tab:clinical_terminology}
\scriptsize
\setlength{\tabcolsep}{5pt}
\renewcommand{\arraystretch}{0.92}
\begin{tabularx}{\textwidth}{@{}
    >{\RaggedRight\arraybackslash}p{0.35\textwidth}
    @{\hspace{2em}}
    >{\RaggedRight\arraybackslash}X
@{}}
\toprule
\textbf{Clinical term} & \textbf{Definition} \\
\midrule
\multicolumn{2}{@{}l}{\textit{Physiological measurements and monitoring}} \\
ASA physical status & Preoperative classification of systemic health status. \\
TEE & Transesophageal echocardiography for intraoperative cardiac imaging. \\
ECG & Electrocardiographic recording of cardiac electrical activity. \\
HR & Heart rate, reported as beats per minute. \\
MAP & Mean arterial pressure over one cardiac cycle. \\
SpO$_2$ & Peripheral oxygen saturation estimated by pulse oximetry. \\
EtCO$_2$ & Carbon-dioxide partial pressure measured at end-expiration. \\
FiO$_2$ & Fraction of oxygen in the inspired gas mixture. \\
RR & Respiratory rate, reported as breaths per minute. \\
PIP & Peak pressure reached in the airway during inspiration. \\
PEEP & Positive pressure maintained at end-expiration. \\
Tidal volume & Gas volume delivered or exhaled during one breath. \\
Minute ventilation & Total gas volume ventilated per minute. \\
CVP & Central venous pressure measured near the right atrium. \\
Cardiac index & Cardiac output normalized to body-surface area. \\
BIS & Processed electroencephalographic index of hypnotic depth. \\
Arterial blood gas & Blood analysis of oxygenation, ventilation, and acid--base status. \\
Lactate & Blood marker associated with anaerobic metabolism and impaired perfusion. \\
\midrule
\multicolumn{2}{@{}l}{\textit{Anesthetic and vasoactive medications}} \\
Propofol & Intravenous hypnotic used to induce or maintain general anesthesia. \\
Remifentanil & Ultra-short-acting opioid used for intraoperative analgesia. \\
Fentanyl & Opioid analgesic used as part of perioperative anesthesia. \\
Sevoflurane / desflurane & Volatile inhalational anesthetics used to maintain anesthesia. \\
Rocuronium & Non-depolarizing neuromuscular blocker. \\
Norepinephrine & Catecholamine vasopressor used to support arterial pressure. \\
Phenylephrine & Alpha$_1$-adrenergic vasopressor that increases vascular tone. \\
Epinephrine & Adrenergic agonist with inotropic and vasopressor effects. \\
Ephedrine & Mixed-acting sympathomimetic used to increase blood pressure. \\
Vasopressin & Non-catecholamine vasopressor that increases vascular tone. \\
Dobutamine & Predominantly beta$_1$-adrenergic inotrope that increases cardiac output. \\
Esmolol & Ultra-short-acting beta$_1$-adrenergic blocker. \\
Atropine & Antimuscarinic agent commonly used to treat bradycardia. \\
Nitroglycerin & Vasodilator that predominantly reduces venous tone and preload. \\
TCI / MAC & Target-controlled infusion and minimum alveolar concentration. \\
\midrule
\multicolumn{2}{@{}l}{\textit{Clinical conditions and diagnoses}} \\
Hypotension / hypertension & Abnormally low or high arterial pressure in clinical context. \\
Hypoxemia & Abnormally low oxygen level in arterial blood. \\
Hypercapnia & Abnormally elevated carbon-dioxide level in arterial blood. \\
Bradycardia / tachycardia & Abnormally slow or rapid heart rate in clinical context. \\
Arrhythmia & Abnormal cardiac rate or rhythm pattern. \\
Apnea / hypoventilation & Absent or inadequate ventilation. \\
Airway obstruction & Partial or complete blockage that impairs airflow. \\
Bronchospasm & Bronchial smooth-muscle constriction that increases airway resistance. \\
Rebreathing & Re-inhalation of exhaled gas containing carbon dioxide. \\
Pneumothorax & Air in the pleural space that may impair lung expansion. \\
Pulmonary embolism & Obstruction of pulmonary arterial flow by embolic material. \\
Hemorrhage & Acute blood loss that may compromise circulating volume. \\
Cardiac arrest & Cessation of effective cardiac activity and circulation. \\
Acidosis & Abnormally low blood pH. \\
Essential hypertension & Chronic arterial hypertension without a secondary cause. \\
Type 2 diabetes mellitus & Chronic disorder of insulin resistance and glucose regulation. \\
Asthma & Chronic inflammatory airway disease with reversible obstruction. \\
Obstructive sleep apnea & Recurrent upper-airway collapse during sleep. \\
Pneumonia & Infection and inflammation of pulmonary parenchyma. \\
Coronary artery disease & Atherosclerotic disease limiting coronary blood flow. \\
Heart failure & Inability of the heart to meet circulatory demand adequately. \\
Atrial fibrillation & Supraventricular arrhythmia with irregular atrial activation. \\
Lung cancer & Malignant neoplasm arising in the bronchus or lung. \\
\midrule
\multicolumn{2}{@{}l}{\textit{Anesthetic procedures and surgery types}} \\
General anesthesia & Drug-induced unconsciousness requiring anesthetic support. \\
Mechanical ventilation & Machine-delivered respiratory support using controlled settings. \\
Lobectomy / wedge resection & Removal of a pulmonary lobe or wedge of lung tissue. \\
Video-assisted thoracoscopic surgery & Minimally invasive thoracic surgery performed through ports. \\
Bronchoscopy & Endoscopic examination or treatment of the tracheobronchial tree. \\
Hepatectomy / liver transplantation & Liver resection or replacement with a donor liver. \\
Gastrectomy / esophagectomy & Surgical removal of part or all of the stomach or esophagus. \\
Colectomy / hemicolectomy & Removal of all or part of the colon. \\
Laparoscopy / laparotomy & Abdominal access through small ports or an open incision. \\
Coronary artery bypass grafting & Graft-based revascularization around coronary stenoses. \\
Partial nephrectomy & Removal of a renal lesion while preserving remaining kidney tissue. \\
\bottomrule
\end{tabularx}
\vspace{-3mm}
\end{table*}
\subsection{Decision Anchors and Episodes}
\label{app:task_construction}
A decision anchor $t_k$ marks a clinically relevant reassessment point. Its visible state is $S_k=(C,M_{t_k},O_{W_k^{\mathrm{obs}}})$, with $W_k^{\mathrm{obs}}=[t_k-\Delta_{\mathrm{obs}},t_k)$ restricted to prior evidence. Persistent abnormalities, phase transitions, and newly available tests define an initial pool of 5,017 candidate anchors and more than 9,000 multimodal instances. The final benchmark units are selected from this pool after artifact screening, temporal validation, and expert review. L2 exposes information available at or before T0, while L3 releases observations turn by turn. Future states, references, and screening labels remain hidden.

An L3 episode is an ordered segment of one recorded surgical trajectory~\citep{bynum2014participation,culwick2020bowtie,runciman2005crisis}. After each response, the next recorded state is released without simulating the effect of the model's recommendation. L1 tests perception from TEE, waveforms, and trends; L2 tests diagnosis, risk, intervention, and reassessment at one anchor; L3 tests repeated state assessment and management as evidence evolves. References specify required elements, acceptable alternatives, and severe errors rather than one wording.

L3 provides six runner-bound context tools for patient profile, procedure/anesthesia context, medication state, the preceding intervention, visible tests, and episode memory, plus four knowledge tools. Full-tool runs expose the same inventory to each evaluated model, whereas per-turn budgets are fixed by the inference profile as detailed in Appendix~\ref{app:agent_framework}. Tools cannot query another patient or future state, and all calls and failures are logged.

\newcommand{\appmodelshade}[2]{{%
  \setlength{\fboxsep}{1.2pt}%
  \colorbox{#1}{\makebox[10em][l]{\strut #2}}}}
\newcommand{\appmodelsmall}[1]{\appmodelshade{ModelSmallBase!12}{#1}}
\newcommand{\appmodelmedium}[1]{\appmodelshade{ModelMediumBase!12}{#1}}
\newcommand{\appmodellarge}[1]{\appmodelshade{ModelLargeBase!14}{#1}}

\begin{table*}[t]
\centering
\caption{Annotation workload of the anesthesiologist team, with task-specific annotation counts and time requirements across AnesTRACE-Bench.}
\label{tab:evaluator_annotation_workload}
\small
\setlength{\tabcolsep}{8pt}
\renewcommand{\arraystretch}{1.1}
\begin{tabular}{lrrrr}
\toprule
\textbf{Task} & \textbf{Annotations} & \textbf{Annotation time} &
\textbf{Avg. time/item} & \textbf{Time/100 items} \\
\midrule
Level One -- TEE & $1000$ & $10$ h & $0.60$ min & $1$ h \\
Level One -- Waveform & $1000$ & $7$ h & $0.42$ min & $0.70$ h \\
Level Two & $2000$ & $26$ h & $0.78$ min & $1.30$ h \\
Level Three & $1659$ & $32$ h & $1.16$ min & $1.93$ h \\
\midrule
\textbf{Total} & $\mathbf{5659}$ & $\mathbf{75}$ h & $\mathbf{0.80}$ min & $\mathbf{1.33}$ h \\
\bottomrule
\end{tabular}
\vspace{-3mm}
\end{table*}

\begin{figure*}[t]
    \centering
    \includegraphics[width=\textwidth]{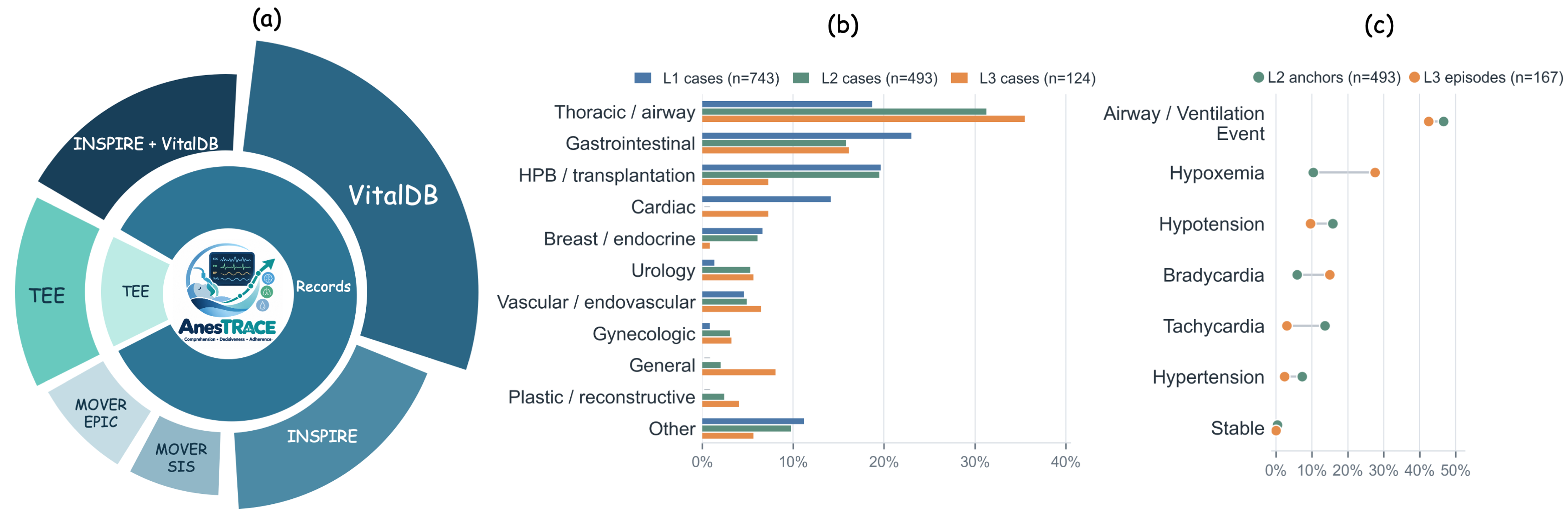}
    \caption{Data sources and distributions in AnesTRACE-Bench: (a) contributing data sources and modalities, (b) surgical-category distributions across the three levels, and (c) intraoperative event distributions in L2 and L3.}
    \label{fig:datasets_disrtibution}
    \vspace{-3mm}
\end{figure*}

\subsection{Expert Annotation and Dataset Statistics}
\label{app:datasets_annotation}
After collecting and preprocessing records from the source datasets, we constructed candidate annotation instances using deterministic temporal and clinical filtering rules. DeepSeek-v4-Pro was used to generate candidate questions and responses for each instance~\citep{xu2026deepseek}.

These candidates were presented to 11 anesthesiologists with varying levels of clinical experience, who reviewed, corrected, and finalized the annotations on the PAI-iTAG platform\footnote{Alibaba Cloud PAI-iTAG: \url{https://www.aliyun.com/product/bigdata/learn/itag}.}. Specifically, the annotation team comprised 5 anesthesiologists with 1--3 years, 3 with 4--5 years, and 3 with more than 5 years of clinical experience. The team contributed 75 physician-hours across 5,659 benchmark annotations, as summarized in Table~\ref{tab:evaluator_annotation_workload}. All final annotations were therefore determined by clinical experts rather than the generating model. The reviewed annotations were subsequently used to construct the benchmark and evaluator-training data.

Figure~\ref{fig:datasets_disrtibution} summarizes the data sources and distributions of AnesTRACE-Bench. Panel (a) presents the contributing data sources and modalities, panel (b) compares the distribution of surgical categories across the three levels, and panel (c) shows the distribution of intraoperative events in L2 and L3. Thoracic/airway, gastrointestinal, and HPB/transplantation procedures account for substantial proportions of the surgical cases. The decision-oriented levels further cover a broad range of intraoperative events, including airway and ventilation events as well as hemodynamic and oxygenation abnormalities such as hypotension, hypoxemia, bradycardia, and tachycardia.

We complement these structured distributions by summarizing the clinical concepts represented in L2 and L3 in Figure~\ref{fig:a3_clinical_wordcloud}.

\par
{\centering
\includegraphics[width=0.70\linewidth]{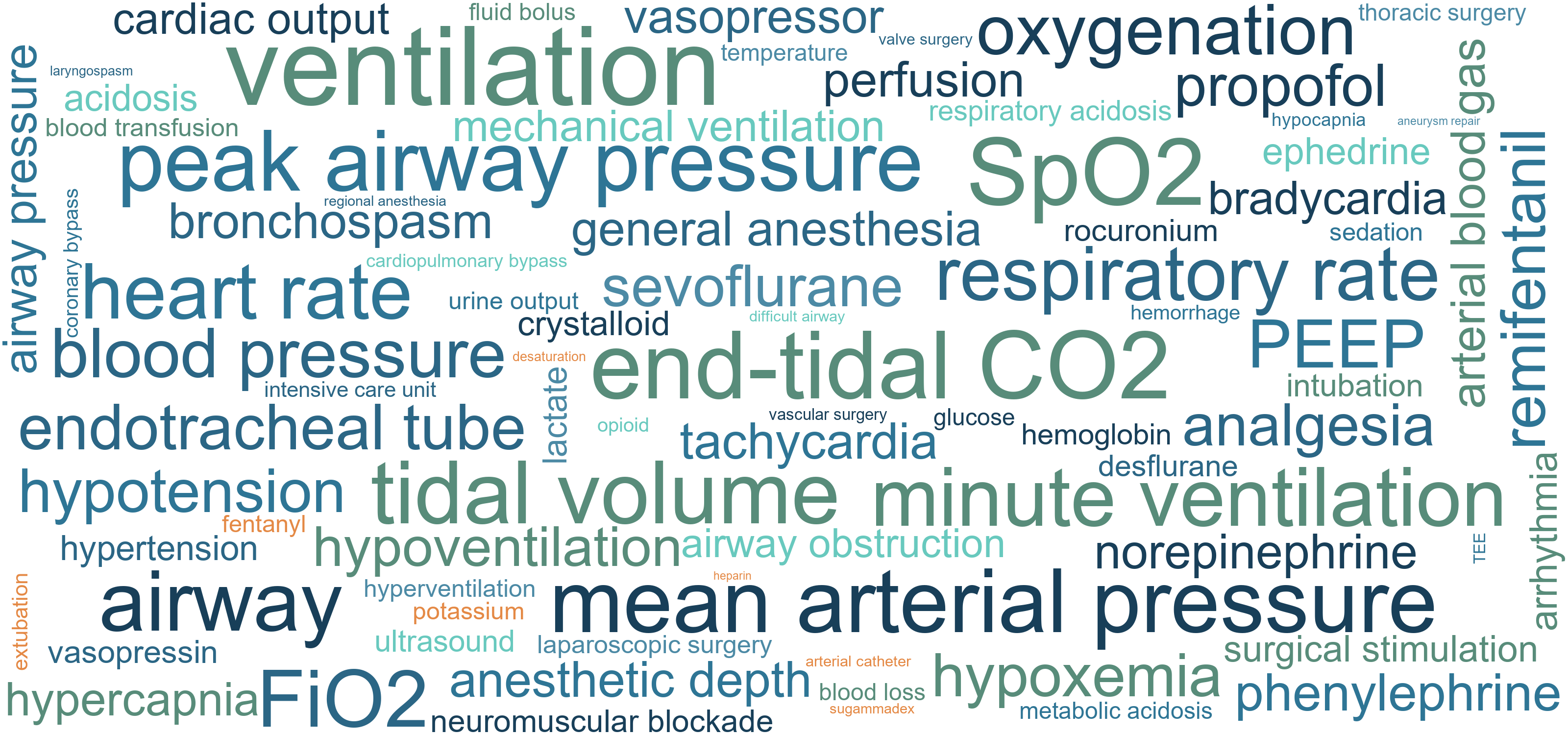}
\captionof{figure}{Concept cloud of frequently occurring clinical terms in AnesTRACE-Bench. Font size reflects term frequency.}
\label{fig:a3_clinical_wordcloud}
\vspace{-3mm}
\par}

\begin{table*}[t]
\centering
\providecommand{\AnesYes}{\ensuremath{\checkmark}}
\providecommand{\AnesPartial}{\ensuremath{\triangle}}
\providecommand{\AnesNo}{\ensuremath{\times}}
\caption{
Comparison of medical and anesthesiology benchmarks.
Temporal Updating denotes evaluation of management decisions as new
patient information becomes available, rather than temporal perception alone.
\AnesYes\ indicates explicit coverage, \AnesPartial\ partial or
supplementary coverage, and \AnesNo\ that the capability is not an
explicit evaluation target. Visual/Signal Input in AnesTRACE are assessed
the performance that direct image, video, or physiological-signal representation are supplied to the model.
}
\renewcommand{\arraystretch}{1.12}
\setlength{\tabcolsep}{3.5pt}
\resizebox{\textwidth}{!}{%
\begin{tabular}{l*{8}{c}}
\toprule
\multirow{2}{*}{\textbf{Benchmark}}
& \multicolumn{3}{c}{\textbf{Anesthesia-Related Capabilities}}
& \multicolumn{4}{c}{\textbf{Evaluation Dimensions}}
& \textbf{Efficiency} \\
\cmidrule(lr){2-4}
\cmidrule(lr){5-8}
\cmidrule(lr){9-9}
& \makecell{\textbf{Intraoperative}\\\textbf{Decision}}
& \makecell{\textbf{Physiological}\\\textbf{Waveform}}
& \textbf{TEE}
& \makecell{\textbf{Visual/Signal}\\\textbf{Input}}
& \makecell{\textbf{Temporal}\\\textbf{Updating}}
& \makecell{\textbf{Open-ended}\\\textbf{Management}}
& \makecell{\textbf{Safety}\\\textbf{Evaluation}}
& \makecell{\textbf{Response}\\\textbf{Latency}} \\
\midrule

\multicolumn{9}{l}{\textit{General medical perception}} \\
GMAI-MMBench~\citep{chen2024gmai}
& \AnesNo & \AnesNo & \AnesNo
& \AnesYes & \AnesNo & \AnesNo & \AnesNo & \AnesNo \\
OmniMedVQA~\citep{omnimedvqa}
& \AnesNo & \AnesNo & \AnesNo
& \AnesYes & \AnesNo & \AnesNo & \AnesNo & \AnesNo \\
ECG-QA~\citep{ecg_qa}
& \AnesNo & \AnesYes & \AnesNo
& \AnesYes & \AnesNo & \AnesNo & \AnesNo & \AnesNo \\

\midrule
\multicolumn{9}{l}{\textit{General clinical evaluation and decision-making}} \\
HealthBench~\citep{healthbench}
& \AnesNo & \AnesNo & \AnesNo
& \AnesNo & \AnesNo & \AnesYes & \AnesYes & \AnesNo \\
LLMEval-Med~\citep{zhang2025llmeval}
& \AnesNo & \AnesNo & \AnesNo
& \AnesNo & \AnesNo & \AnesPartial & \AnesYes & \AnesNo \\
MIMIC-CDM~\citep{hager2024mimiccdm}
& \AnesNo & \AnesNo & \AnesNo
& \AnesNo & \AnesPartial & \AnesYes & \AnesPartial & \AnesNo \\
LiveClin~\citep{wang2026liveclin}
& \AnesNo & \AnesNo & \AnesNo
& \AnesYes & \AnesPartial & \AnesNo & \AnesNo & \AnesNo \\
AgentClinic~\citep{agentclinic}
& \AnesNo & \AnesNo & \AnesNo
& \AnesYes & \AnesPartial & \AnesPartial & \AnesNo & \AnesNo \\
MedAgentBench~\citep{medagentbench}
& \AnesNo & \AnesNo & \AnesNo
& \AnesNo & \AnesNo & \AnesPartial & \AnesNo & \AnesNo \\
LongMedBench~\citep{longmedbench}
& \AnesNo & \AnesNo & \AnesNo
& \AnesNo & \AnesYes & \AnesNo & \AnesNo & \AnesNo \\

\midrule
\multicolumn{9}{l}{\textit{Surgical visual benchmarks}} \\
SurgVLM-Bench~\citep{zeng2025surgvlm}
& \AnesNo & \AnesNo & \AnesNo
& \AnesYes & \AnesNo & \AnesNo & \AnesNo & \AnesNo \\
SurgCoTBench~\citep{low2026surgraw}
& \AnesNo & \AnesNo & \AnesNo
& \AnesYes & \AnesNo & \AnesNo & \AnesNo & \AnesNo \\

\midrule
\multicolumn{9}{l}{\textit{Anesthesiology benchmarks}} \\
CAB~\citep{CAB}
& \AnesPartial & \AnesNo & \AnesNo
& \AnesNo & \AnesNo & \AnesPartial & \AnesYes & \AnesNo \\
Hypnos benchmark~\citep{wang2025hypnos}
& \AnesPartial & \AnesNo & \AnesNo
& \AnesNo & \AnesNo & \AnesPartial & \AnesYes & \AnesNo \\
AnesBench~\citep{Anessuite}
& \AnesPartial & \AnesNo & \AnesNo
& \AnesNo & \AnesNo & \AnesPartial & \AnesNo & \AnesNo \\

\midrule
\textbf{AnesTRACE (ours)}
& \AnesYes & \AnesYes & \AnesYes
& \AnesYes & \AnesYes & \AnesYes & \AnesYes & \AnesYes \\

\bottomrule
\end{tabular}%
}
\label{tab:benchmark_comparison}
\vspace{-3mm}
\end{table*}

\section{Task Definitions and Evaluation Criteria}
\label{app:evaluation_metrics}

We provide detailed task specifications for AnesTRACE-Bench in this section.
Table~\ref{tab:anestrace_task_taxonomy} summarizes the tasks across its three levels, including their inputs, answer formats, scales, and evaluation dimensions.
Table~\ref{tab:benchmark_comparison} compares AnesTRACE-Bench with existing medical, surgical, and anesthesiology benchmarks in terms of clinical scope, input modalities, temporal decision updating, safety evaluation, and response latency.

\subsection{Task Taxonomy and Input-Output Formats}
\label{app:task_specifications}

Table~\ref{tab:anestrace_task_taxonomy} summarizes the task taxonomy, input--output formats, evaluation scales, and metrics across the three levels of AnesTRACE-Bench. L1 evaluates intraoperative perception from TEE and physiological waveforms, covering recognition, localization, functional assessment, and waveform understanding. L2 evaluates single-point anesthesia decision-making from clinical context and pre-decision physiological trends, with open-ended outputs for risk prediction, diagnosis, intervention, and reassessment. L3 extends this setting to longitudinal decision-making, conditioning each turn on the current state and preceding history and additionally evaluating consistency across the full decision trajectory.

\begin{table*}[t]
\centering
\caption{Detailed three-level task hierarchy of AnesTRACE-Bench, including task inputs, answer formats, scales, and evaluation dimensions.}
\label{tab:anestrace_task_taxonomy}
\scriptsize
\renewcommand{\arraystretch}{1.08}
\setlength{\tabcolsep}{3pt}

\begin{tabularx}{\textwidth}{@{}
c
>{\raggedright\arraybackslash}p{0.16\textwidth}
>{\raggedright\arraybackslash}p{0.26\textwidth}
>{\raggedright\arraybackslash}p{0.14\textwidth}
>{\centering\arraybackslash}p{0.09\textwidth}
>{\raggedright\arraybackslash}X
@{}}
\toprule
\textbf{Level} & \textbf{Input} & \textbf{Task Type}
& \textbf{Answer Format} & \textbf{Scale}
& \textbf{Evaluation Dimensions} \\
\midrule

\multirow[c]{12}{*}{\textbf{L1}}
& TEE frame & Probe Position
& Single choice & 82 & Accuracy \\
& TEE frame & Imaging Axis
& Single choice & 79 & Accuracy \\
& TEE video & CFD Presence
& Single choice & 30 & Accuracy \\
& TEE frame & Standard Imaging Plane
& Single choice & 99 & Accuracy \\
& TEE frame & Structure Visibility
& Single choice & 89 & Accuracy \\
& TEE frame & Structure Localization
& Bounding box & 193 & mIoU \\
& TEE video & Assessability Judgment
& Single choice & 80 & Accuracy \\
& TEE video & Functional Assessment
& Labels + text & 83 & Accuracy; Factual-F1 \\
& TEE video & Abnormality Detection
& Labels + text & 82 & Accuracy; Factual-F1 \\
\cmidrule(lr){2-6}
& Waveform plot & Value Extraction
& Single choice & 200 & Accuracy \\
& Waveform plot & Trend Recognition
& Multiple selection & 400 & Set-F1 \\
& Waveform plot & Abnormality Detection
& Structured response & 400 & Metric-F1 \\

\midrule
\multirow[c]{4}{*}{\textbf{L2}}
& \multirow[c]{4}{0.16\textwidth}
  {Clinical context + pre-decision trends}
& Risk Prediction
& Open-ended
& \multirow[c]{4}{*}{\shortstack{493\\anchors}}
& \multirow[c]{4}{=}{%
  \makecell[l]{
    Clinical Correctness\\
    Evidence Grounding\\
    Task Completeness\\
    Safety Error (B3/B4)
  }} \\
& & Clinical Diagnosis & Open-ended & & \\
& & Intervention Decision & Open-ended & & \\
& & Reassessment Plan & Open-ended & & \\

\midrule
\multirow[c]{2}{*}{\textbf{L3}}
& Current state + history
& Turn-level Assessment and Intervention
& Structured JSON
& 517 turns
& \makecell[l]{
    Clinical Correctness\\
    Evidence Grounding\\
    Task Completeness\\
    Safety Error
  } \\
& Recorded episode
& Whole-trajectory Consistency
& Decision sequence
& \shortstack{167\\episodes}
& \makecell[l]{
    Temporal Evidence and\\
    Response Adaptation\\
    Longitudinal Management\\
    Coherence
  } \\

\bottomrule
\end{tabularx}
\vspace{-3mm}
\end{table*}

\subsection{Perception Metrics and Clinical Triplet Matching}
\label{app:perception_metrics}
AnesTRACE-Bench uses task-specific evaluation metrics. Single-choice perception tasks are evaluated using accuracy. For a multiple selection question with reference set $Y_i$ and predicted set $\hat{Y}_i$, the sample-level Set-F1 is
\begin{equation}
\mathrm{SetF1}_i
=
\frac{2|Y_i\cap\hat{Y}_i|}
{|Y_i|+|\hat{Y}_i|}.
\end{equation}
The reported Set-F1 is the macro-average over questions. If both sets are empty, the score is defined as one; an invalid or missing prediction receives zero.

Spatial grounding is evaluated using intersection over union:
\begin{equation}
\mathrm{IoU}_i
=
\frac{|B_i\cap\hat{B}_i|}
{|B_i\cup\hat{B}_i|},
\qquad
\mathrm{mIoU}
=
\frac{1}{N}\sum_{i=1}^{N}\mathrm{IoU}_i,
\end{equation}
where $B_i$ and $\hat{B}_i$ are the reference and predicted bounding boxes. 
Open-ended perception responses are represented as collections of atomic findings
\begin{equation}
\tau=(m,o,p),
\end{equation}
where $m$ identifies a physiological metric or anatomical structure, $o$ denotes the associated observation or clinical finding, and $p$ indicates its presence state. We use task-specific projections of this representation. For TEE functional assessment and abnormality detection, \emph{Factual-F1} matches all three fields after normalization. For waveform abnormality perception, \emph{Metric-F1} matches the canonical metric identity $m$ only.

For each metric, matched findings count as true positives (TP), while unmatched predictions and references count as false positives (FP) and false negatives (FN), respectively:
\begin{equation}
\mathrm{F1}_i
=
\frac{2\mathrm{TP}_i}
{2\mathrm{TP}_i+\mathrm{FP}_i+\mathrm{FN}_i}.
\end{equation}
TEE Factual-F1 is macro-averaged over fact-scoring-eligible questions within each task; invalid or missing predictions receive zero. When an aggregate TEE Factual-F1 is reported, the functional-assessment and abnormality-detection task scores are averaged with equal weight. Metric and anatomical aliases are normalized before matching.

\subsection{Clinical Rubrics and Score Aggregation}
\label{app:clinical_rubrics}

AnesTRACE-Eval evaluates open-ended L2 and L3 responses against the patient
information visible at the decision point, anesthesiologist-defined atomic
criteria, the expert reference, and retrieved medical evidence. Patient-specific
evidence is primary and references and external knowledge define acceptable
clinical boundaries rather than a unique answer.

The five evaluation dimensions are defined in
Table~\ref{tab:clinical_evaluation_dimensions}. The three turn-level ordinal
dimensions apply to L2 and L3, safety is assessed for management outputs, and
Temporal Consistency is evaluated only for complete L3 trajectories.

\begin{table*}[t]
\centering
\caption{Evaluation dimensions of AnesTRACE-Eval and their detailed definitions.}
\label{tab:clinical_evaluation_dimensions}
\footnotesize
\setlength{\tabcolsep}{8pt}
\renewcommand{\arraystretch}{1.18}
\begin{tabularx}{\textwidth}{@{}
    >{\RaggedRight\arraybackslash}p{0.21\textwidth}
    |>{\itshape\RaggedRight\arraybackslash}X
@{}}
\toprule
\textbf{Dimension} & \textbf{Definition} \\
\midrule
Clinical Correctness &
Evaluate the primary clinical conclusion and severity, together with the
direction, priority, and key actions of the proposed management; a diagnosis or
severity error that materially changes management, or a dangerous or
contraindicated action, constitutes a decisive defect. \\
\midrule
Evidence Grounding &
Assess whether conclusions and actions follow from the available case evidence
with appropriate temporal and causal reasoning, clinical prioritization, and
calibration; unsupported claims, reversed evidence, unavailable information,
and neglected relevant history are penalized. \\
\midrule
Task Completeness &
Examine whether all required content categories are present, including the
diagnosis and its rationale or the required actions and plan details; factual
or evidential errors are scored under the preceding dimensions rather than
double-counted as missing content. \\
\midrule
Safety &
Identify avoidable harm arising from an unsafe action, consequential omission,
inappropriate delay, dosing error, or failure to escalate, and classify its
potential consequence as \emph{safe}, \emph{minor}, \emph{major}, or
\emph{critical}. \\
\midrule
Temporal Consistency (L3) &
Determine whether later decisions incorporate new evidence and observed
responses to recorded interventions, and whether management remains coherent by
appropriately maintaining, escalating, de-escalating, or revising its strategy
across turns. \\
\bottomrule
\end{tabularx}
\vspace{-3mm}
\end{table*}

Together, these dimensions assess response quality at the current decision
point and, for L3, the coherence of decision updating across the trajectory.

We score each non-safety dimension on $\{0,1,2\}$. A score of 2 indicates that
the response satisfies the relevant criteria without a clinically material
defect; 1 indicates partial satisfaction with a bounded omission or
imprecision; and 0 indicates a decisive error, unsupported conclusion, or
failure to provide a required component.

Safety is reported separately as \emph{safe}, \emph{minor}, \emph{major}, or \emph{critical}. Because safety severity is not assumed to be linearly equivalent to the ordinal clinical dimensions, it is not converted into a compensatory numeric score. The principal safety outcome is the Major/Critical Safety Error Rate:
\begin{equation}
\mathrm{Safety\ Error}
=
\frac{1}{N}
\sum_{i=1}^{N}
\mathbb{I}
\left[
s^{(i)}_{\mathrm{saf}}
\in
\{\textit{major},\textit{critical}\}
\right].
\end{equation}
The denominator includes only responses with a valid safety label.
Missing or failed safety evaluations are excluded from this denominator
and reported separately.

For each ordinal dimension $d$, model-level performance is reported as the mean over its scored evaluation units:
\begin{equation}
\bar{s}_d
=
\frac{1}{N_d}
\sum_{i=1}^{N_d}s_{i,d}.
\end{equation}
Ordinal scores are normalized to a 0--100 scale by multiplying
the mean raw score by 50. The turn-level average equally weights
Clinical Correctness, Evidence Grounding, and Task Completeness;
Safety Error is reported separately.
L2 uses decision anchors as evaluation units, whereas L3 uses individual decision turns. Dimension scores and Safety Error are reported separately so that a high average clinical score cannot obscure the frequency of serious safety errors.

\section{AnesTRACE-Eval Training and Validation}
\label{app:evaluator_training}

\subsection{Evaluator Data and Clinical Evidence}
\label{app:evaluator_data}

AnesTRACE-Eval evaluates open-ended Level~2 and Level~3 responses. Each input contains the decision-point context, question, candidate response, atomic reference criteria, and, for Level~3, the preceding trajectory and newly revealed state. The scoring rubric is defined in Appendix~\ref{app:clinical_rubrics}.

We use DeepSeek-V4-Pro to construct the initial evaluator-training data from benchmark cases, reference annotations, and the predefined evaluation rubric. The model generates candidate scores and rationales for SFT, as well as contrasting response pairs for preference annotation. All generated candidates are then reviewed, corrected, and formally annotated by anesthesiologists, with preference pairs additionally ranked by the experts. The resulting corpus contains approximately 7k expert-annotated SFT examples and 4k expert-ranked DPO~\citep{dpo} pairs.

The human-validation set is constructed separately from actual benchmark
responses produced by five generator models whose outputs are excluded
from evaluator training.
Neither the validation responses nor outputs from these five generators
are included in the SFT examples or DPO preference pairs.
The validation models and sampling procedure are detailed in
Appendix~\ref{app:evaluator_validation}.

For each case, a retrieval module supplies case-relevant passages from Miller's Anesthesia, ASA guidance, pharmacological references, and related literature.  Retrieved passages are provided as auxiliary evidence to the evaluator; patient-specific observations visible at the decision point remain primary, and future-state information is excluded.  The retriever is not treated as a separately fine-tuned clinical judge: SFT and DPO train Qwen3.5-9B to use the supplied evidence when producing rubric scores and rationales.

\subsection{Training Configuration}
\label{app:evaluator_configuration}

All experiments were conducted on 2 NVIDIA H100 80GB GPUs.  SFT and DPO
required approximately 17 and 2 hours of wall-clock time, corresponding to
approximately 34 and 4 H100 GPU-hours, respectively.  The complete settings
are summarized in Table~\ref{tab:evaluator_training_config}.

\begin{table*}[t]
\centering
\caption{Training configuration for AnesTRACE-Eval.}
\label{tab:evaluator_training_config}
\small
\setlength{\tabcolsep}{7pt}
\renewcommand{\arraystretch}{1.08}
\begin{tabular}{lcc}
\toprule
\textbf{Hyperparameter} & \textbf{SFT} & \textbf{DPO} \\
\midrule
Initialization & Qwen3.5-9B & SFT checkpoint \\
Fine-tuning & LoRA & LoRA \\
LoRA rank & $16$ & $8$ \\
LoRA $\alpha$ & $32$ & $16$ \\
LoRA dropout & $0.05$ & $0.05$ \\
LoRA target & All & All \\
Maximum sequence length & $8192$ & $6144$ \\
Per-device batch size & $2$ & $1$ \\
Gradient accumulation & $8$ & $16$ \\
Learning rate & $0.0001$ & $0.000003$ \\
Epochs & $3$ & $1$ \\
LR scheduler & Cosine & Cosine \\
Warmup ratio & $0.05$ & $0.05$ \\
Precision & BF16 & BF16 \\
Gradient checkpointing & Yes & Yes \\
Seed & $42$ & $42$ \\
DPO $\beta$ & -- & $0.1$ \\
Preference loss & -- & Sigmoid \\
DeepSpeed & -- & ZeRO-3 \\
\bottomrule
\end{tabular}
\end{table*}

\paragraph{Supervised fine-tuning.}
Qwen3.5-9B is SFT-trained on approximately 7,000 reviewed examples using the settings in Table~\ref{tab:evaluator_training_config}. Data are split at the patient or clinical-episode level, sequence packing is disabled, and checkpoints are saved every 50 optimizer steps.

\begin{figure*}[t]
  \centering
  \begin{minipage}[t]{0.48\textwidth}
    \centering
    \includegraphics[width=\linewidth]{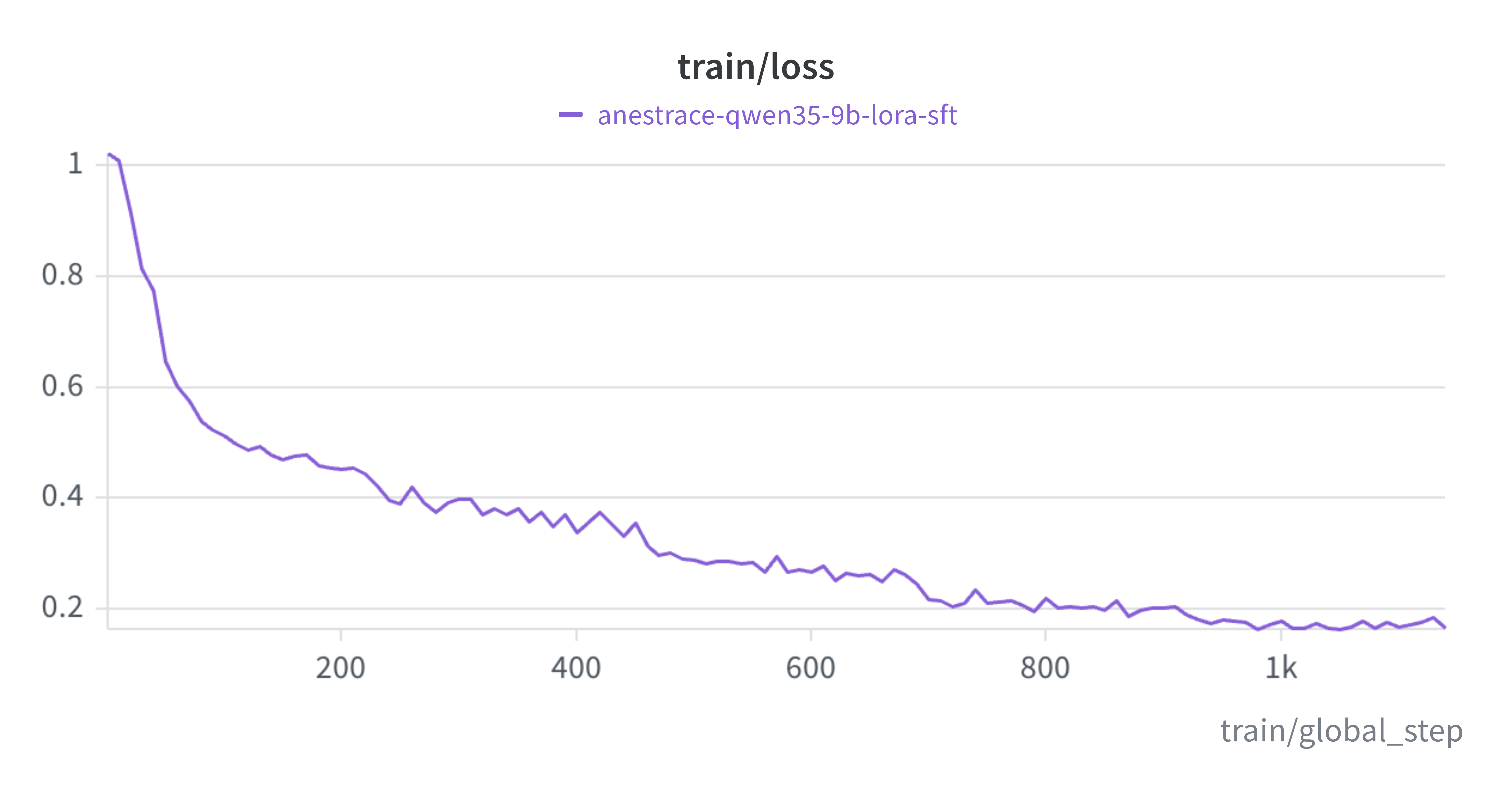}\\[-1mm]
    {\small (a) SFT training loss}
  \end{minipage}\hfill
  \begin{minipage}[t]{0.48\textwidth}
    \centering
    \includegraphics[width=\linewidth]{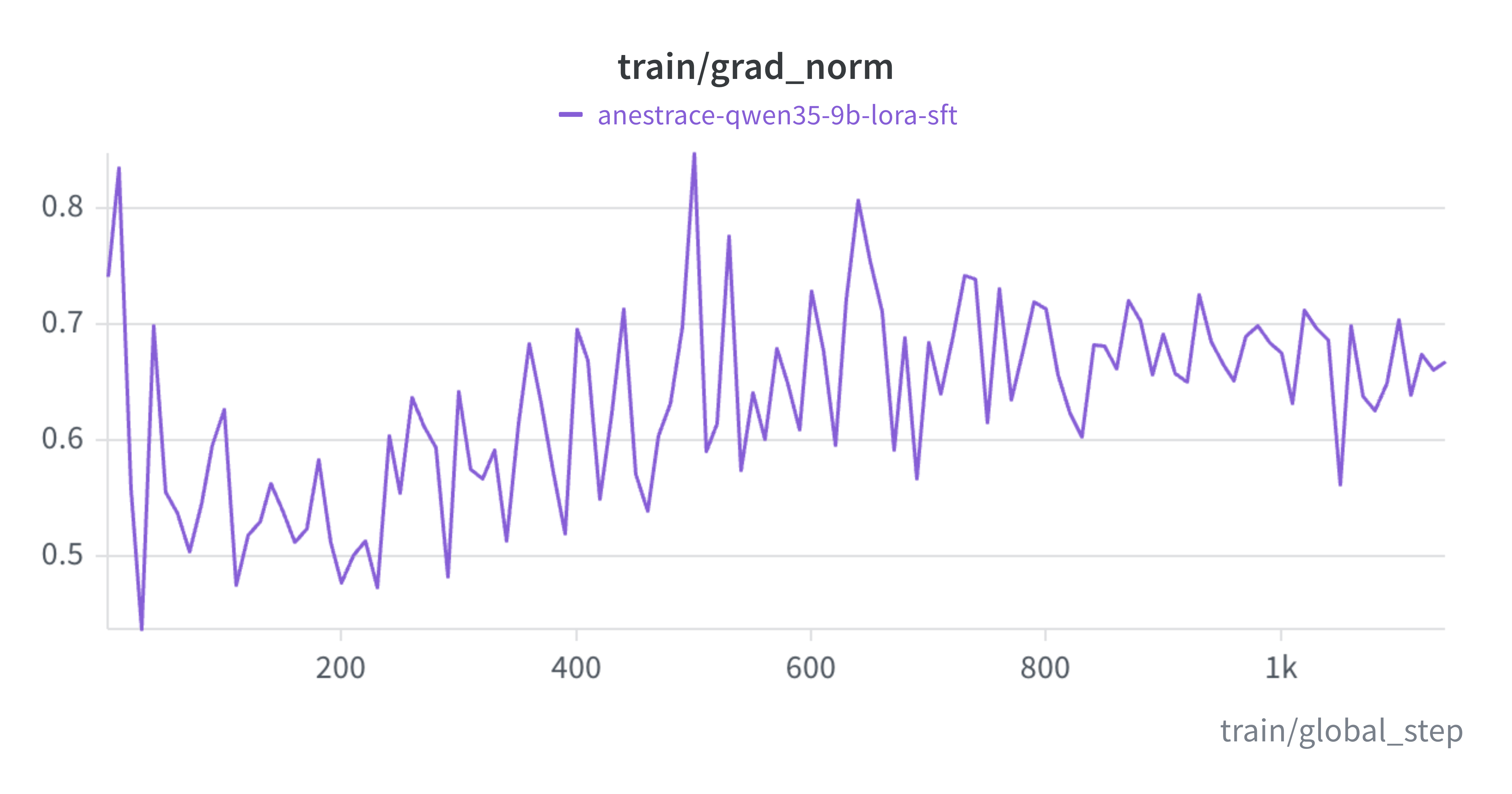}\\[-1mm]
    {\small (b) SFT gradient norm}
  \end{minipage}
  \caption{Optimization traces for supervised fine-tuning of AnesTRACE-Eval.}
  \label{fig:anestrace_eval_sft_training}
\end{figure*}

Figure~\ref{fig:anestrace_eval_sft_training} shows the available logged SFT traces.  The displayed loss is a logged/smoothed minibatch statistic.  The downward loss trajectory and the absence of sustained gradient-norm growth indicate numerically stable optimization, but do not by themselves measure held-out clinical validity.

\begin{figure*}[t]
  \centering
  \begin{minipage}[t]{0.48\textwidth}
    \centering
    \includegraphics[width=\linewidth]{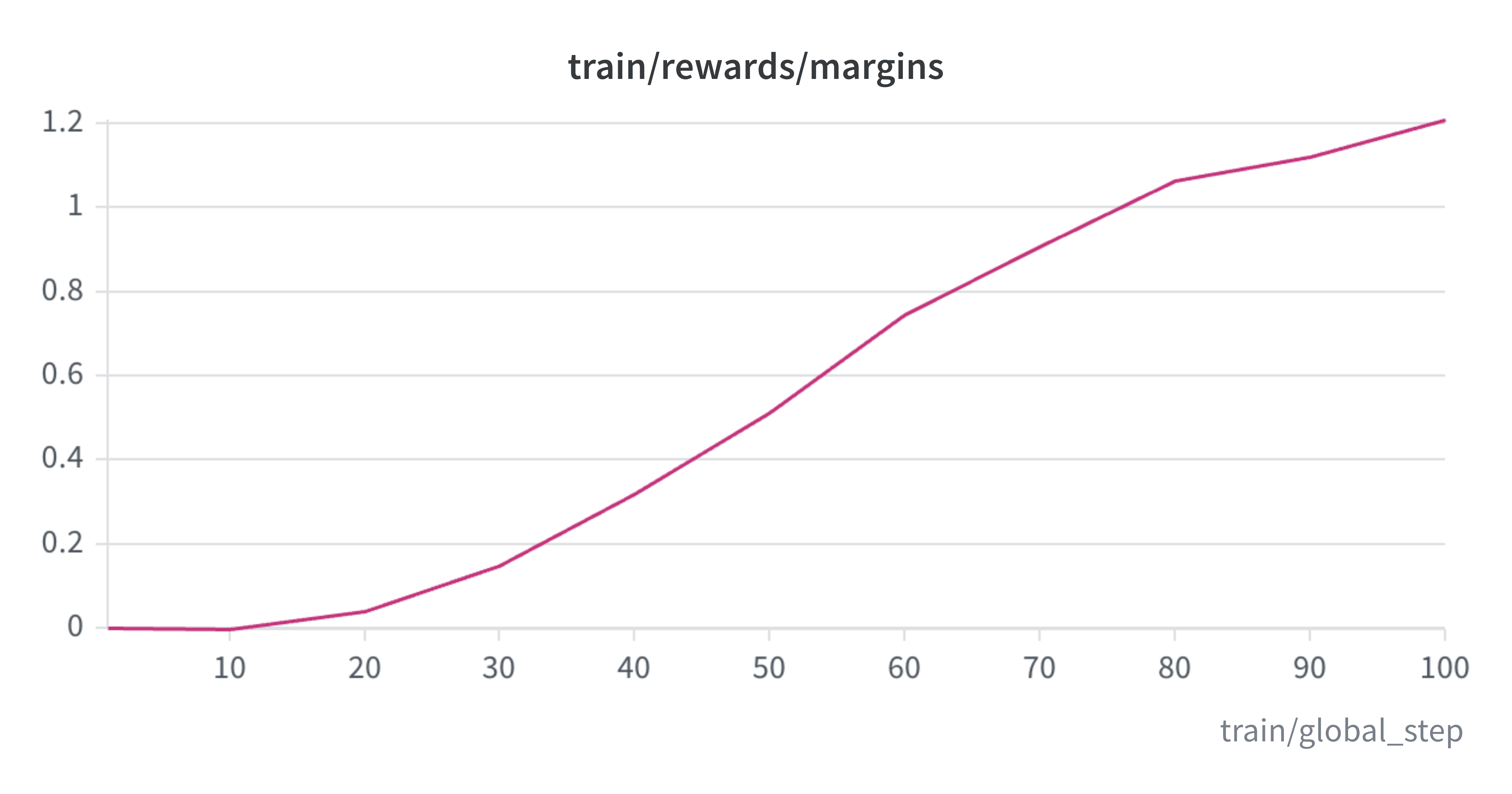}\\[-1mm]
    {\small (a) Reward margin}
  \end{minipage}\hfill
  \begin{minipage}[t]{0.48\textwidth}
    \centering
    \includegraphics[width=\linewidth]{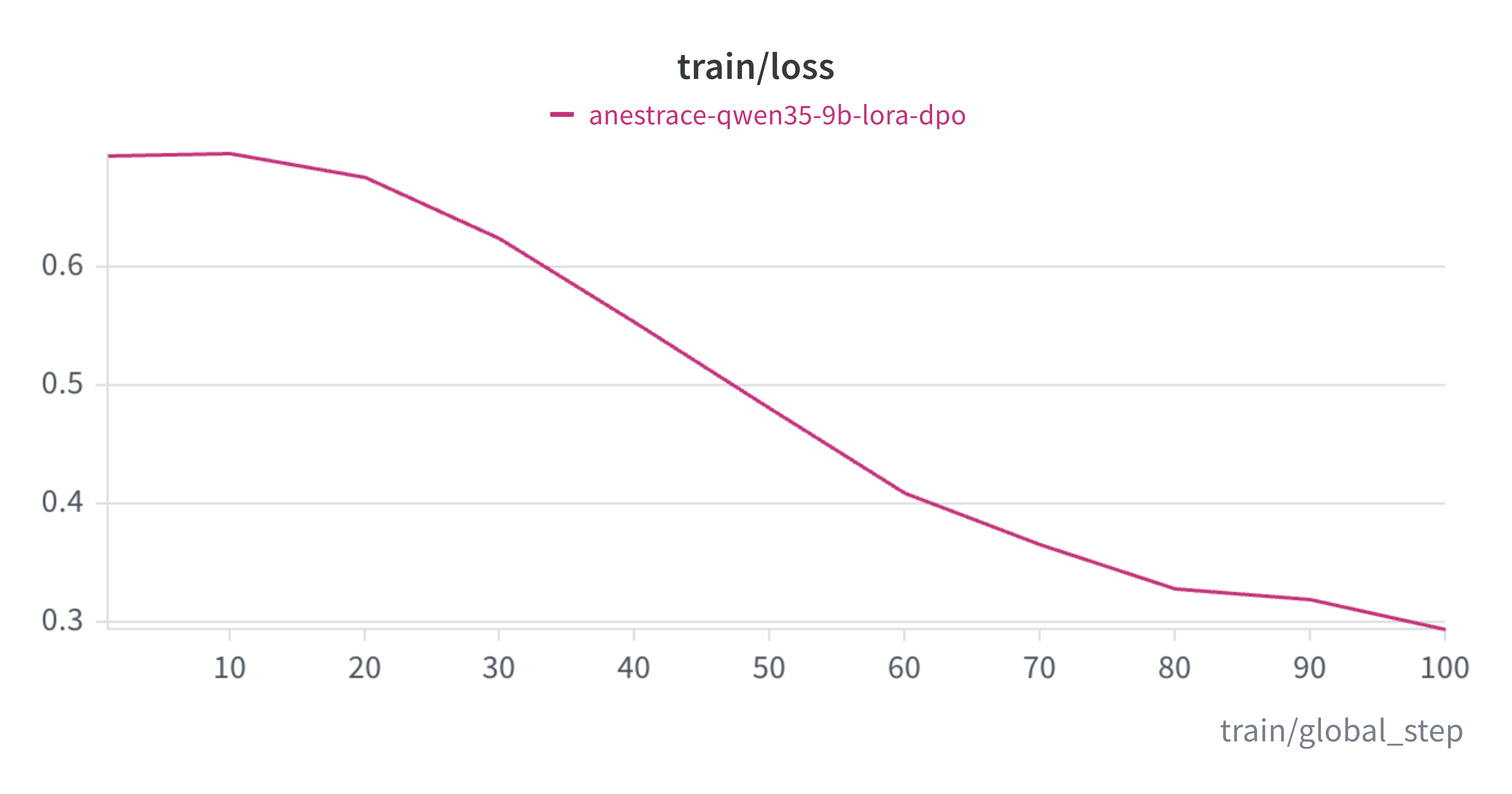}\\[-1mm]
    {\small (b) DPO training loss}
  \end{minipage}

  \vspace{2mm}
  \begin{minipage}[t]{0.48\textwidth}
    \centering
    \includegraphics[width=\linewidth,height=0.22\textheight,keepaspectratio]{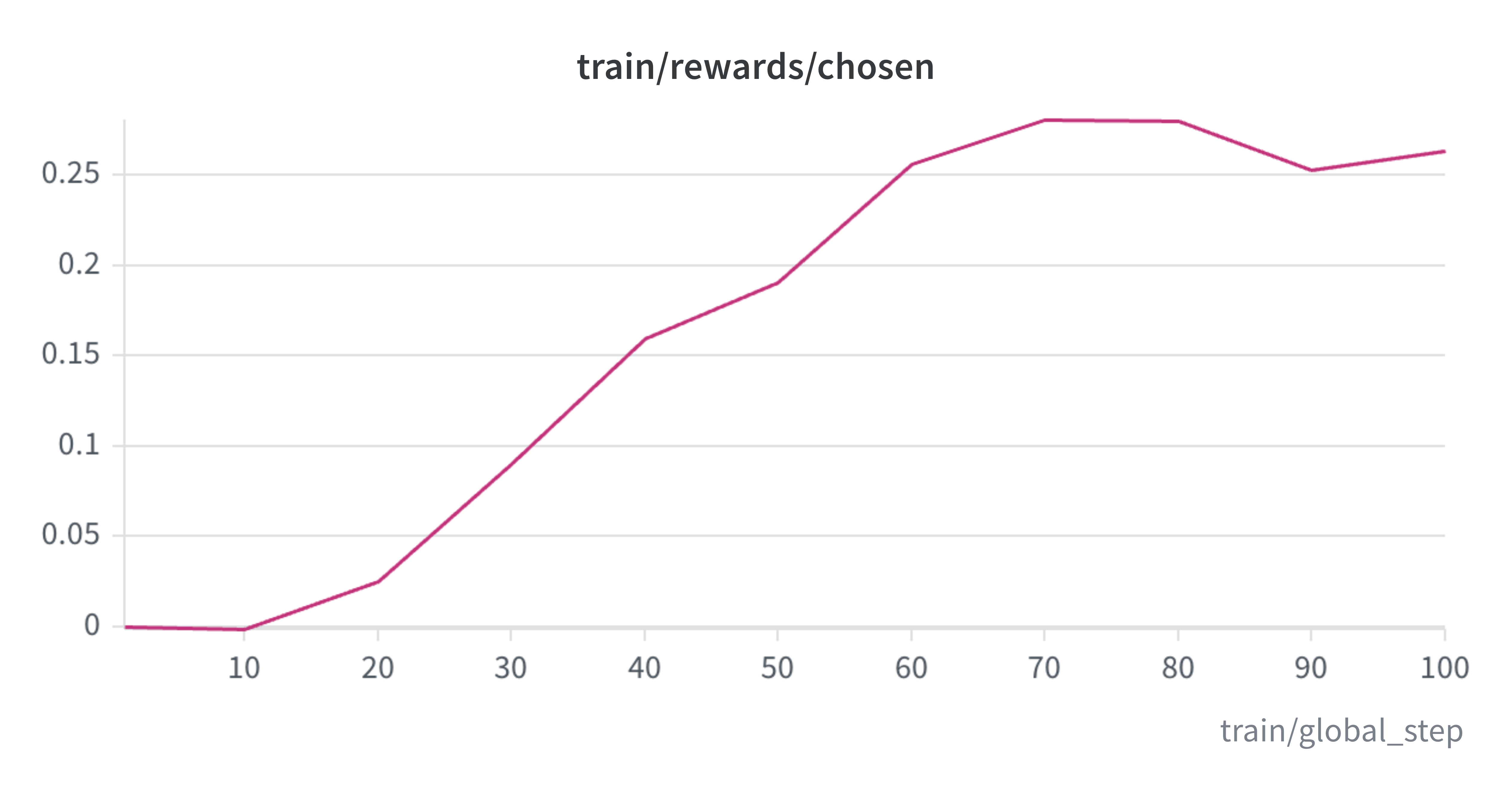}\\[-1mm]
    {\small (c) Chosen reward}
  \end{minipage}\hfill
  \begin{minipage}[t]{0.48\textwidth}
    \centering
    \includegraphics[width=\linewidth,height=0.22\textheight,keepaspectratio]{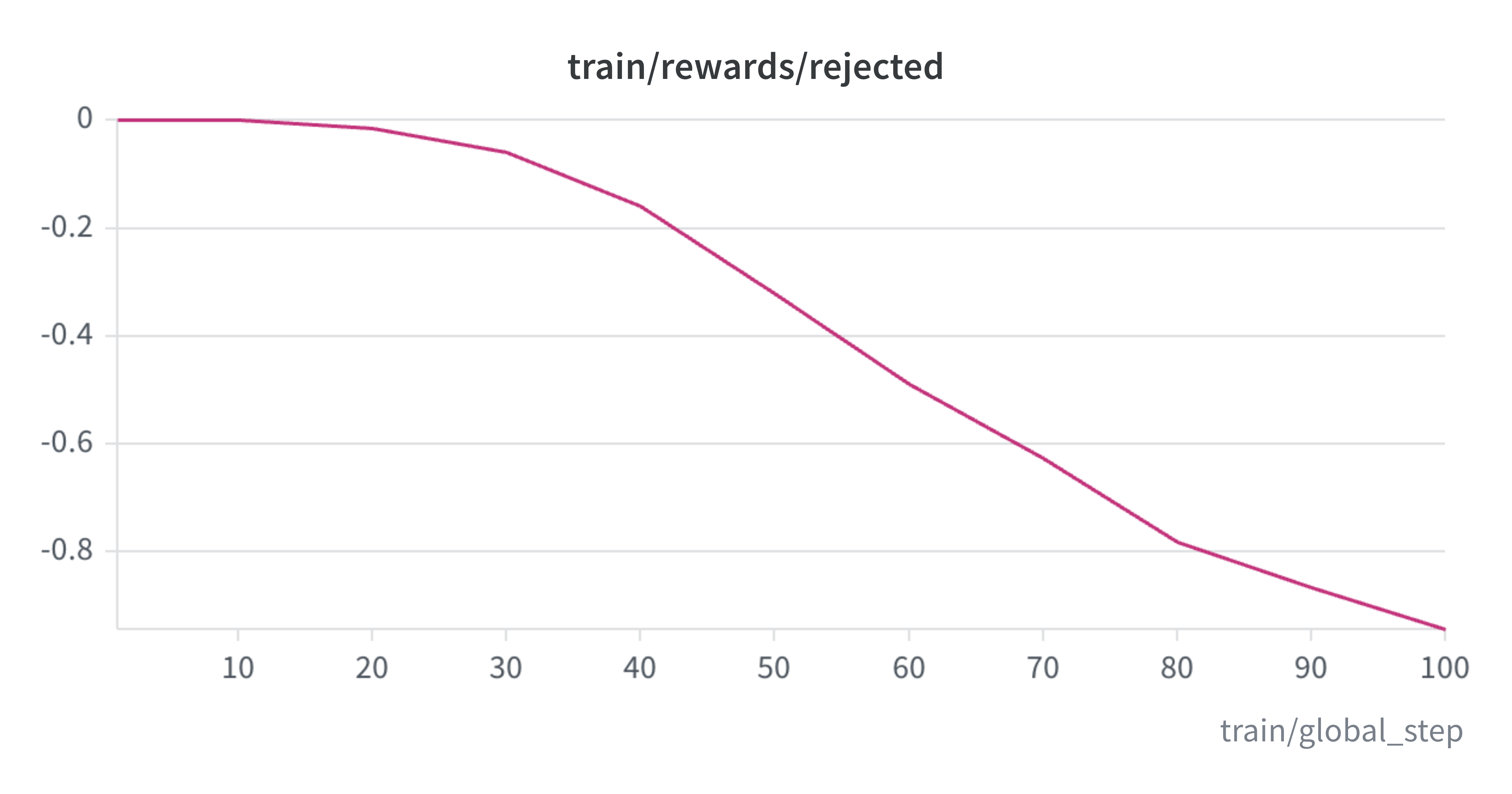}\\[-1mm]
    {\small (d) Rejected reward}
  \end{minipage}
  \caption{DPO optimization traces: (a) the reward margin, (b) training loss, (c) the chosen reward, and (d) the rejected reward.}
  \label{fig:anestrace_eval_dpo_training}
\end{figure*}

\paragraph{Preference alignment.}
Starting from the SFT checkpoint, DPO is applied to approximately 4,000 anesthesiologist-ranked pairs using the settings in Table~\ref{tab:evaluator_training_config}.

The supplied DPO logs are visualized in Figure~\ref{fig:anestrace_eval_dpo_training}.  Across the displayed 100 steps, the chosen reward increases, the rejected reward decreases, the reward margin expands to approximately 1.2, and the DPO loss declines from approximately 0.69 to 0.29.  These trajectories show that the optimization objective separates the constructed preferred and rejected responses. 

\subsection{Agreement with Anesthesiologists}
\label{app:evaluator_validation}

\begin{table*}[t]
\centering
\caption{
Agreement of AnesTRACE-Eval variants with anesthesiologist ratings.
Each cell reports Spearman's $\rho$ / Kendall's $\tau_b$.
Overall aggregates Clinical Correctness, Evidence Grounding,
and Task Completeness. Temporal Consistency applies only to L3.
}
\scriptsize
\renewcommand{\arraystretch}{1.18}
\setlength{\tabcolsep}{3pt}

\begin{tabular*}{\textwidth}{@{\extracolsep{\fill}}llcccccc@{}}
\toprule
\multirow{2}{*}{\textbf{Level}}
& \multirow{2}{*}{\textbf{Evaluator}}
& \multicolumn{6}{c}{\textbf{Agreement with Anesthesiologist Ratings}} \\
\cmidrule(lr){3-8}
& &
\makecell{\textbf{Clinical}\\\textbf{Correctness}} &
\makecell{\textbf{Evidence}\\\textbf{Grounding}} &
\makecell{\textbf{Task}\\\textbf{Completeness}} &
\textbf{Overall} &
\makecell{\textbf{Safety}\\\textbf{Severity}} &
\makecell{\textbf{Temporal}\\\textbf{Consistency}} \\
\midrule

\multirow{3}{*}{L2}
& Base      & .46/.43 & .44/.41 & .39/.32 & .49/.42 & .38/.34 & -- \\
& SFT       & .57/.53 & .48/.45 & .50/.47 & .53/.45 & .48/.43 & -- \\
& SFT + DPO & \textbf{.60/.56} & \textbf{.52/.49}
            & \textbf{.59/.54} & \textbf{.58/.50}
            & \textbf{.51/.47} & -- \\
\midrule
\multirow{3}{*}{L3}
& Base      & .48/.44 & .35/.32 & .45/.38
            & .47/.39 & .47/.41 & .37/.34 \\
& SFT       & .68/.63 & .66/.62 & .46/.42
            & .72/.62 & .77/.70 & .55/.52 \\
& SFT + DPO & \textbf{.71/.66} & \textbf{.69/.65}
            & \textbf{.68/.63} & \textbf{.75/.65}
            & \textbf{.80/.73} & \textbf{.58/.55} \\
\bottomrule
\end{tabular*}
\label{tab:full_evaluator_human_agreement}
\end{table*}
We construct the human-validation set by sampling actual benchmark
responses from five models: Fleming-R1-32B, GPT-OSS-20B,
HuatuoGPT-3-32B, MedGemma-27B-text-it, and Qwen3.5-27B.
These generators are held out from evaluator training: their outputs,
including all sampled validation responses, are excluded from both
SFT and DPO.
The set contains 100 L2 responses and 117 L3 decision-turn responses
from 39 episodes.
Each L2 response includes the four clinical tasks, while L3 responses
are reviewed within their corresponding episode context.
Three anesthesiologists divide the annotation workload, with each
item receiving one annotation under the same clinical rubric.
These ratings provide the human reference for comparing the Base,
SFT, and SFT+DPO evaluators.

Table~\ref{tab:full_evaluator_human_agreement} presents the agreement results from Table~\ref{tab:evaluator_human_agreement} separately for L2 and L3. We quantify agreement using Spearman's rank correlation coefficient~\citep{spearman1987proof} and Kendall's $\tau_b$~\citep{kendall1945treatment}. Both measures generally improve after SFT and increase further after DPO across evaluation dimensions, including overall quality and safety assessment. These results show that AnesTRACE-Eval achieves stronger agreement with anesthesiologist judgments after training, with both SFT and preference optimization contributing to the improvement.

\section{Experimental and Agent Protocols}
\label{app:model_setup}

\subsection{Model Inventory and Runtime Configuration}
\label{app:model_configuration}
For intraoperative perception tasks, we evaluate 16 MLLMs. The API-accessed models are GPT-6-Astra,
Gemini 3 Pro Preview, Claude-Fable-5, and
Qwen3.8-Max~\citep{openai2026gpt6astra,google2025gemini3pro,anthropic2026fable5,alibaba2026qwen38max}.
General-purpose open-weight models comprise
Qwen3-VL-8B/32B~\citep{bai2025qwen3}, InternVL3.5-8B/38B~\citep{wang2025internvl3},
LLaVA-OneVision-2-8B~\citep{li2024llava}, Qwen3.5-9B/27B~\citep{qwen35},
and Qwen3.8-27B~\citep{qwen2026qwen3827b}. The medical-domain MLLMs are
Lingshu-I-8B and Lingshu-32B~\citep{xu2026lingshu}, and Fleming-VL-8B
and Fleming-VL-38B~\citep{shu2025fleming}.

For the decision-making tasks, we evaluate 19 LLMs. The API-accessed models
include GPT-6-Astra, Gemini 3 Pro Preview, Claude-Fable-5, Qwen3.8-Max, and
DeepSeek-V4-Pro~\citep{deepseek2026v4}. General-purpose locally deployed
models include Qwen3-8B~\citep{yang2025qwen3},
Qwen3.5-9B/27B~\citep{qwen35}, Qwen3.8-27B~\citep{qwen2026qwen3827b},
GLM-4.7-Flash~\citep{zai2026glm47flash}, and
GPT-OSS-20B~\citep{agarwal2025gpt}. Medical-domain models include
Fleming-R1-7B/32B~\citep{liu2025fleming},
HuatuoGPT-3-8B/32B~\citep{chen2026onepo},
MedGemma-27B-text-it, Baichuan-M2-32B~\citep{sellergren2025medgemma, dou2025baichuan}, and the anesthesia-specialized
Morpheus-7B/32B~\citep{Anessuite}.

Open-weight checkpoints are downloaded from their official Hugging Face repositories without task-specific adaptation and deployed on NVIDIA H100 GPUs in bfloat16 for inference. Locally deployed models use an OpenAI-compatible vLLM interface and API-accessed models are evaluated through their official API endpoints. We use temperature 0 where supported and fix local
top-$p=1$, top-$k=0$, and seed 42, with explicit thinking disabled for
compatible models.

\subsection{Inference Protocols and Prompt Templates}
\label{app:inference_protocols}
\begin{figure*}[!t]
\centering
\begin{minipage}{0.90\textwidth}
\textbf{L1: Multimodal perception.}
The model receives the original TEE image or video, waveform, or trend plot and
must answer using visible evidence only. The output contract is selected by
task type: option identifier, spatial coordinates, or structured findings.
\end{minipage}

\vspace{0.25em}
\includegraphics[
    width=0.90\textwidth
]{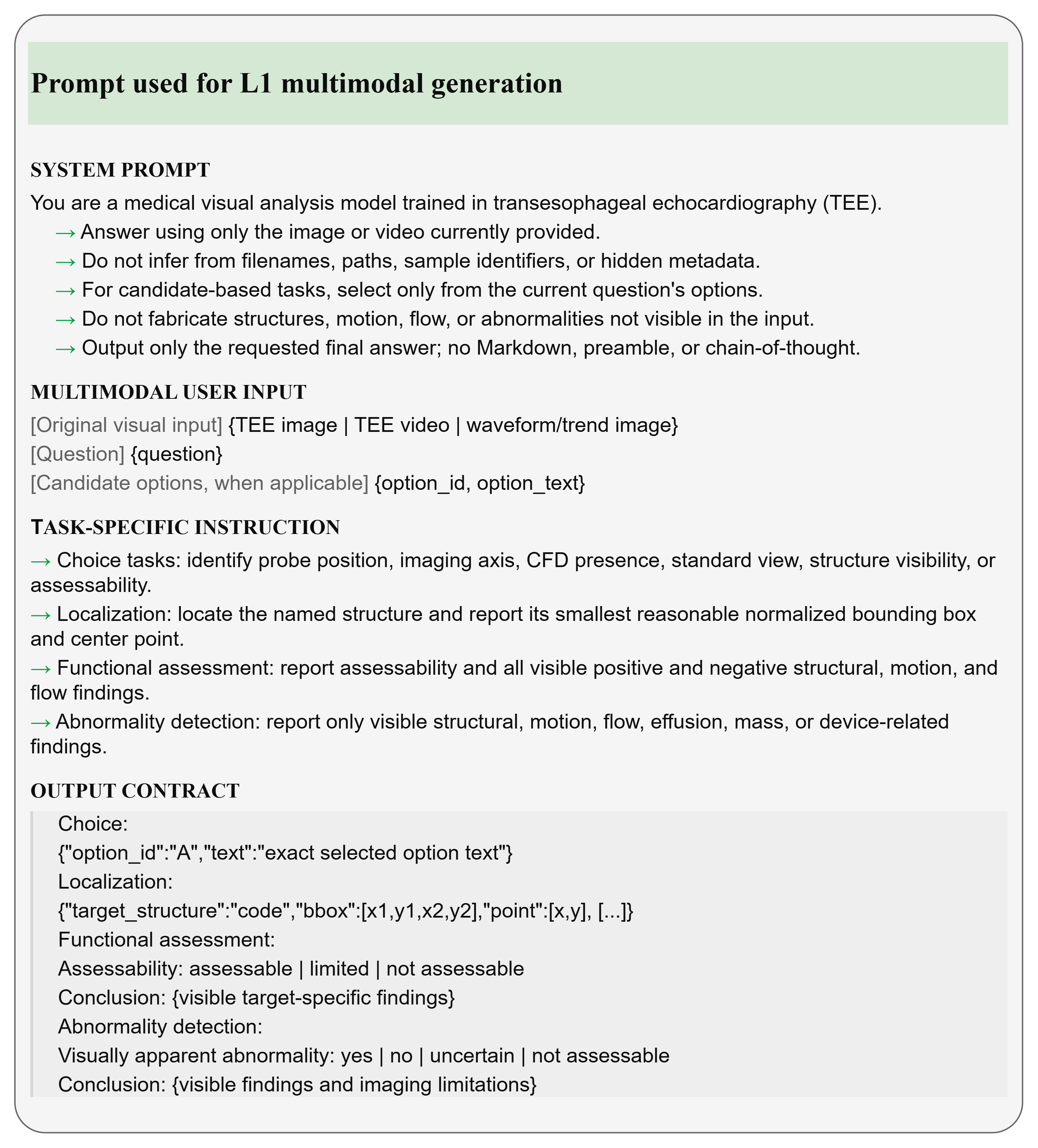}
\vspace{-0.4em}
\caption{Prompt used for L1 multimodal perception.}
\label{fig:prompt_level_one}
\end{figure*}

\begin{figure*}[!t]
\centering
\begin{minipage}{0.90\textwidth}
\textbf{L2: Single-point clinical decision-making.}
The prompt inserts the available patient state at one decision anchor and asks
for one complete plan. Eight ordered sections separate risk, evidence,
diagnosis, management, reassessment, and escalation without requiring JSON.
\end{minipage}

\vspace{0.25em}
\includegraphics[
    width=0.90\textwidth
]{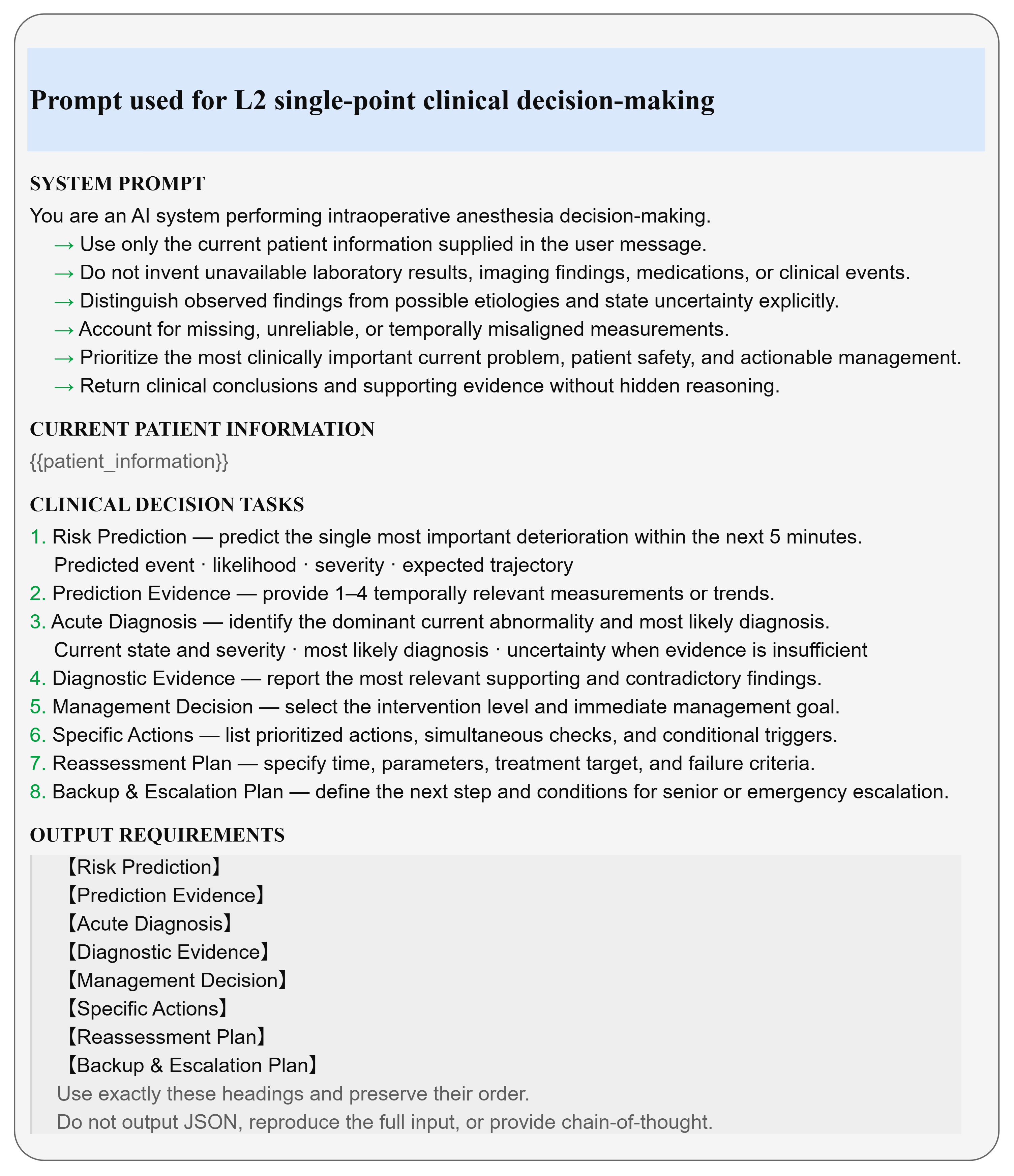}
\vspace{-0.4em}
\caption{Prompt used for L2 single-point clinical decision-making.}
\label{fig:prompt_level_two}
\end{figure*}

\begin{figure*}[!t]
\centering
\begin{minipage}{0.90\textwidth}
\textbf{L3: Tool-using sequential decision-making.}
The initial prompt contains only the current partial observation. Bounded tools
provide permitted context and external knowledge, and each turn ends with a
structured state assessment and intervention in strict JSON.
\end{minipage}

\vspace{0.25em}
\includegraphics[
    width=0.90\textwidth
]{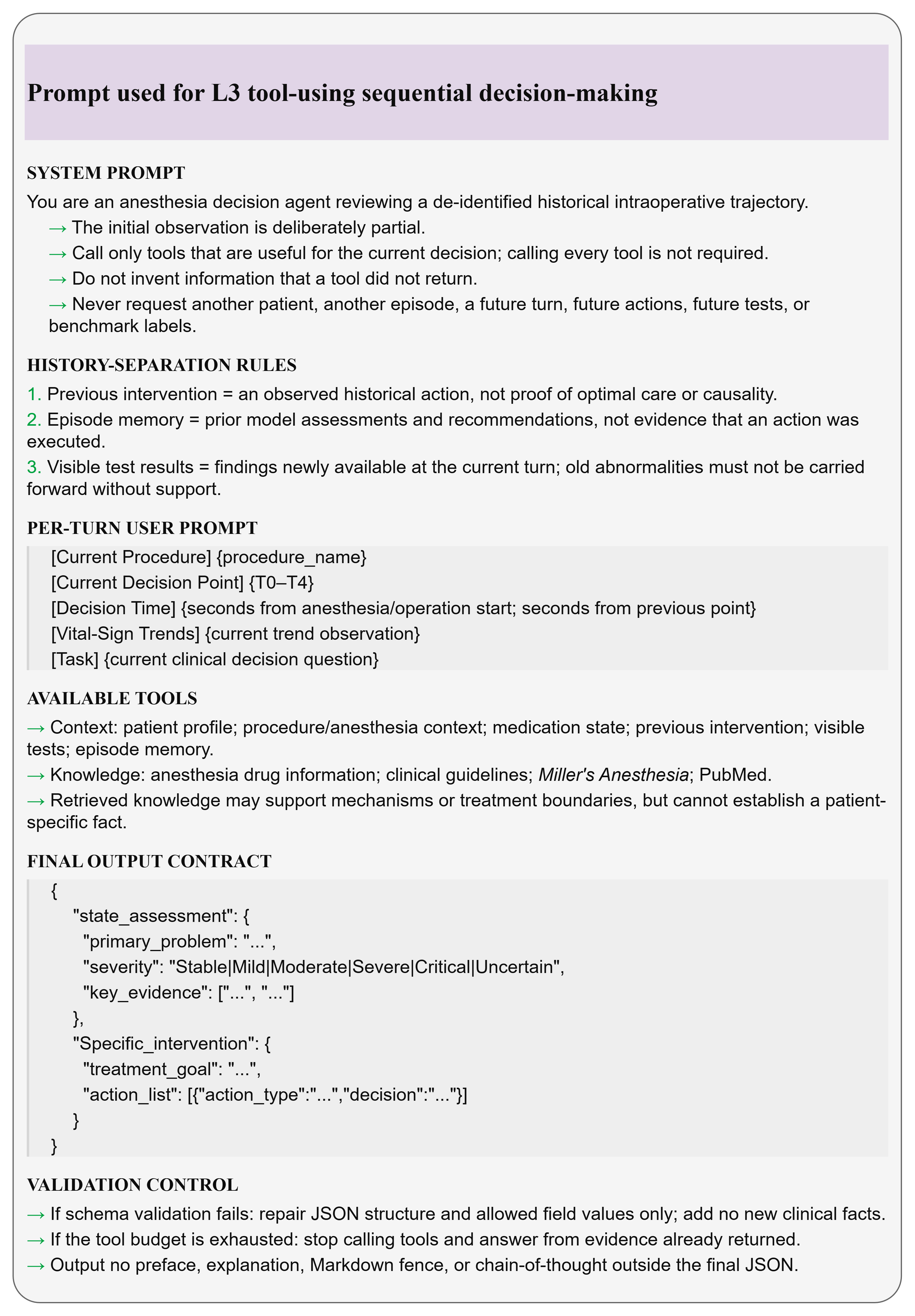}
\vspace{-0.4em}
\caption{Prompt used for L3 tool-using sequential decision-making.}
\label{fig:prompt_level_three}
\end{figure*}

We present the model-facing prompts used at the three benchmark levels in the figures~\ref{fig:prompt_level_one}--\ref{fig:prompt_level_three}. Reference answers, hidden labels, and future information are excluded from every prompt.

\begin{table*}[t]
\centering
\footnotesize
\renewcommand{\arraystretch}{1.18}
\setlength{\tabcolsep}{3pt}
\caption{
Tools available to the intraoperative agent. Patient-specific tools are restricted to the current turn.
}
\begin{tabularx}{\textwidth}{@{}
>{\centering\arraybackslash}m{0.14\textwidth}
>{\raggedright\arraybackslash}p{0.33\textwidth}
>{\raggedright\arraybackslash}p{0.16\textwidth}
>{\raggedright\arraybackslash}X
@{}}
\toprule
\textbf{Tool Type} & \textbf{Tool} & \textbf{Data Source}
& \textbf{Returned Information} \\
\midrule

\multirow[c]{9}{=}{\shortstack{Patient\\context}}
& \makecell[l]{Patient profile\\[-2pt]
  {\scriptsize\texttt{get\_patient\_profile}}}
& Current episode
& Demographics, ASA status, and comorbidities. \\

& \makecell[l]{Procedure and anesthesia context\\[-2pt]
  {\scriptsize\texttt{get\_procedure\_anesthesia\_context}}}
& Current episode
& Procedure, position, airway, and anesthesia context. \\

& \makecell[l]{Medication state\\[-2pt]
  {\scriptsize\texttt{get\_anesthesia\_medication\_state}}}
& Current turn
& Active medications, infusion rates, and device settings. \\

& \makecell[l]{Previous intervention\\[-2pt]
  {\scriptsize\texttt{get\_previous\_intervention}}}
& Previous turn
& The preceding documented clinical intervention. \\

& \makecell[l]{Investigation results\\[-2pt]
  {\scriptsize\texttt{get\_visible\_test\_results}}}
& Current turn
& Newly available laboratory and other test results. \\

\midrule
\multirow[c]{1}{=}{\shortstack{Episode\\memory}}
& \makecell[l]{Episode memory\\[-2pt]
  {\scriptsize\texttt{get\_episode\_memory}}}
& Runtime memory
& Prior agent assessments and recommendations. \\

\midrule
\multirow[c]{8}{=}{\shortstack{External\\knowledge}}
& \makecell[l]{Drug information\\[-2pt]
  {\scriptsize\texttt{get\_anesthesia\_drug\_information}}}
& RxNorm, openFDA
& Drug labels, contraindications, and safety guidance. \\

& \makecell[l]{Clinical guidelines\\[-2pt]
  {\scriptsize\texttt{search\_guidelines}}}
& Local guideline corpus
& Cited guideline passages and source metadata. \\

& \makecell[l]{Miller's Anesthesia\\[-2pt]
  {\scriptsize\texttt{search\_miller\_anesthesia}}}
& Local textbook corpus
& Cited passages with chapter and page information. \\

& \makecell[l]{PubMed literature\\[-2pt]
  {\scriptsize\texttt{search\_pubmed}}}
& NCBI PubMed
& Article metadata and abstracts. \\

\bottomrule
\end{tabularx}
\label{tab:l3_agent_tools}
\vspace{-3mm}
\end{table*}

\subsection{Agent Framework, Memory, and Tools}
\label{app:agent_framework}

We implement L3 as a single-agent LangGraph workflow~\citep{langchain2026langgraph}. One graph invocation handles one decision point: turns are processed chronologically within an episode, while independent episodes can run concurrently. The initial prompt contains the procedure, decision time, and current vital-sign trends. The agent may alternate between model inference and tool execution before submitting a structured assessment and management recommendation. The graph then validates the response, attempts one format repair if needed, updates episode memory, and records the turn result. Only after submission is the next recorded turn revealed; model-proposed actions do not alter the replayed trajectory.

Episode memory carries summaries of the agent's previous diagnoses and recommendations across turns and is reset between episodes. It is retrieved separately from the documented intervention at the preceding decision point. Thus, a prior recommendation is not treated as an executed action, and the agent can revise its assessment when new observations become available.

Table~\ref{tab:l3_agent_tools} lists the ten available tools. Five state-bound tools expose the patient profile, procedural and anesthesia context, current medication state, preceding documented intervention, and newly available test results; a sixth retrieves episode memory. These tools take no patient identifier or time-range arguments and return information scoped to the current episode or turn. Four knowledge tools accept model-generated drug names or queries to retrieve drug information, clinical guidelines, passages from \textit{Miller's Anesthesia}~\citep{miller_anesthesia_2024}, and PubMed literature. Retrieved references can inform the recommendation but do not replace patient-specific observations.
When the provided textual descriptions of visual observations are
insufficient, the agent can invoke a supplementary visual perception
tool backed by Qwen3.5-9B to extract additional information from the
underlying visual inputs. Most evaluated models did not invoke this tool during the experiments.

The default per-turn budget permits at most six tool rounds and eight calls, including no more than six context calls and four knowledge calls. Each context tool can be called once per turn. Rejected calls and tool failures remain visible to the agent, while model requests, tool results, validation outcomes, and resource use are retained in the run traces.

\subsection{Agent Behavior and Efficiency Metrics}
\label{app:agent_metrics}

For L3, we measure agent behavior from execution traces separately from
the clinical quality scores produced by AnesTRACE-Eval. Let $R_i$ denote
the reference-related patient facts available at decision turn $i$, and
let $A_i$ denote the facts recorded as acquired by the agent before its
recommendation. For turns with $|R_i|>0$, Evidence Acquisition (EA) is
reported at two aggregation levels:
\begin{equation}
\mathrm{EA}_{\mathrm{micro}}
=
\frac{\sum_{i\in\mathcal{I}}|A_i\cap R_i|}
     {\sum_{i\in\mathcal{I}}|R_i|},
\qquad
\mathrm{EA}_{\mathrm{macro}}
=
\frac{1}{|\mathcal{I}|}
\sum_{i\in\mathcal{I}}
\frac{|A_i\cap R_i|}{|R_i|},
\label{eq:evidence_acquisition}
\end{equation}
where $\mathcal{I}=\{i:|R_i|>0\}$. Micro EA measures aggregate
reference-fact coverage and gives greater weight to turns with more
reference facts; Macro EA gives each eligible decision turn equal weight.
Because fact acquisition is determined by automatic matching, these
metrics are proxies for information acquisition rather than
physician-validated measures of clinically necessary evidence.

Invalid Tool Call Rate is defined as the proportion of recorded tool calls that fail protocol validation or execution checks; rejections caused by budget limits or duplicate-call constraints are tracked separately. Tool Calls / Turn denotes the mean number of recorded tool calls per decision turn.

P95 Latency denotes the empirical 95th percentile of recorded per-turn latency. For each evaluated model, we replay the full set of 167 L3 episodes, comprising 517 decision turns, once under the same inference configuration. Per-turn latency is computed as the sum of recorded model-request time and tool-execution time for each decision turn, and P95 is calculated over all turns with valid timing records for that model. This metric characterizes the upper tail of per-turn latency across the L3 benchmark rather than latency aggregated over an entire episode. For L2, the metric is computed over question-level response times. Output Tokens denotes the mean number of model-generated tokens per decision turn.

\section{Complete Experiments Results}
\label{app:experiments}

\subsection{Intraoperative Perception Results}
\label{app:l1_results}
We present disaggregates the L1 results in Table~\ref{tab:model_comparison} by TEE and waveform task, reporting the corresponding task-level metrics.

\subsection{Single-point Decision-Making Results}
\label{app:single_point_decision_results}
We present the complete results of the two decision-making levels.
Table~\ref{tab:appendix_l2_task_dimensions}
reports L2 performance for each of the four clinical tasks, disaggregated
into clinical correctness, evidence grounding, and task completeness.
Table~\ref{tab:appendix_l2_mllm_task_dimensions} reports the corresponding
results for MLLMs receiving paired-image inputs. Because the two tables
use different input protocols, their scores are presented separately.

\subsection{Multi-step Decision-Making Results}
\label{app:multi_step_decision_results}

For L3, Table~\ref{tab:appendix_l3_clinical} reports turn-level clinical scores, trajectory-level temporal consistency, and
safety error rates. Table~\ref{tab:appendix_l3_agent_efficiency} complements these clinical outcomes with evidence acquisition, tool-use behavior, latency, and output-token consumption. Together, these tables separate the quality of sequential management recommendations from the information gathering and computational costs of the agent workflow.

\begin{table*}[t]
\centering
\caption{Detailed L1 performance by perception task. All scores are reported as percentages, and higher values indicate better performance. Background colors denote model size:
{\setlength{\fboxsep}{0.8pt}\colorbox{ModelSmallBase!12}{\small\strut Small}} ($<10$B),
{\setlength{\fboxsep}{0.8pt}\colorbox{ModelMediumBase!12}{\small\strut Medium}} ($10$--$40$B), and
{\setlength{\fboxsep}{0.8pt}\colorbox{ModelLargeBase!14}{\small\strut Large}} ($>40$B).
}
\renewcommand{\arraystretch}{1.25}
\setlength{\tabcolsep}{3pt}
\resizebox{\textwidth}{!}{%
\begin{tabular}{ll*{14}{c}}
\toprule
& \multirow{2}{*}{\textbf{Model}}
& \multicolumn{11}{c}{\textbf{TEE}}
& \multicolumn{3}{c}{\textbf{Waveform}} \\
\cmidrule(lr){3-13}\cmidrule(lr){14-16}
& & \makecell{Probe\\ACC}
& \makecell{Axis\\ACC}
& \makecell{CFD\\ACC}
& \makecell{Plane\\ACC}
& \makecell{Visibility\\ACC}
& \makecell{Localization\\mIoU}
& \makecell{Assessability\\ACC}
& \makecell{Function\\ACC}
& \makecell{Function\\Factual-F1}
& \makecell{Abnormality\\ACC}
& \makecell{Abnormality\\Factual-F1}
& \makecell{Value\\ACC}
& \makecell{Trend\\Set-F1}
& \makecell{Anomaly\\Metric-F1} \\
\midrule
\multirow[c]{3}{*}{\rotatebox[origin=c]{90}{\scriptsize\makecell{\textbf{Proprietary}\\\textbf{MLLMs}}}}
& \appmodellarge{GPT-6-Astra}
& 58.5 & 70.9 & 93.3 & 72.7 & 62.9 & 32.2 & 68.8 & 30.1 & 13.0 & 23.2 & 15.7
& 91.5 & 61.1 & 64.8 \\
& \appmodellarge{Gemini 3 Pro Preview}
& 69.5 & 67.1 & 93.3 & 56.6 & 46.1 & 24.6 & 46.3 & 27.7 & 13.5 & 29.3 & 12.7
& 77.0 & 62.6 & 54.8 \\
& \appmodellarge{Claude-Fable-5}
& 70.7 & 75.9 & 93.3 & 68.4 & 64.2
& 23.4 & 42.7
& 40.1 & 27.8
& 35.5 & 26.0
& 80.5 & 68.7 & 59.6 \\
\midrule
\multirow[c]{9}{*}{\rotatebox[origin=c]{90}{\scriptsize\makecell{\textbf{General}\\\textbf{Open-Weight}\\\textbf{MLLMs}}}}
& \appmodellarge{Qwen3.8-MAX}
& 68.3 & 64.6 & 93.3 & 66.7 & 51.7 & 22.8 & 62.5 & 57.8 & 12.5 & 24.4 & 10.1
& 72.0 & 69.6 & 42.8 \\
& \appmodelmedium{InternVL3.5-38B}
& 61.0 & 27.8 & 93.3 & 32.3 & 37.1 & 16.8 & 47.5 & 47.0 & 21.1 & 24.4 & 10.1
& 57.0 & 69.6 & 23.0 \\
& \appmodelmedium{Qwen3-VL-32B}
& 52.4 & 49.7 & 93.3 & 46.8 & 32.3
& 18.2 & 23.6
& 45.9 & 7.6
& 41.5 & 6.4
& 55.0 & 71.7 & 28.6 \\
& \appmodelmedium{Qwen3.5-27B}
& 36.6 & 59.5 & 93.3 & 44.4 & 43.8 & 18.9 & 43.8 & 56.6 & 16.8 & 29.3 & 10.7
& 56.0 & 65.3 & 36.5 \\
& \appmodelmedium{Qwen3.8-27B}
& 58.5 & 48.1 & 93.3 & 34.3 & 33.7 & 14.8 & 42.5 & 51.8 & 9.3 & 39.0 & 14.2
& 57.5 & 72.5 & 31.3 \\
& \appmodelsmall{Qwen3.5-9B}
& 59.8 & 30.4 & 93.3 & 33.3 & 52.8 & 15.5 & 40.0 & 57.8 & 8.8 & 28.0 & 6.2
& 54.5 & 60.7 & 33.1 \\
& \appmodelsmall{Qwen3-VL-8B}
& 58.5 & 29.1 & 90.0 & 28.3 & 23.6 & 14.4 & 45.0 & 37.3 & 4.6 & 24.4 & 5.1
& 45.0 & 58.2 & 23.5 \\
& \appmodelsmall{InternVL3.5-8B}
& 59.8 & 31.6 & 93.3 & 31.3 & 48.3 & 13.9 & 41.2 & 33.7 & 10.5 & 29.3 & 6.2
& 41.0 & 70.1 & 17.3 \\
& \appmodelsmall{LLaVA-OneVision-2-8B}
& 54.9 & 30.4 & 90.0 & 24.2 & 50.6 & 14.3 & 32.5 & 47.0 & 1.6 & 31.7 & 4.9
& 49.5 & 60.6 & 23.6 \\
\midrule
\multirow[c]{4}{*}{\rotatebox[origin=c]{90}{\tiny\makecell{\textbf{Medical}\\\textbf{Domain}\\\textbf{Open-Weight}\\\textbf{MLLMs}}}}
& \appmodelmedium{Fleming-VL-38B}
& 46.3 & 19.0 & 93.3 & 42.4 & 37.1 & 14.8 & 36.2 & 28.9 & 8.3 & 19.5 & 1.2
& 41.5 & 59.4 & 24.0 \\
& \appmodelmedium{Lingshu-32B}
& 51.2 & 39.2 & 93.3 & 36.4 & 46.1 & 12.1 & 41.2 & 33.7 & 9.8 & 30.5 & 12.0
& 49.5 & 64.6 & 20.0 \\
& \appmodelsmall{Lingshu-I-8B}
& 58.5 & 49.4 & 76.7 & 29.3 & 56.2 & 13.5 & 33.8 & 41.0 & 4.1 & 25.6 & 11.7
& 39.0 & 59.5 & 23.3 \\
& \appmodelsmall{Fleming-VL-8B}
& 34.1 & 32.9 & 83.3 & 38.4 & 46.1 & 8.7 & 2.5 & 24.1 & 4.7 & 12.2 & 0.0
& 38.5 & 41.1 & 17.6 \\
\bottomrule
\end{tabular}%
}
\label{tab:app_l1_complete}
\vspace{-3mm}
\end{table*}

\begin{table*}[t]
\centering
\caption{
Detailed L2 performance by task and evaluation dimension.
Clinical scores and Safety Error are reported as percentages; P95 Latency is reported in seconds. Background colors denote model size:
{\setlength{\fboxsep}{0.8pt}\colorbox{ModelSmallBase!12}{\small\strut Small}} ($<10$B),
{\setlength{\fboxsep}{0.8pt}\colorbox{ModelMediumBase!12}{\small\strut Medium}} ($10$--$40$B), and
{\setlength{\fboxsep}{0.8pt}\colorbox{ModelLargeBase!14}{\small\strut Large}} ($>40$B).
}
\renewcommand{\arraystretch}{1.25}
\setlength{\tabcolsep}{1pt}
\resizebox{\textwidth}{!}{%
\begin{tabular}{@{}c@{\hspace{1pt}}l*{14}{c}@{}}
\toprule
& \multirow{2}{*}{\textbf{Model}}
& \multicolumn{3}{c}{\textbf{Risk Prediction}}
& \multicolumn{3}{c}{\textbf{Clinical Diagnosis}}
& \multicolumn{3}{c}{\textbf{Intervention Decision}}
& \multicolumn{3}{c}{\textbf{Reassessment Plan}}
& \multirow{2}{*}{\makecell{\textbf{Safety}\\\textbf{Error}$\downarrow$}}
& \multirow{2}{*}{\makecell{\textbf{P95}\\\textbf{Latency (s)}$\downarrow$}} \\
\cmidrule(lr){3-5}
\cmidrule(lr){6-8}
\cmidrule(lr){9-11}
\cmidrule(lr){12-14}
& & \textbf{Corr.}$\uparrow$
& \textbf{Ground.}$\uparrow$
& \textbf{Complete.}$\uparrow$
& \textbf{Corr.}$\uparrow$
& \textbf{Ground.}$\uparrow$
& \textbf{Complete.}$\uparrow$
& \textbf{Corr.}$\uparrow$
& \textbf{Ground.}$\uparrow$
& \textbf{Complete.}$\uparrow$
& \textbf{Corr.}$\uparrow$
& \textbf{Ground.}$\uparrow$
& \textbf{Complete.}$\uparrow$
& & \\
\midrule

\multirow[c]{3}{*}{\rotatebox[origin=c]{90}{\scriptsize\makecell{\textbf{Proprietary}\\\textbf{LLMs}}}}
& \appmodellarge{GPT-6-Astra}
& 76.2 & 66.3 & 100.0
& 91.5 & 89.9 & 99.7
& 93.9 & 94.2 & 99.9
& 97.8 & 97.6 & 99.7
& 3.4 & 33.408 \\

& \appmodellarge{Gemini 3 Pro Preview}
& 64.6 & 51.8 & 97.0
& 66.9 & 60.4 & 96.7
& 58.6 & 57.2 & 90.8
& 81.5 & 81.5 & 95.3
& 13.5 & 26.946 \\

& \appmodellarge{Claude-Fable-5}
& 86.2 & 76.2 & 98.8
& 90.1 & 86.8 & 98.8
& 82.5 & 82.0 & 98.8
& 95.5 & 95.9 & 97.4
& 9.4 & 39.964 \\

\midrule
\multirow[c]{8}{*}{\rotatebox[origin=c]{90}{\small\makecell{\textbf{General}\\\textbf{Open-Weight}\\\textbf{LLMs}}}}
& \appmodellarge{DeepSeek-V4-Pro}
& 72.9 & 61.6 & 100.0
& 78.7 & 72.1 & 99.7
& 71.7 & 69.1 & 96.1
& 91.9 & 91.7 & 97.8
& 14.8 & 14.040 \\

& \appmodellarge{Qwen3.8-Max}
& 61.9 & 51.7 & 98.4
& 73.0 & 67.5 & 98.2
& 61.7 & 60.5 & 92.7
& 83.5 & 83.2 & 95.9
& 15.1 & 23.514 \\

& \appmodelmedium{Qwen3.5-27B}
& 59.0 & 49.3 & 100.0
& 65.2 & 58.8 & 99.7
& 61.4 & 59.8 & 89.3
& 70.1 & 69.1 & 94.5
& 28.6 & 22.139 \\

& \appmodelmedium{Qwen3.8-27B}
& 63.1 & 51.1 & 100.0
& 66.2 & 60.8 & 99.6
& 54.6 & 52.4 & 91.2
& 67.4 & 66.7 & 90.8
& 21.5 & 22.935 \\

& \appmodelmedium{GPT-OSS-20B}
& 60.3 & 47.1 & 100.0
& 52.1 & 45.7 & 99.5
& 31.4 & 29.1 & 91.7
& 54.3 & 52.9 & 91.2
& 37.5 & 17.904 \\

& \appmodelmedium{GLM-4.7-Flash}
& 41.9 & 33.8 & 99.9
& 40.7 & 36.1 & 98.9
& 26.5 & 23.9 & 87.8
& 49.3 & 47.4 & 82.5
& 48.2 & 36.615 \\

& \appmodelsmall{Qwen3.5-9B}
& 53.7 & 42.1 & 100.0
& 58.7 & 51.5 & 99.7
& 41.8 & 39.2 & 90.1
& 63.6 & 63.1 & 89.1
& 27.7 & 15.181 \\

& \appmodelsmall{Qwen3-8B}
& 40.9 & 33.6 & 100.0
& 31.2 & 26.5 & 99.0
& 18.5 & 14.8 & 79.4
& 40.1 & 39.1 & 73.1
& 57.6 & 24.501 \\

\midrule
\multirow[c]{8}{*}{\rotatebox[origin=c]{90}{\scriptsize\makecell{\textbf{Medical}\\\textbf{Domain}\\\textbf{Open-Weight}\\\textbf{LLMs}}}}
& \appmodelmedium{Morpheus-32B}
& 38.1 & 27.0 & 100.0
& 27.5 & 23.8 & 99.9
& 16.0 & 14.0 & 95.1
& 43.4 & 41.4 & 89.1
& 56.9 & 20.305 \\

& \appmodelmedium{Fleming-R1-32B}
& 51.5 & 40.2 & 100.0
& 48.9 & 43.1 & 99.7
& 30.3 & 27.0 & 82.0
& 56.8 & 56.8 & 87.4
& 37.6 & 14.551 \\

& \appmodelmedium{HuatuoGPT-3-32B}
& 49.7 & 38.4 & 100.0
& 44.4 & 37.8 & 100.0
& 25.7 & 23.6 & 93.6
& 41.7 & 41.4 & 95.9
& 50.9 & 25.049 \\

& \appmodelmedium{Baichuan-M2-32B}
& 41.0 & 32.8 & 100.0
& 46.5 & 40.2 & 99.2
& 28.7 & 25.4 & 90.5
& 36.8 & 34.7 & 83.0
& 42.1 & 75.099 \\

& \appmodelmedium{MedGemma-27B-text-it}
& 52.7 & 41.6 & 99.8
& 45.3 & 39.4 & 98.8
& 25.1 & 22.0 & 77.3
& 52.4 & 50.2 & 87.3
& 43.7 & 12.443 \\

& \appmodelsmall{HuatuoGPT-3-8B}
& 18.6 & 14.4 & 100.0
& 9.3 & 7.0 & 100.0
& 4.8 & 4.5 & 97.6
& 19.0 & 19.4 & 98.2
& 85.9 & 9.874 \\

& \appmodelsmall{Morpheus-7B}
& 23.4 & 17.7 & 99.9
& 14.9 & 12.0 & 99.9
& 10.9 & 9.6 & 96.4
& 32.6 & 30.9 & 87.7
& 67.4 & 26.802 \\

& \appmodelsmall{Fleming-R1-7B}
& 34.9 & 30.2 & 99.9
& 32.2 & 25.8 & 99.5
& 13.6 & 12.2 & 95.6
& 41.9 & 40.2 & 83.6
& 59.0 & 9.193 \\

\bottomrule
\end{tabular}%
}
\vspace{-3mm}
\label{tab:appendix_l2_task_dimensions}
\end{table*}

\begin{table*}[t]
\centering
\caption{
Detailed L2 performance for MLLMs with paired-image input.
Clinical scores and Safety Error are reported as percentages; P95 Latency is reported in seconds. Background colors denote model size:
{\setlength{\fboxsep}{0.8pt}\colorbox{ModelSmallBase!12}{\small\strut Small}} ($<10$B),
{\setlength{\fboxsep}{0.8pt}\colorbox{ModelMediumBase!12}{\small\strut Medium}} ($10$--$40$B), and
{\setlength{\fboxsep}{0.8pt}\colorbox{ModelLargeBase!14}{\small\strut Large}} ($>40$B).
}
\renewcommand{\arraystretch}{1.25}
\setlength{\tabcolsep}{1pt}
\resizebox{\textwidth}{!}{%
\begin{tabular}{@{}c@{\hspace{1pt}}l*{14}{c}@{}}
\toprule
& \multirow{2}{*}{\textbf{Model}}
& \multicolumn{3}{c}{\textbf{Risk Prediction}}
& \multicolumn{3}{c}{\textbf{Clinical Diagnosis}}
& \multicolumn{3}{c}{\textbf{Intervention Decision}}
& \multicolumn{3}{c}{\textbf{Reassessment Plan}}
& \multirow{2}{*}{\makecell{\textbf{Safety}\\\textbf{Error}$\downarrow$}}
& \multirow{2}{*}{\makecell{\textbf{P95}\\\textbf{Latency (s)}$\downarrow$}} \\
\cmidrule(lr){3-5}
\cmidrule(lr){6-8}
\cmidrule(lr){9-11}
\cmidrule(lr){12-14}
& & \textbf{Corr.}$\uparrow$
& \textbf{Ground.}$\uparrow$
& \textbf{Complete.}$\uparrow$
& \textbf{Corr.}$\uparrow$
& \textbf{Ground.}$\uparrow$
& \textbf{Complete.}$\uparrow$
& \textbf{Corr.}$\uparrow$
& \textbf{Ground.}$\uparrow$
& \textbf{Complete.}$\uparrow$
& \textbf{Corr.}$\uparrow$
& \textbf{Ground.}$\uparrow$
& \textbf{Complete.}$\uparrow$
& & \\
\midrule

\multirow[c]{5}{*}{\rotatebox[origin=c]{90}{\scriptsize\makecell{\textbf{General}\\\textbf{Open-Weight}\\\textbf{MLLMs}}}}
& \appmodelmedium{Qwen3-VL-32B-Instruct}
& 19.5 & 9.9 & 100.0
& 19.7 & 12.0 & 99.9
& 38.6 & 32.8 & 85.6
& 51.4 & 50.0 & 92.4
& 40.9 & 48.546 \\

& \appmodelmedium{Qwen3.8-27B}
& 17.0 & 8.3 & 99.5
& 21.3 & 13.9 & 99.4
& 40.9 & 36.2 & 85.9
& 52.3 & 50.6 & 83.7
& 36.2 & 18.318 \\

& \appmodelsmall{Qwen3.5-9B}
& 11.7 & 6.0 & 99.8
& 15.0 & 7.8 & 99.6
& 34.8 & 29.8 & 81.0
& 47.2 & 45.8 & 77.6
& 42.7 & 10.098 \\

& \appmodelsmall{Qwen3-VL-8B-Instruct}
& 13.5 & 6.3 & 100.0
& 12.1 & 5.0 & 99.5
& 27.3 & 23.6 & 76.8
& 43.3 & 43.0 & 75.2
& 50.3 & 14.260 \\

& \appmodelsmall{InternVL3.5-8B}
& 8.8 & 4.9 & 100.0
& 9.1 & 6.0 & 99.0
& 12.9 & 10.1 & 84.6
& 26.1 & 25.5 & 50.6
& 69.5 & 14.967 \\

\midrule
\multirow[c]{3}{*}{\rotatebox[origin=c]{90}{\tiny\makecell{\textbf{Medical}\\\textbf{Domain}\\\textbf{Open-Weight}\\\textbf{MLLMs}}}}
& \appmodelmedium{Lingshu-32B}
& 13.2 & 10.6 & 99.9
& 15.2 & 11.7 & 99.3
& 16.6 & 13.7 & 78.0
& 33.5 & 32.9 & 52.7
& 66.8 & 26.080 \\

& \appmodelsmall{Lingshu-I-8B}
& 5.0 & 3.8 & 100.0
& 5.6 & 3.0 & 98.7
& 4.0 & 2.7 & 90.0
& 13.7 & 13.5 & 49.0
& 88.4 & 6.597 \\

& \appmodelsmall{Fleming-VL-8B}
& 11.6 & 7.8 & 59.4
& 18.1 & 14.8 & 60.4
& 19.3 & 18.6 & 54.2
& 29.9 & 29.5 & 41.2
& 66.8 & 17.099 \\

\bottomrule
\end{tabular}%
}
\vspace{-3mm}
\label{tab:appendix_l2_mllm_task_dimensions}
\end{table*}

\begin{table*}[t]
\centering
\caption{
Detaidled L3 clinical decision performance. Clinical scores and Safety Error are reported as percentages. The three clinical dimensions average the corresponding diagnosis and intervention scores. Background colors denote model size:
{\setlength{\fboxsep}{0.8pt}\colorbox{ModelSmallBase!12}{\small\strut Small}} ($<10$B),
{\setlength{\fboxsep}{0.8pt}\colorbox{ModelMediumBase!12}{\small\strut Medium}} ($10$--$40$B), and
{\setlength{\fboxsep}{0.8pt}\colorbox{ModelLargeBase!14}{\small\strut Large}} ($>40$B).
}
\renewcommand{\arraystretch}{1.25}
\setlength{\tabcolsep}{4.5pt}
\resizebox{\textwidth}{!}{%
\begin{tabular}{ll*{6}{c}}
\toprule
& \multirow{2}{*}{\textbf{Model}}
& \multicolumn{4}{c}{\textbf{Turn-level Clinical Scores}}
& \multirow{2}{*}{\makecell{\textbf{Temporal}\\\textbf{Consistency}$\uparrow$}}
& \multirow{2}{*}{\makecell{\textbf{Safety}\\\textbf{Error}$\downarrow$}} \\
\cmidrule(lr){3-6}
& & \makecell{\textbf{Clinical}\\\textbf{Correctness}$\uparrow$}
& \makecell{\textbf{Evidence}\\\textbf{Grounding}$\uparrow$}
& \makecell{\textbf{Task}\\\textbf{Completeness}$\uparrow$}
& \textbf{Avg.}$\uparrow$
& & \\
\midrule
\multirow{4}{*}{\rotatebox[origin=c]{90}{\small\makecell{\textbf{Proprietary}\\\textbf{LLMs}}}}
& \appmodellarge{GPT-6-Astra}          & 80.65 & 80.54 & 86.92 & 82.70 & 87.6 & 17.54 \\
& \appmodellarge{Gemini 3 Pro Preview} & 63.23 & 60.70 & 74.31 & 66.08 & 78.01 & 38.52 \\
& \appmodellarge{Claude-Fable-5}       & 76.07 & 79.27 & 82.54 & 79.40 & 82.5 & 24.3 \\

\midrule
\multirow{7}{*}{\rotatebox[origin=c]{90}{\small\makecell{\textbf{General}\\\textbf{LLMs}}}}
& \appmodellarge{DeepSeek-V4-Pro}      & 77.52 & 74.85 & 87.65 & 80.01 & 81.92 & 19.17 \\
& \appmodellarge{Qwen3.8-Max}          & 64.63 & 58.41 & 78.34 & 67.13 & 85.33 & 40.09 \\
& \appmodelmedium{Qwen3.5-27B}         & 66.16 & 61.80 & 80.61 & 69.52 & 81.59 & 32.50 \\
& \appmodelmedium{Qwen3.8-27B}         & 62.81 & 60.52 & 74.91 & 66.08 & 69.16 & 34.05 \\
& \appmodelmedium{GPT-OSS-20B}         & 53.12 & 49.13 & 68.91 & 57.05 & 67.22 & 53.41 \\
& \appmodelmedium{GLM-4.7-Flash}       & 33.33 & 29.12 & 69.70 & 44.05 & 49.85 & 70.78 \\
& \appmodelsmall{Qwen3.5-9B}           & 50.92 & 45.63 & 68.69 & 55.08 & 64.22 & 61.94 \\
& \appmodelsmall{Qwen3-8B}             & 23.05 & 19.54 & 48.02 & 30.23 & 45.21 & 64.05 \\
\midrule
\multirow{8}{*}{\rotatebox[origin=c]{90}{\scriptsize\makecell{\textbf{Medical-Domain}\\\textbf{Open-Weight}\\\textbf{LLMs}}}}
& \appmodelmedium{Morpheus-32B}        & 36.86 & 29.98 & 71.78 & 46.21 & 55.99 & 66.02 \\
& \appmodelmedium{Fleming-R1-32B}      & 48.77 & 43.53 & 66.20 & 52.83 & 66.87 & 56.52 \\
& \appmodelmedium{HuatuoGPT-3-32B}     & 43.25 & 37.92 & 68.59 & 49.92 & 61.68 & 64.69 \\
& \appmodelmedium{Baichuan-M2-32B}     & 40.40 & 36.94 & 71.20 & 49.51 & 56.14 & 66.67 \\
& \appmodelmedium{MedGemma-27B-text-it} & 45.14 & 41.67 & 59.41 & 48.74 & 54.34 & 59.41 \\
& \appmodelsmall{HuatuoGPT-3-8B}       & 19.02 & 14.16 & 68.19 & 33.79 & 48.20 & 86.58 \\
& \appmodelsmall{Morpheus-7B}          & 23.72 & 17.58 & 64.05 & 35.12 & 46.69 & 78.19 \\
& \appmodelsmall{Fleming-R1-7B}        & 36.76 & 31.92 & 64.67 & 44.45 & 56.59 & 69.37 \\
\bottomrule
\end{tabular}%
}
\label{tab:appendix_l3_clinical}
\end{table*}

\begin{table*}[t]
\centering
\caption{
Detailed L3 agent behavior and efficiency. Evidence Acquisition and Invalid Call Rate are percentages. Evidence Acquisition automatically matches acquired
reference-related patient facts and remains exploratory pending source-data version reconciliation. Output Tokens is the mean generated token count per turn. Background colors denote model size:
{\setlength{\fboxsep}{0.8pt}\colorbox{ModelSmallBase!12}{\small\strut Small}} ($<10$B),
{\setlength{\fboxsep}{0.8pt}\colorbox{ModelMediumBase!12}{\small\strut Medium}} ($10$--$40$B), and
{\setlength{\fboxsep}{0.8pt}\colorbox{ModelLargeBase!14}{\small\strut Large}} ($>40$B).
}
\renewcommand{\arraystretch}{1.25}
\setlength{\tabcolsep}{4.5pt}
\resizebox{\textwidth}{!}{%
\begin{tabular}{ll*{6}{c}}
\toprule
& \multirow{2}{*}{\textbf{Model}}
& \multicolumn{2}{c}{\textbf{Evidence Acquisition}}
& \multicolumn{2}{c}{\textbf{Tool Use}}
& \multicolumn{2}{c}{\textbf{Efficiency}} \\
\cmidrule(lr){3-4}
\cmidrule(lr){5-6}
\cmidrule(lr){7-8}
& & \textbf{Micro}$\uparrow$
& \textbf{Macro}$\uparrow$
& \makecell{\textbf{Invalid Call}\\\textbf{Rate}$\downarrow$}
& \makecell{\textbf{Calls}\\\textbf{/ Turn}}
& \makecell{\textbf{P95}\\\textbf{Latency (s)}$\downarrow$}
& \makecell{\textbf{Output}\\\textbf{Tokens}} \\
\midrule
\multirow{5}{*}{\rotatebox[origin=c]{90}{\small\makecell{\textbf{Proprietary}\\\textbf{LLMs}}}}
& \appmodellarge{GPT-6-Astra}          & 61.84 & 62.51 & 0.00 & 2.64 & 74.29  & 893.23 \\
& \appmodellarge{Gemini 3 Pro Preview} & 74.44 & 76.36 & 0.00 & 3.30 & 101.37 & 2718.95 \\
& \appmodellarge{Claude-Fable-5}       & 84.44 & 82.66 & 0.00 & 3.53 & 69.15  & 2944.37 \\

\midrule
\multirow{6}{*}{\rotatebox[origin=c]{90}{\small\makecell{\textbf{General}\\\textbf{Open-Weight LLMs}}}}
& \appmodellarge{DeepSeek-V4-Pro}      & 89.74 & 87.88 & 0.05 & 4.03 & 14.41  & 508.54 \\
& \appmodellarge{Qwen3.8-Max}          & 62.84 & 61.95 & 0.00 & 2.07 & 22.87  & 674.32 \\
& \appmodelmedium{Qwen3.5-27B}         & 83.30 & 84.13 & 0.00 & 4.00 & 34.38  & 535.21 \\
& \appmodelmedium{Qwen3.8-27B}         & 86.92 & 85.57 & 0.00 & 4.01 & 39.88  & 567.01 \\
& \appmodelmedium{GPT-OSS-20B}         & 59.49 & 60.97 & 0.00 & 2.86 & 106.55 & 1169.30 \\
& \appmodelmedium{GLM-4.7-Flash}       & 95.44 & 97.18 & 0.00 & 7.03 & 63.60  & 554.91 \\
& \appmodelsmall{Qwen3.5-9B}           & 89.54 & 90.93 & 0.00 & 3.63 & 27.22  & 792.89 \\
& \appmodelsmall{Qwen3-8B}             & 32.7  & 34.8  & 0.00 & 0.71 & 10.29  & 539.72 \\
\midrule
\multirow{8}{*}{\rotatebox[origin=c]{90}{\scriptsize\makecell{\textbf{Medical-Domain}\\\textbf{Open-Weight}\\\textbf{LLMs}}}}
& \appmodelmedium{Morpheus-32B}        & 67.54 & 68.76 & 0.00 & 3.72 & 86.91 & 7118.69 \\
& \appmodelmedium{Fleming-R1-32B}      & 11.33 & 15.11 & 0.00 & 1.76 & 22.80  & 521.56 \\
& \appmodelmedium{HuatuoGPT-3-32B}     & 73.44 & 74.00 & 0.00 & 4.67 & 31.30  & 716.44 \\
& \appmodelmedium{Baichuan-M2-32B}     & 95.98 & 96.63 & 0.00 & 7.01 & 32.98  & 558.96 \\
& \appmodelmedium{MedGemma-27B-text-it} & 40.49  & 42.96  & 0.00 & 1.11 & 34.37  & 457.87 \\
& \appmodelsmall{HuatuoGPT-3-8B}       & 29.11 & 31.00 & 0.00 & 2.77 & 99.89  & 3548.53 \\
& \appmodelsmall{Morpheus-7B}          & 8.32  & 6.70  & 0.00 & 0.44 & 50.40  & 3679.19 \\
& \appmodelsmall{Fleming-R1-7B}        & 4.63  & 9.32  & 0.00 & 1.96 & 33.12  & 1280.85 \\
\bottomrule
\end{tabular}%
}
\vspace{-3mm}
\label{tab:appendix_l3_agent_efficiency}
\end{table*}

\section{Ablation and Performance Analysis}
\label{app:performance_analyses}

\subsection{Model Scale and Domain Specialization}
\label{app:scale_specialization}
Within individual model families, increasing model scale generally improves decision-making quality, although the gains are not uniform across all capabilities. For example, scaling Qwen3-VL from 8B to 32B improves waveform anomaly perception on L1, while Qwen3.5-27B substantially outperforms Qwen3.5-9B on both L2 average performance and L3 turn-level decision quality. However, improvements in overall clinical quality do not necessarily translate into better safety: despite the sizable gain in L2 average from Qwen3.5-9B to Qwen3.5-27B, the corresponding Safety Error Rate remains essentially unchanged (27.7 vs.\ 28.6).

And we also find that the evaluated medical-domain models do not show a consistent performance advantage over general-purpose models on AnesTRACE-Bench. For example, Qwen3.5-27B exceeds the anesthesia-specialized
Morpheus-32B in both L2 average (73.0 versus 51.3) and L3 turn-level
average (69.5 versus 46.2), and also outperforms Fleming-R1-32B
(60.3 and 52.8, respectively). The pattern is not uniform:
Fleming-R1-7B exceeds Qwen3-8B on both decision-making levels.
\subsection{Dimension-wise Performance Analysis}
\label{app:dimension_analysis}
Across L2, \emph{Evidence Grounding} is consistently the weakest clinical dimension for most model--task combinations (Table~\ref{tab:appendix_l2_task_dimensions}). This gap is especially apparent in \emph{Risk Prediction}: for example, GPT-6-Astra achieves 0.762 in Clinical Correctness but only 0.663 in Evidence Grounding. This suggests that models may identify plausible short-term risks without adequately linking them to the physiological trends and patient-specific context available at the decision point. A second gap emerges between recognizing a clinical problem and selecting an appropriate intervention. Qwen3.5-9B scores 0.587 in diagnostic correctness but only 0.418 in intervention correctness, indicating that accurate state assessment does not necessarily translate into appropriate management.

The same flaw persists in L3, where \emph{Evidence Grounding} remains lower than \emph{Clinical Correctness} (Table~\ref{tab:appendix_l3_clinical}); Qwen3.5-27B scores 61.80 versus 66.16, while GLM-4.7-Flash scores 29.12 versus 33.33. In the multi-step setting, however, grounding additionally requires integrating newly revealed observations with preceding clinical states and documented interventions, making failures of evidence integration more consequential as the episode evolves. \emph{Temporal Consistency} captures a complementary trajectory-level weakness: models such as GLM-4.7-Flash (49.85) and Morpheus-7B (46.69) also struggle to maintain or appropriately revise management across successive decision points. These results indicate that correct local clinical judgments, evidence-grounded reasoning, and coherent longitudinal management remain distinct challenges for current models.

\subsection{Perception--Reasoning Bottleneck Analysis}
\label{app:perception_reasoning}
We compare visual and structured textual representations for the same models and L2 cases (Fig.~\ref{fig:perception_impact}). All five paired models achieve higher decision quality with structured text, with average scores improving by 4.4--19.0 points and Safety Error Rates decreasing by 5.0--15.0 percentage points. The largest gap is observed for Qwen3.5-9B, whose average score increases from 46.3 to 65.3 while its Safety Error Rate decreases from 42.7\% to 27.7\%.

The complete results indicate that these gains are concentrated in clinical state interpretation and short-term risk assessment (Tables~\ref{tab:appendix_l2_mllm_task_dimensions} and~\ref{tab:appendix_l2_task_dimensions}). For Qwen3.5-9B, structured text substantially improves both correctness and evidence grounding for \emph{Risk Prediction} and \emph{Clinical Diagnosis}, whereas the improvement in \emph{Intervention Decision} correctness is comparatively modest (0.348 to 0.418). Notably, completeness for risk prediction and diagnosis is already near ceiling under visual input (0.998 and 0.996), suggesting that the low clinical scores are not primarily caused by omitted response components, but by difficulty correctly interpreting and grounding the underlying physiological evidence.
These results support the presence of an input-representation bottleneck: imperfect extraction of clinically relevant information from visual physiological inputs appears to substantially limit downstream state recognition and risk assessment.

\subsection{Quality--Efficiency Trade-offs}
\label{app:quality_efficiency}

Figure~\ref{fig:quality_latency_illustration} shows variation in
decision quality and P95 Latency across the evaluated systems. Within the Qwen3.5
family, moving from 9B to 27B increases the L3 turn-level average
from 56.6 to 69.1 and reduces Safety Error from 61.1\% to 32.5\%,
while recorded P95 Latency rises from 27.2 to 34.4 seconds
(Table~\ref{tab:l2_l3_results}). In this comparison, higher decision
quality and fewer safety errors are accompanied by a modest increase
in measured time per turn.

The complete agent results show that greater information acquisition
or generation effort does not necessarily yield better management
(Table~\ref{tab:appendix_l3_agent_efficiency}). Baichuan-M2-32B
records higher Evidence Acquisition than Qwen3.5-27B (96.0\% versus
83.3\%) and makes more tool calls per turn (7.01 versus 4.00).
Despite similar P95 Latency (33.0 versus 34.4 seconds), its
turn-level score is lower (49.5 versus 69.5) and its Safety Error
Rate is higher (66.7\% versus 32.5\%). Output length is likewise
not a proxy for clinical quality: HuatuoGPT-3-8B generates an
average of 3,549 tokens per turn, compared with 716 for
HuatuoGPT-3-32B, yet has lower turn-level quality (33.8 versus
49.9) and higher P95 Latency (99.9 versus 31.3 seconds).
These comparisons motivate reporting clinical quality, safety,
information acquisition, and computational cost as distinct
outcomes. L3 latency sums recorded model-request and tool-execution
times and is not directly comparable with the L2 response-time measure.

\begin{figure*}[t]
    \centering
    \includegraphics[width=\textwidth]{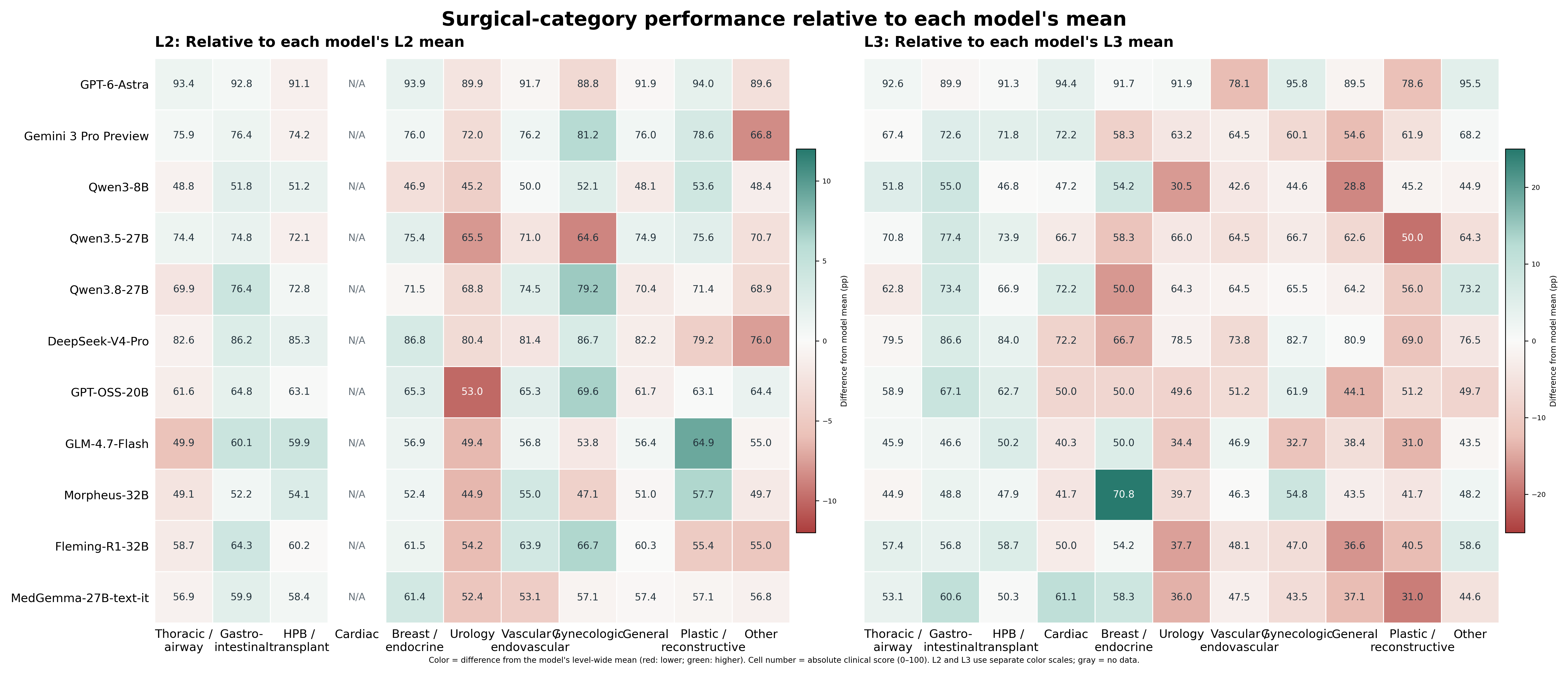}
    \caption{Performance across surgical categories in L2 and L3.
    Cell values show mean clinical scores (0--100). Colors indicate
    the difference, in percentage points, from each model's overall
    mean within the same level.}
    \vspace{-3mm}
    \label{fig:surgical_category_performance}
\end{figure*}

\subsection{Models performance in Different Surgical Subgroup}
\label{app:surgical_subgroup}
Figure~\ref{fig:surgical_category_performance} shows how clinical scores
vary across surgical categories relative to each model's mean within the
same level. In L2, urologic cases score below that mean for every model
shown; the pattern persists for most models in L3. Plastic and
reconstructive cases also receive lower relative scores in L3, whereas
several models score above their respective means on gastrointestinal
cases. For example, Qwen3.5-27B scores 77.4 on gastrointestinal cases
but 50.0 on plastic and reconstructive cases in L3. Thus, aggregate
scores obscure substantial variation across the surgical contexts
represented in the benchmark. However, because subgroup sizes, case complexity, and clinical-event composition differ across categories, these results should be interpreted as evidence of subgroup-dependent performance variation rather than an intrinsic ranking of surgical difficulty.

\begin{table*}[t]
\centering
\caption{\textbf{Error Summary}.
Distribution of primary error categories across five representative models.
}
\renewcommand{\arraystretch}{1.15}
\setlength{\tabcolsep}{4pt}
\resizebox{\textwidth}{!}{%
\begin{tabular}{lccccc}
\toprule
\textbf{Primary Error Category}
& \textbf{GPT-6-Astra}
& \textbf{DeepSeek-V4-Pro}
& \textbf{Qwen3.5-27B}
& \textbf{Qwen3.5-9B}
& \textbf{Morpheus-32B} \\
\midrule

\multicolumn{6}{l}{\textbf{L2: Single-Point Task Answers}} \\
Severity miscalibration
& 45.4\% & 24.2\% & 24.1\% & 21.4\% & 17.5\% \\
Problem misidentification
& 11.8\% & 20.6\% & 19.3\% & 17.9\% & 33.1\% \\
Inappropriate intervention
& 4.0\% & 10.4\% & 15.8\% & 18.1\% & 18.4\% \\
Observation misread
& 7.0\% & 15.5\% & 11.4\% & 15.6\% & 14.9\% \\
Unsupported evidence
& 18.3\% & 12.3\% & 8.1\% & 5.7\% & 4.7\% \\
Necessary action omitted
& 2.8\% & 9.6\% & 8.3\% & 9.6\% & 5.3\% \\

\midrule
\multicolumn{6}{l}{\textbf{L3: Turn-Level Task Answers}} \\
Severity miscalibration
& 50.2\% & 33.9\% & 26.4\% & 22.6\% & 18.8\% \\
Necessary action omitted
& 23.3\% & 22.4\% & 20.5\% & 21.6\% & 15.1\% \\
Inappropriate intervention
& 1.8\% & 8.7\% & 16.0\% & 22.3\% & 24.2\% \\
Problem misidentification
& 6.8\% & 9.6\% & 12.6\% & 13.0\% & 20.8\% \\
Observation misread
& 4.6\% & 8.3\% & 10.2\% & 9.9\% & 11.4\% \\

\midrule
\multicolumn{6}{l}{\textbf{L3: Episode-Level Temporal Errors}} \\
Longitudinal inconsistency
& 40.6\% & 28.3\% & 34.7\% & 33.1\% & 40.5\% \\
New evidence ignored
& 28.1\% & 41.3\% & 34.7\% & 37.8\% & 32.4\% \\
Prior action or response ignored
& 6.2\% & 13.0\% & 15.3\% & 17.3\% & 14.2\% \\
Escalation gap
& 12.5\% & 10.9\% & 8.3\% & 5.5\% & 8.8\% \\
Reassessment gap
& 9.4\% & 6.5\% & 6.9\% & 4.7\% & 4.1\% \\
\bottomrule
\end{tabular}%
}
\vspace{-3mm}
\label{tab:anestrace_error_taxonomy}
\end{table*}

\subsection{Error Analysis for Decision-Making Tasks}
\label{app:error_type_distribution}

To characterize decision-making failures, we analyze five representative
models spanning proprietary, general-purpose, and medical-domain systems
(Table~\ref{tab:anestrace_error_taxonomy}). Among flagged units, two
patterns recur: miscalibration of clinical severity and failure to update
management as evidence evolves. In L2, severity miscalibration accounts
for 17.5--45.4\% of flagged answers; it is the largest category for
GPT-6-Astra (45.4\%), whereas problem misidentification is most frequent
for Morpheus-32B (33.1\%). In L3, severity miscalibration remains common
at the turn level (18.8--50.2\%). At the episode level, ignoring new
evidence (28.1--41.3\%) and longitudinal inconsistency (28.3--40.6\%)
account for substantial shares of flagged trajectories. These findings
point to persistent difficulty in assessing clinical urgency and revising
decisions across successive patient states.

\begin{figure*}[t]
\centering
\includegraphics[width=0.94\textwidth]{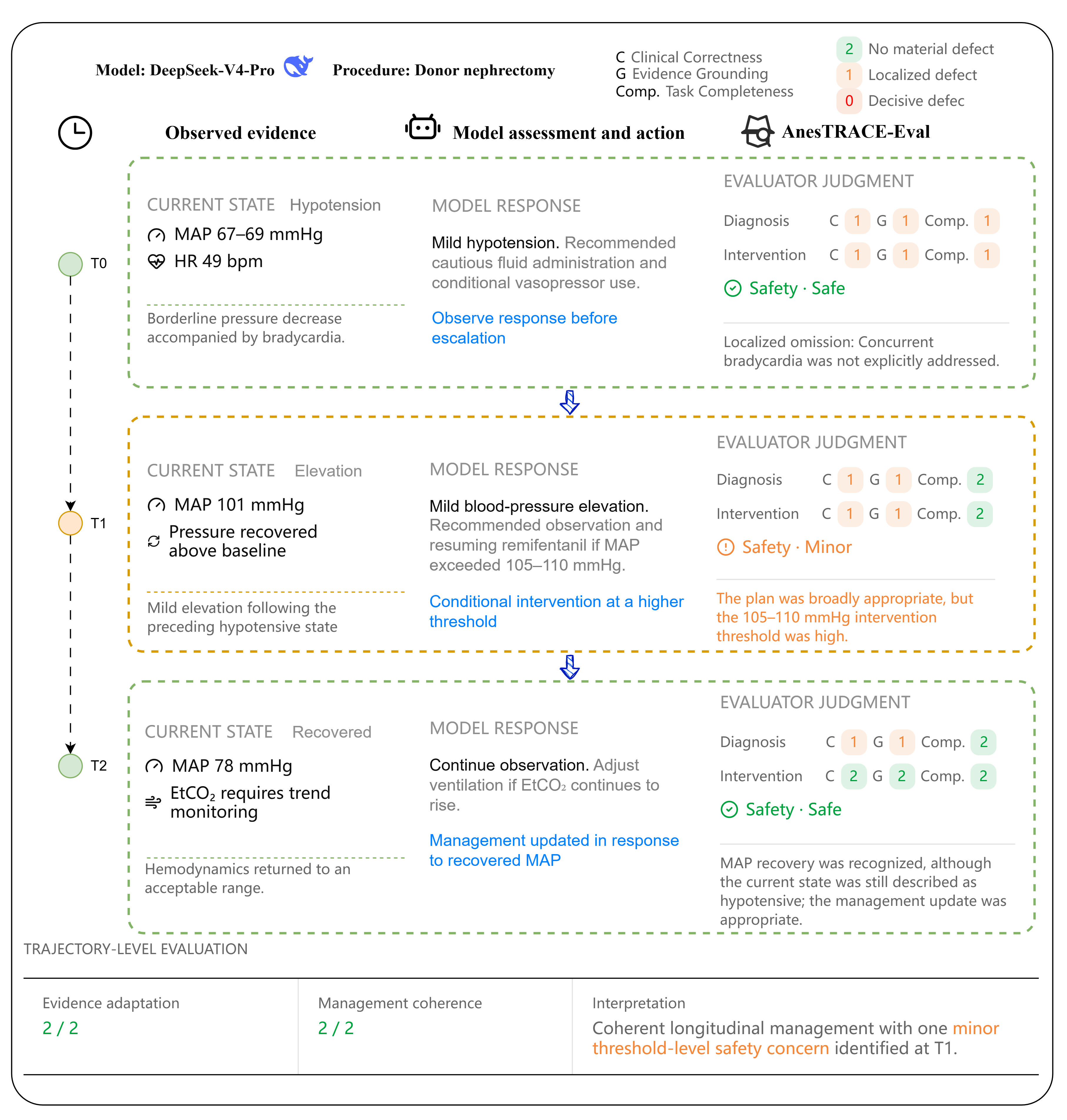}
\caption{Turn- and trajectory-level evaluation of a representative L3 episode.
DeepSeek-V4-Pro with full tools adapted its management across a
donor-nephrectomy trajectory, while AnesTRACE-Eval identified localized
turn-level defects and a minor safety concern at T1.}
\label{fig:representative_l3_evaluation}
\vspace{-3mm}
\end{figure*}

\section{Qualitative Case Study}
\label{app:case_studies}

\textbf{Representative Multi-step Episode}. Figure~\ref{fig:representative_l3_evaluation} illustrates the turn- and
trajectory-level assessment of a three-turn donor-nephrectomy episode from the
linked VitalDB--INSPIRE records. It aligns the evolving clinical evidence with
the model's assessment and action at each turn, allowing AnesTRACE-Eval to
distinguish localized response defects from the coherence of the overall
trajectory.

The retained L3 records include the observation, model response, turn-level
rubric judgment, safety label, and trajectory-level assessment. Figure
\ref{fig:representative_l3_evaluation} presents a three-turn donor-nephrectomy
episode from linked VitalDB--INSPIRE records, illustrating how AnesTRACE-Eval
distinguishes localized defects from overall trajectory quality.

The evaluator considered the proposed management safe at T0 and T2, while
identifying a minor concern at T1 because the proposed intervention threshold
could delay treatment. Despite these localized defects, it assigned full scores
to both evidence adaptation and management coherence across the trajectory.
This example illustrates how turn-level rubric scores and safety labels
complement trajectory-level assessment without reducing the episode to a
single failure label.

\end{document}